\documentclass[pdflatex,sn-mathphys-num]{sn-jnl}
\usepackage{graphicx}%
\usepackage{multirow}%
\usepackage{amsmath,amssymb,amsfonts}%
\usepackage{amsthm}%
\usepackage{mathrsfs}%
\usepackage[title]{appendix}%
\usepackage{xcolor}%
\usepackage{textcomp}%
\usepackage{manyfoot}%
\usepackage{booktabs}%
\usepackage{algorithm}%
\usepackage{algorithmicx}%
\usepackage{algpseudocode}%
\usepackage{listings}%
\usepackage{tabularx}
\usepackage{adjustbox}
\usepackage[algo2e]{algorithm2e} 
\usepackage{multicol}
\newtheorem{lemma}{\bf Lemma}[section]
\newtheorem{theorem}{Theorem}%  meant for continuous numbers
\begin{document}

\title[Article Title]{Epidemic-guided deep learning for spatiotemporal forecasting of Tuberculosis outbreak}


\author[1]{\fnm{Madhab} \sur{Barman}} %\email{mat18d004@iiitdm.ac.in}
\equalcont{These authors contributed equally to this work.}

\author[1]{\fnm{Madhurima} \sur{Panja}} %\email{madhurima.panja@sorbonne.ae}
\equalcont{These authors contributed equally to this work.}

\author[2]{\fnm{Nachiketa} \sur{Mishra}} %\email{mishra.nachiketa@gmail.com}

\author*[1,3]{\fnm{Tanujit} \sur{Chakraborty}} \email{tanujit.chakraborty@sorbonne.ae}

\affil[1]{\orgname{SAFIR, Sorbonne University Abu Dhabi}, \orgaddress{\country{UAE}}} % \orgdiv{Department of Science and Engineering}

\affil[2]{\orgname{IIITDM Kancheepuram}, \orgaddress{\city{Chennai}, \country{India}}} % \orgdiv{Department of Science and Humanities}, , 
%Indian Institute of Information Technology, Design and Manufacturing,

\affil[3]{\orgdiv{Sorbonne Center for Artificial Intelligence}, \orgname{Sorbonne University}, \orgaddress{\country{France}}}

\abstract{
Tuberculosis (TB) remains a formidable global health challenge, driven by complex spatiotemporal transmission dynamics and influenced by factors such as population mobility and behavioral changes. We propose an Epidemic-Guided Deep Learning (EGDL) approach that fuses mechanistic epidemiological principles with advanced deep learning techniques to enhance early warning systems and intervention strategies for TB outbreaks. Our framework is built upon a modified networked Susceptible-Infectious-Recovered (MN-SIR) model augmented with a saturated incidence rate and graph Laplacian diffusion, capturing both long-term transmission dynamics and region-specific population mobility patterns. Compartmental model parameters are rigorously estimated using Bayesian inference via the Markov Chain Monte Carlo approach. Theoretical analysis leveraging the comparison principle and Green’s formula establishes global stability properties of the disease-free and endemic equilibria. Building on these epidemiological insights, we design two forecasting architectures, EGDL-Parallel and EGDL-Series, that integrate the mechanistic outputs of the MN-SIR model within deep neural networks. This integration mitigates the overfitting risks commonly encountered in data-driven methods and filters out noise inherent in surveillance data, resulting in reliable forecasts of real-world epidemic trends. Experiments conducted on TB incidence data from 47 prefectures in Japan and 31 provinces in mainland China demonstrate that our approach delivers robust and accurate predictions across multiple time horizons (short to medium-term forecasts), supporting its generalizability across regions with different population dynamics. Additionally, incorporating uncertainty quantification through conformal prediction and explainability via temporal gradient-based class activation maps enhances the model’s practical utility for guiding targeted public health interventions.
}


% Other Journals \keywords{Tuberculosis, graph Laplacian, global stability, epidemic-informed machine learning, time series forecasting}



\maketitle

\section{Introduction}\label{sec1}
Tuberculosis (TB), caused by Mycobacterium tuberculosis complex, is a chronic infectious disease primarily transmitted through aerosols produced by coughing \cite{dheda2014global}. In 2023, the World Health Organization (WHO) reported 8.2 million new TB cases globally, a significant increase from 7.5 million in 2022\footnote{\url{https://www.who.int/teams/global-tuberculosis-programme/tb-reports}}. This makes TB the leading cause of death from a single infectious disease, surpassing COVID-19. To support the Sustainable Development Goal 3: ``Good Health and Well-being'' and to address the global health burden of high TB incidence rate, the WHO introduced the ``End TB strategy'' in 2015. This initiative aims to reduce TB incidence cases below 0.01\% by 2035 and 0.001\% by 2050 \cite{uplekar2015s}. However, the current global rate of decline is not on track to achieve these goals. In the era of rapid globalization, the existence of a TB-free region is impractical unless a global mitigation approach is successfully implemented. While low and medium-income countries are traditionally considered high TB-burden regions, TB remains a persistent challenge even in high-income countries due to increased population mobility \cite{pareek2016impact}. Among the high-income nations, Japan continues to be classified as a TB middle-burden country, with an incidence rate of 10-49 cases per 100,000 population\footnote{\url{https://iris.who.int/handle/10665/341980}}. High TB incidences and related mortality in Japan are primarily attributed to an aging population \cite{hagiya2019trends}, as well as the rising number of immigrants from high TB-burden countries, accounting for a growing proportion of cases among younger populations \cite{ota2021epidemiology}. {\color{black}Alongside Japan, mainland China remains one of the countries with a high-burden of TB incidences owing to its large population, latent infections, and disparities in healthcare access across regions \cite{ma2023influential}.} To mitigate the health and socio-economic consequences of TB, various attempts have been made to develop early warning systems that predict the future TB dynamics \cite{zheng2015forecast, wang2017hybrid}. An early warning system provides quick hindsight, allowing for timely intervention and better disease control in its nascent stages. Epidemic modeling and forecasting are central to these efforts, as they enable more accurate predictions of disease trends, guide public health responses, and help allocate resources more effectively to reduce the TB transmission rate.

Compartmental epidemiological models are widely used to characterize infectious disease dynamics by dividing the population into distinct and mutually exclusive subsets based on their infection status \cite{grassly2008mathematical, brauer2019mathematical}. The compartmental SIR model categorizes individuals as Susceptible ($S$), those who are vulnerable to infection; Infectious ($I$), those who are currently infected and actively transmitting the disease; and Recovered ($R$), those who have gained immunity after recovering from the disease \cite{kermack1927contribution}. Numerous extensions of the classical SIR model, including additional compartments and refined incidence functions, have been developed to model TB transmission dynamics in Indonesia \cite{side2018numerical} and Kazakhstan \cite{kalizhanova2024modeling}. Although compartmental epidemiological models yield fundamental insights into infectious disease outbreaks, they are more suited for understanding the disease dynamics, rather than real-time forecasting of the disease incidences \cite{keeling2008modeling, ghosh2025developing}. To overcome the problem of limited predictability of the compartmental approaches, data-centric forecasting methods, including statistical and deep learning models, have emerged as powerful tools. For example, \cite{abade2024comparative} demonstrated the effectiveness of deep learning frameworks in forecasting TB infections in Brazil, while \cite{wang2017hybrid} implemented a data-driven hybrid framework to predict TB incidence in China. These approaches leverage diverse data sources to uncover hidden patterns in epidemic data, showcasing their flexibility. However, this flexibility can often be a double-edged sword, making them prone to overfitting, particularly in low-data regimes, a common challenge in epidemic forecasting \cite{rodriguez2024machine}. Additionally, these models often generate unrealistic predictions for non-stationary systems as they solely rely on lagged values of disease incidences. The primary disadvantage of data-driven methods for epidemic forecasting is that they are not guided by the mechanisms governing disease transmission. By focusing solely on the dynamics of surveillance data without integrating epidemiological principles, these approaches often generate unrealistic forecasts and struggle to differentiate genuine trends from noise introduced during data collection \cite{qian2025physics}. To bridge this gap, a recent line of research has explored combining epidemic dynamics from compartmental models into data-centric forecasting frameworks \cite{ye2025integrating, delli2022hybrid}. For instance, Epidemic-Informed Neural Networks (EINNs) embed epidemiological knowledge from ordinary differential equation (ODE) based compartmental models into Recurrent Neural Networks (RNN) via gradient matching, enhancing model flexibility and resilience against noisy data \cite{rodriguez2023einns}. Similarly, \cite{qian2025physics} employed Physics Informed Neural Networks (PINNs) to incorporate epidemiological principles from a modified SIR framework with observed infectious data, improving forecast accuracy through transfer learning.

Despite these advancements, most studies in this domain rely on traditional SIR-based compartmental models, which do not account for spatial dynamics, such as population mobility. As a result, these models often fail to capture the spatiotemporal trajectory of an epidemic, limiting their ability to predict real-world transmission patterns accurately. Recent epidemiological studies have incorporated network-based structures into SIR models to address this limitation in compartmental models, enhancing their ability to capture spatial disease transmission \cite{pastor2015epidemic, keeling2005networks}. By leveraging network structures, the SIR model can estimate epidemic dynamics at the population level rather than focusing solely on individual-level infection status, leading to a more comprehensive understanding of disease spread. To represent the spatial disease dynamics, previous studies have often relied on Laplacian operators to model population mobility, assuming isotropic diffusion and uniform movement probabilities \cite{li2018diffusive, lei2018theoretical, allen2007asymptotic}. However, these assumptions are frequently violated in real-world scenarios, where human mobility is influenced by local environmental factors such as lockdowns, travel restrictions, and other region-specific circumstances. To address these challenges, recent studies have extended the global stability theory of the SIR model by incorporating graph Laplacian diffusion, leading to more efficient frameworks for modeling spatial epidemic dynamics \cite{tian2020global, barman2024network}. In existing networked epidemic models, bilinear incidence rate ($\beta S I$) and standard bilinear incidence rate ($\beta S I/N$), with transmission rate $\beta$ and population size $N$, are widely used to analyze the global dynamics of disease spread \cite{liu2020weighted, tian2023asymptotic}. However, these incidence functions fail to account for behavioral changes and crowding effects, often leading to overestimated infection rates in highly connected populations. {\color{black}To overcome these limitations, we adopt a saturated incidence function \cite{capasso1978generalization} and formulate a networked SIR model by incorporating graph Laplacian diffusion, we name it modified networked SIR (MN-SIR) model}. This integration better captures long-term transmission dynamics and provides an adaptive framework for modeling dynamic changes in population behavior. Building on this epidemiological foundation, we further introduce the Epidemic-Guided Deep Learning (EGDL) approach, which combines spatial epidemiological insights with data-driven forecasting techniques. EGDL generates accurate epidemic forecasts by considering both spatial dynamics and temporal interactions. The proposed EGDL architectures integrate the long-term infection dynamics of the MN-SIR model as auxiliary information within deep learning frameworks. This integration combines the flexibility of deep learning techniques and the epidemiological principles captured by the MN-SIR model, improving spatiotemporal disease incidence forecasting. Furthermore, using epidemic dynamics as auxiliary information, instead of solely relying on gradient-based learning, restricts the overfitting risks in EGDL frameworks, allowing the model to learn robust representations. This effectively filters out noise from historical incidence data and aligns spatiotemporal disease forecasts with real-world epidemic trends. The key contributions of the study can be summarized as follows:
% \begin{itemize}
%     \item We introduce an extended version of the networked SIR model by incorporating a saturated incidence rate and graph Laplacian diffusion to understand epidemic data's spatiotemporal dynamics. We validate the key epidemiological properties such as positivity and boundedness. The global stability of disease-free and endemic equilibria is established using Green's formula and the comparison principle.
%     \item To enhance spatiotemporal disease forecasting, we propose the EGDL-Parallel and EGDL-Series architectures, combining the networked SIR model with the historical surveillance data-driven deep learning frameworks (e.g., Transformers, NBeats, etc.).
%     % \item We study the global features of TB incidence data across 47 prefectures in Japan and employ the Granger causality test and wavelet coherence plots to assess the causal relationship between the networked SIR model output and active TB cases.
%     \item The forecasting performance of the EGDL architectures is evaluated using a rolling window approach across four test horizons and compared against data-centric models, with robustness validated through non-parametric statistical tests. The uncertainty of the EGDL frameworks is quantified using the conformal prediction approach.
% \end{itemize}

\begin{itemize}
    \item We introduce a MN-SIR model by incorporating a saturated incidence rate and graph Laplacian diffusion to understand epidemic data's spatiotemporal dynamics. We validate the key epidemiological properties, such as positivity and boundedness. The global stability of disease-free and endemic equilibria is established using Green's formula and the comparison principle.
    \item To enhance spatiotemporal disease forecasting, we propose the EGDL-Parallel and EGDL-Series architectures, combining the MN-SIR model with the historical surveillance data-driven deep learning frameworks (e.g., Transformers, NBeats).
    \item {\color{black}The forecasting performance of the EGDL architectures is evaluated using real-world TB incidence datasets from 47 prefectures of Japan and 31 provinces of mainland China via a rolling window approach across four test horizons. We compared the performance the EGDL frameworks against data-centric models and validated the robustness through a non-parametric statistical test. The uncertainty of the EGDL frameworks is quantified using the conformal prediction approach, while the explainability of the model output is obtained via temporal gradient-based class activation maps.}
    
\end{itemize}

\section{Motivating Examples}\label{data}
\begin{figure}[!ht]
    \centering
    \includegraphics[width=0.99\linewidth]{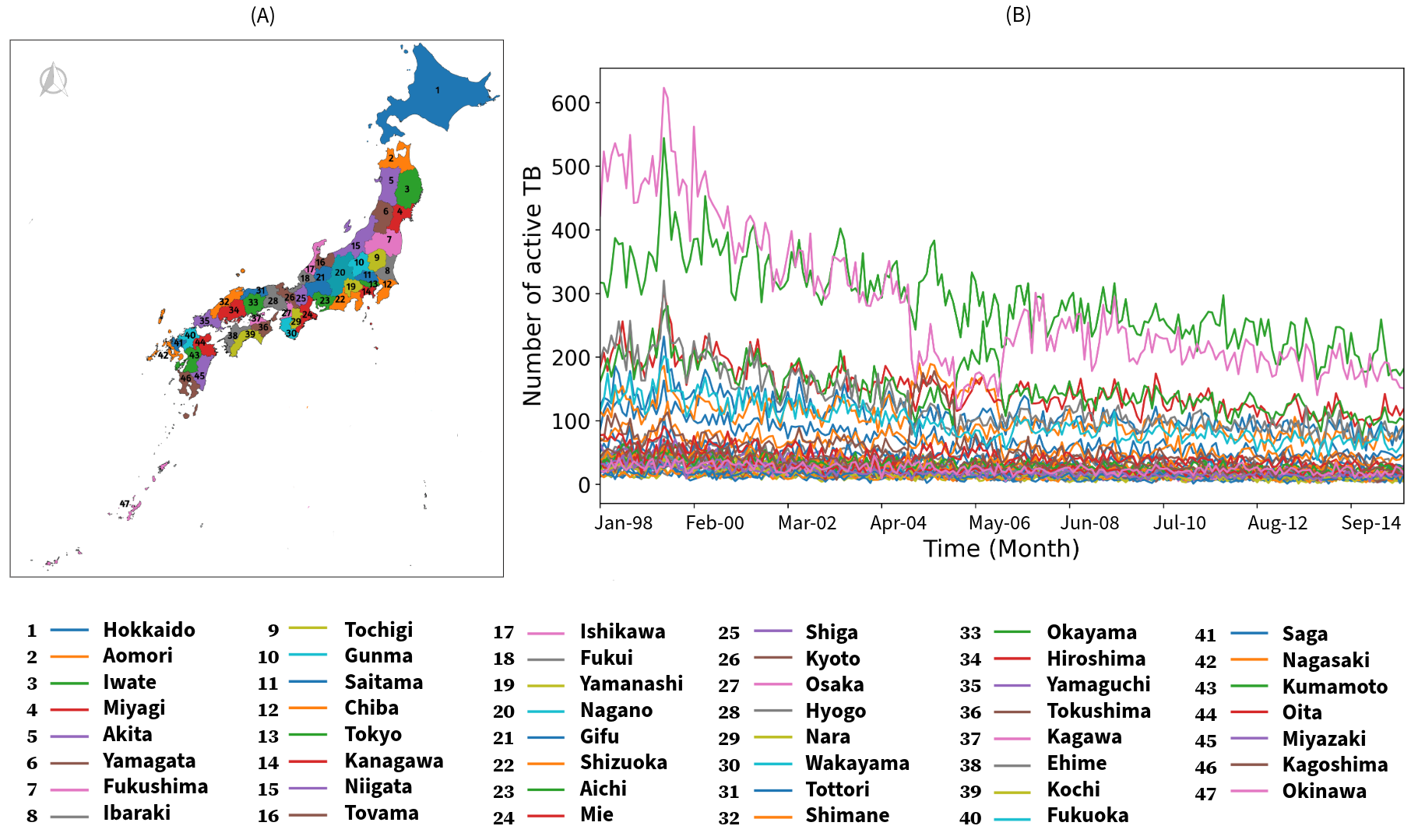}
    \caption{{\color{black}(A) Geographic distribution of Japan's 47 prefectures, shown for illustrative purposes only, without implying any political assertions on Japan's territorial boundaries. (B) Monthly active tuberculosis (TB) cases were recorded in each of Japan's 47 prefectures from January 1998 to December 2015.}}
    \label{fig:TB47}
\end{figure}

\begin{figure}[!ht]
    \centering
    \includegraphics[width=0.9\linewidth]{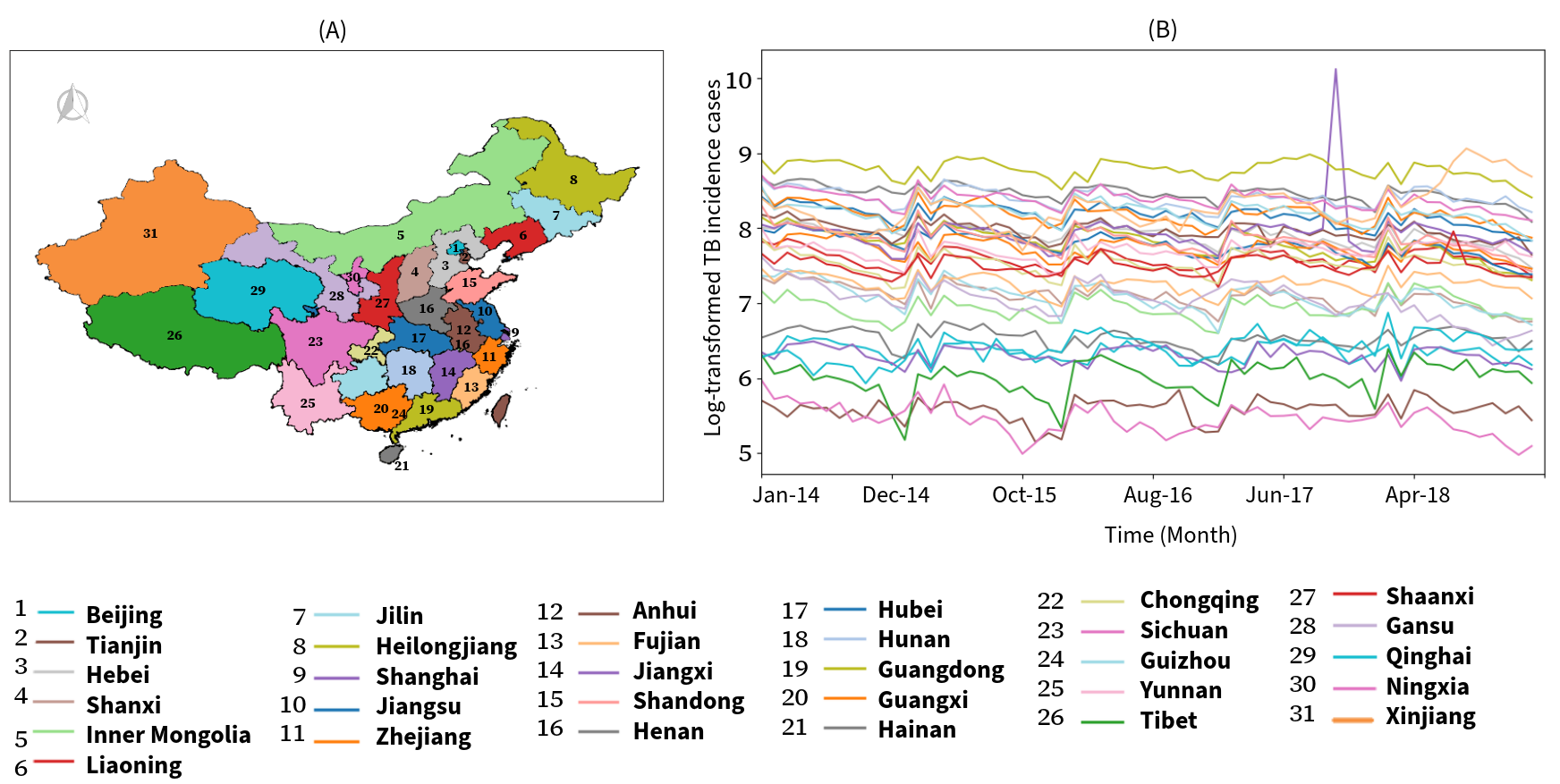}
    \caption{{\color{black}(A) Geographic distribution of 31 provinces of mainland China, shown for illustrative purposes only, without implying any political assertions on China's territorial boundaries. (B) Log-transformed values of monthly active tuberculosis (TB) cases recorded in each of China's 31 provinces from January 2014 to December 2018 (log transformation is applied only for visual clarity).}}
    \label{Fig_China_MAP}
\end{figure}

TB remains one of the leading causes of infectious disease-related mortality, posing a significant global public health challenge. {\color{black}This study focuses on TB incidence across two major Asian countries, Japan and mainland China, which represent contrasting TB burden levels and demographic profiles.} Japan, classified as a TB medium-burden country, reported 11,519 new TB cases in 2022\footnote{\url{https://jata-ekigaku.jp/english/tb-in-japan}}. To control the rapid spread of TB, Japan has established a robust nationwide surveillance system that collects granular data at the prefectural level, facilitated by the country’s administrative structure of 47 prefectures. This hierarchical setup enables precise geographic and temporal monitoring of TB trends, enabling timely public health responses. In this study, we analyze the monthly counts of newly registered active TB cases (of all forms) across Japan’s 47 prefectures, as illustrated in Fig. \ref{fig:TB47}(A). The dataset spans from January 1998 to December 2015 and is publicly available at \cite{data_monthly, sumi2019time}. Fig. \ref{fig:TB47}(B) displays monthly TB cases for each prefecture, with distinct colors to facilitate regional comparison. {\color{black}In addition to Japan, we analyze TB incidence in mainland China, a high-burden country that ranked third globally in estimated TB cases in 2022\footnote{\href{https://www.who.int/teams/global-programme-on-tuberculosis-and-lung-health/tb-reports/global-tuberculosis-report-2023/tb-disease-burden/1-1-tb-incidence}{TB Disease Burden (WHO).}}. Despite steady progress in TB prevention, the disease remains a major public health concern due to China’s large population and the widespread prevalence of latent infections. Consequently, TB is classified as a Class B respiratory infectious disease in China, necessitating ongoing monitoring \cite{ma2023influential}. Additionally, the COVID-19 pandemic severely disrupted healthcare systems, affecting the continuity and reliability of TB surveillance. To ensure data consistency, we restrict our analysis to the pre-pandemic period, focusing on monthly notified TB cases from 31 provinces between January 2014 and December 2018, which were collected from the Chinese Center for Disease Control and Prevention\footnote{\url{https://en.chinacdc.cn/}} \cite{ma2023influential}. Fig. \ref{Fig_China_MAP} displays the TB incidence trends across these provinces during the study period, with each province uniquely color-coded to highlight regional variations. Together, the TB datasets from Japan and China offer complementary insights into disease dynamics across East Asia, ranging from a medium-burden, aging society to a high-burden, populous nation facing latent infections. This regional heterogeneity presents a challenging spatiotemporal forecasting problem. Our study aims to solve this problem by developing an integrated solution, based on epidemiological models and deep learning approaches, that can accurately predict future TB transmission patterns.}

\section{Preliminaries}\label{Sec_Prelim}
Epidemic modeling and forecasting approaches can be broadly categorized as mechanistic (or compartmental) frameworks and phenomenological models. Mechanistic models use causal frameworks to describe disease states and understand the surveillance data-generating process. In contrast, phenomenological models focus on directly modeling surveillance data (past lagged observations of the epidemic time series) without explicitly incorporating the underlying epidemiological mechanisms. The following subsections provide a brief overview of the mechanistic frameworks (Section \ref{Sec_Epi_model_Review}), spatiotemporal forecasting task (Section \ref{Prelim_Spatiotemporal_Forecasting}), and phenomenological deep learning models (Section \ref{Sec_Phenom_Review}) to be used as building blocks for the proposed EGDL frameworks.

% This model provides the mathematical foundation for understanding the propagation, stabilization, and eventual resolution of disease outbreaks, driven by critical parameters such as transmission and recovery rates as follows:

\subsection{Epidemic Models: A Review} \label{Sec_Epi_model_Review}
The SIR model, developed by Kermack and McKendrick \cite{kermack1927contribution}, serves as a foundational framework for analyzing the temporal dynamics of epidemic outbreaks. The model is governed by the following system of differential equations:
% \begin{subequations}\label{SIR}
% \begin{align}
%     \dfrac{dS}{dt}  = -\lambda S,\quad
%     \dfrac{dI}{dt}  = \lambda S  -\gamma I, \quad
%     \dfrac{dR}{dt}  = \gamma I,
% \end{align}
% \end{subequations}
\begin{equation}\label{SIR}
    \dfrac{dS}{dt}  = -\lambda S,\quad
    \dfrac{dI}{dt}  = \lambda S  -\gamma I, \quad
    \dfrac{dR}{dt}  = \gamma I,
\end{equation}
where $\lambda$ denotes the force of infection which measures the rate at which susceptible individuals contract the infection and $\gamma$ is the recovery rate. The force of infection is commonly formulated as a function of the number of infected individuals and encapsulates the dynamics of interactions leading to the transmission of the infection. This force function $\lambda (I)$ can determined as:
\begin{align}\label{force}
    \lambda = \beta_0 P(N) {I}/{N},
\end{align}
where $\beta_0$ denotes the probability of transmission per contact between a susceptible and an infected individual, $P(N)$ is the contact rate, defined as the average number of contacts adequate for disease transmission by an individual per unit time \cite{ma2009dynamical}.
 If \( P(N) = cN \), where \( c \) is a positive constant, this implies that the contact rate is directly proportional to the total population size. Consequently, the force of infection in Eq. 
 \eqref{force} becomes $\lambda = \beta_0 c I = \beta I,$ where \( \beta = \beta_0 c \) is referred to as the transmission coefficient. Thus, the incidence is represented as  $\beta S I$, commonly called bilinear incidence. Furthermore, the above model in Eq. \eqref{SIR} can be extended by incorporating vital dynamics, including the birth rate ($\Lambda$) and death rate ($\mu$), along with a saturated incidence rate as proposed in \cite{capasso1978generalization}. The SIR epidemic model with a saturated incidence rate can be expressed as:
\begin{equation}\label{Sat_SIR}
% \begin{align}
    \dfrac{dS}{dt} = \Lambda -\dfrac{\beta SI}{1 + \alpha I} - \mu S, \; \quad
    \dfrac{dI}{dt} = \dfrac{\beta SI}{1 + \alpha I} -(\gamma + \mu) I, \; \quad
    \dfrac{dR}{dt} = \gamma I - \mu R,
% \end{align}
\end{equation}
where $\alpha$  represents the saturation factor. {\color{black}We assume that $\Lambda$ represents the constant recruitment or birth rate into the susceptible population at each node. The parameter $\mu$ is the natural death rate, included in all compartments to model uniform mortality. The term $(\gamma + \mu)I$ accounts for the removal of infected individuals due to recovery ($\gamma$) or death ($\mu$), while $\gamma I$ contributes to the recovered class.} This model also assumes homogeneous mixing among individuals, meaning that every susceptible individual is equally likely to interact with any other individual in the population at any given time. This assumption simplifies the representation of contact patterns, treating all interactions as uniformly distributed across the population. In real-world population dynamics, contact patterns are often heterogeneous rather than homogeneous. Building upon this basic framework, heterogeneity can be introduced by further subdividing the compartments to account for diverse population characteristics \cite{anderson1988epidemiology}. Typically, the number of contacts each individual has is significantly smaller than the total population size. In such cases, the homogeneous-mixing assumption becomes inefficient. The simple SIR model describes the temporal dynamics of an infectious disease in a population; however, it fails to capture the spatial dynamics of disease incidence. Epidemic models incorporating network structures address this limitation by assigning each individual a finite set of permanent contacts within a localized region, providing a more accurate depiction of infection dynamics.

{\color{black}\subsection{Spatiotemporal Forecasting: Problem Definition}\label{Prelim_Spatiotemporal_Forecasting}
Time series data consists of a sequence of observations recorded at uniformly spaced time intervals \cite{hyndman2018forecasting}. Based on the number of variables measured at each time step, temporal datasets are broadly categorized into univariate and multivariate time series. A univariate time series contains a sequence of scalar observations collected over time, whereas a multivariate time series comprises sequences of vector-valued observations, where multiple variables are recorded simultaneously.
Let $V = \{1, 2, \dots, n\}$ denote a set of spatial locations (e.g., prefectures or provinces), and let $\mathcal{T} = \{1, 2, \dots, T\}$ denote discrete and equally spaced time points. A univariate time series at location $x \in V$ is a sequence $\{Y(x, t)\}_{t \in \mathcal{T}}$, where $Y(x, t) \in \mathbb{R}_{\geq 0}$ represents the observed number of TB cases at location $x$ and time $t$. A multivariate time series refers to a sequence of vectors $\{\mathbf{y}_t\}_{t=1}^T$, where $\mathbf{y}_t \in \mathbb{R}^d$ encodes $d$-dimensional observations at time $t$. In our setup, this corresponds to stacking TB incidence values from all locations at time $t$, i.e., $\mathbf{y}_t = [Y(1, t), Y(2, t), \dots, Y(n, t)]^\top \in \mathbb{R}^n$, where $d = n$ is the number of spatial units. With these multiple variables observed across different spatial locations over time, the multivariate time series naturally extends to a spatiotemporal dataset. The collection $\{Y(x, t): x \in V, t \in \mathcal{T} \}$ forms a spatiotemporal time series, which captures both temporal dynamics (evolution over time) and spatial interactions (across locations) \cite{jin2024survey}. One of the primary tasks in time series analysis is forecasting, which involves predicting future values based on historical observations. Depending on the nature of the temporal data, forecasting can be categorized into univariate and multivariate forecasting. Univariate forecasting focuses on predicting future values of a single time series using only its past observations. In contrast, multivariate forecasting predicts target variables by leveraging historical data from multiple related time series, capturing the inter-dependencies among correlated variables and often improving forecast accuracy. 

Spatiotemporal forecasting further extends the multivariate forecasting task by jointly modeling multiple interrelated time series distributed across both space and time, thereby capturing complex spatial and temporal patterns. The multi-step ahead forecasting task is defined as follows: given historical data up to time $T$, i.e., $\{Y(x, t)\}_{x \in V, t \leq T}$, the objective is to predict future values $\{\widehat{Y}(x, T + v)\}_{v=1}^q$ for each location $x \in V$, where $q \in \{1, 2, \ldots\}$ is the forecasting horizon. This is a multi-output forecasting problem, involving the simultaneous prediction of future values across multiple spatial units and time steps \cite{panja2024stgcn}. % Let $f: \mathbb{R}^{n \times w} \rightarrow \mathbb{R}^{n \times q}$ be a global forecasting model that takes a matrix of past observations over a rolling window of size $w$ and outputs forecasts for the next $q$ time steps, i.e., 
% $
% \hat{\mathbf{Y}}_{T+1:T+q} = f(\mathbf{Y}_{T-w+1:T}) = f\left( [\mathbf{y}_{T-w+1}, \dots, \mathbf{y}_T] \right),
% $
% where $\hat{\mathbf{Y}}_{T+1:T+q} \in \mathbb{R}^{n \times q}$ is the predicted future incidence matrix. Note that the above formulation is consistent with the state-of-the-art multivaraite forecasting models and also generalizable to spatiotemporal variants considered in this study. 
Let $f_{M} \in \mathcal{F}$ be a forecasting model, where $\mathcal{F}$ denotes the class of data-driven global forecasting architectures (for e.g. Transformers, TCN, NBeats, and NHits). The model $f_{M}$ takes as input a sequence of past observations over a rolling window of size $t_w$ and generates forecasts for the next $q$ time steps. Thus using the historical data $\{Y(x, t)\}_{x \in V, t \leq T}$ for training $f_{M}$ the future dynamics can be predicted as follows: 
\begin{equation}\label{Eq_Fore_Data}
    \widehat{Y}\left(x, \tilde{t} + v\right) = f_{M}\left(\left\{Y\left(x, \tilde{t} - t_0 \right)\right\}_{t_0 = 0}^{t_w - 1}\right),
\end{equation}
where $x \in V, \; \tilde{t} = t_w, t_w+1, \ldots, T, \; v = 1, 2, \ldots, q$, and $t_w$ denotes the number of historical inputs used for predicting the subsequent observations. Note that the above formulation is consistent with the state-of-the-art multivariate forecasting models and also generalizable to spatiotemporal variants considered in this study. 
}

\subsection{Deep Learning Models for Time Series Forecasting} \label{Sec_Phenom_Review}
% , which leverages historical data to predict future trends,

Time series forecasting has become a crucial area of research in public health, particularly for modeling the spread and impact of infectious diseases. The increasing availability of epidemic-related data has supported the development of data-driven forecasting methods aimed at early detection of abnormal disease patterns and mitigation of public health risks \cite{chakraborty2020real, rosenfeld2021epidemic, panja2023epicasting}. These methods are essential for timely interventions and effective surveillance systems. Recently, deep learning approaches have emerged as powerful tools due to their ability to capture complex, non-linear dependencies in high-dimensional time series data, often outperforming traditional statistical models \cite{rodriguez2024machine}. Architectural innovations in deep learning-based forecasting have taken various forms. Encoder-decoder models with attention mechanisms, most notably Transformers, have proven effective in learning long-range temporal dependencies \cite{vaswani2017attention, wu2020deep}. These models process input sequences using positional encodings and stacked self-attention layers, followed by decoder components that leverage look-ahead masking to ensure predictions depend only on past observations. Similarly, Temporal Convolutional Networks (TCNs) offer an alternative by using dilated causal convolutions within residual blocks to extract temporal features and model uncertainty in sequential data \cite{chen2020probabilistic}. On the other hand, attention-free and convolution-free models with deep stacks of fully connected layers have also shown strong performance in capturing input-output relationships while implicitly prioritizing resource-efficient computations. Among these models, the Neural Basis Expansion for Time Series (NBeats) framework uses a block-based architecture to forecast time series data through residual learning \cite{oreshkin2019n}. It consists of stacked blocks, where the initial block models the input time series and generates forward and backward outputs. Each subsequent block refines the residuals of the previous one. Within each block, dense layers with ReLU activations produce expansion coefficients, which are then passed through backward and forward basis layers to generate the backcast and forecast. Building upon this architecture, the Neural Hierarchical Interpolation for Time Series (NHits) framework introduces a multi-rate signal sampling approach to improve long-range forecasting \cite{challu2023nhits}. Each NHits block comprises a Multilayer Perceptron (MLP) followed by MaxPool operations to focus on analyzing low-frequency components crucial for stable long-range forecasting.

The architectural designs of these scalable deep learning forecasters have demonstrated superior performance in capturing long-term dependencies of epidemic datasets. For instance, Transformers have been shown to accurately forecast influenza cases \cite{wu2020deep, sasal2022w}, while attention mechanisms have been utilized for predicting dengue incidence in Vietnam \cite{nguyen2022deep}. Similarly, a modified NBeats framework was developed to forecast COVID-19 hospitalizations in Canada \cite{motavali2023dsa}, and the NHits framework has been applied to provide precise forecasts of COVID-19 incidence and mortality in Brazil \cite{souza2023deep}. In addition to these advancements, there has been a growing interest in modeling spatio-temporal dependencies in epidemic data. Traditional methods, such as the Generalized Spatiotemporal Autoregressive (GSTAR) \cite{imro2023determination} and Fast Gaussian Process (GpGp) \cite{senanayake2016predicting} models, have been used to capture spatial and temporal patterns. Modern deep learning architectures, such as Spatiotemporal Graph Convolutional Networks (STGCN), have further enhanced the ability to model these complex relationships \cite{wang2022causalgnn}. A key challenge remains to include the epidemiological knowledge and spatial dynamics of the disease inside the deep learning frameworks so that they can be used for informed decision-making in public health departments. %This study combines a networked SIR model with deep learning-based forecasters to model, analyze, and forecast TB outbreaks and builds a real-time disease monitoring system.

\section{Modified Networked SIR (MN-SIR) Model} \label{Network_SIR_Model}
The networked epidemic model with graph Laplacian diffusion offers significant advantages over classical epidemic models by incorporating the spatial and network structure of populations \cite{keeling2008modeling}. Classical models typically assume a well-mixed population, where interactions occur uniformly among individuals. In contrast, network-based models represent individuals as nodes connected by edges, where the edges capture the interaction or movement between individuals or regions. This approach enables the incorporation of spatial heterogeneity and connectivity patterns, allowing for a more realistic representation of disease dynamics across communities or geographic regions.  Unlike classical models, which often rely on uniform diffusion assumptions, the graph Laplacian approach reflects localized disease transmission and varying intensities of disease spread. This is particularly important in fragmented regions, where isolated communities or uneven connectivity can significantly influence epidemic profiles.

\subsection{MN-SIR Model Formulation}
The model begins by considering a standard weighted, connected, and undirected finite graph  $G:=\langle {V}, {E} \rangle$ without self-loops, where $V$ represents the set of nodes (or vertices) with a total of $n$ nodes, and $E$ denotes the set of edges. Taking the graph as a spatial structure, we define the graph Laplacian operator ($\Delta$) acting on a function $h$  from continuous space to a finite graph as follows:
\begin{align}\label{hfun}
    \Delta h(x) = \sum_{{y \in V, y \sim x}} \left[ h(y) - h(x)\right],
\end{align}
where $y \sim x$ describes node $y$ is adjacent to node $x$ and $h$ is a function such that $h : V \to \mathbb{R}$, where $\mathbb{R}$ represents the set of real numbers \cite{tian2020global, liu2020weighted}. The operator $\Delta$ describes population mobility between regions. Building on this framework, combined with the model in Eq. \eqref{Sat_SIR}, we develop the MN-SIR epidemic model with a saturation incidence rate, incorporating graph Laplacian diffusion, as follows:  
\begin{equation}\label{model}
    \begin{aligned}
     \dfrac{\partial S (x, t)}{\partial t}   - \sigma\Delta S(x, t) &= \Lambda - \dfrac{\beta S(x, t) I(x, t)}{1 + \alpha I(x, t)} - \mu S(x, t),\quad  S(x ,0) = S_0(x) > 0 \mbox{ for } x \in V,   \\
  \dfrac{\partial I (x, t)}{\partial t} 
    - \sigma \Delta I(x, t) &= \dfrac{\beta S(x, t) I(x, t)}{1 + \alpha I(x, t)} - (\gamma + \mu)I(x, t), \quad I(x ,0) = I_0(x) \geq 0 \mbox{ for } x \in V,  \\
   \dfrac{\partial R (x, t)}{\partial t} - \sigma \Delta R(x, t) &= \gamma I(x, t) - \mu R(x, t), \quad  R(x ,0) = R_0(x) \geq 0 \mbox{ for } x \in V,
    \end{aligned}
\end{equation}
{\color{black} where $S_0(x), I_0(x),$ and $R_0(x)$ denote the initial values of the susceptible, infected, and recovered populations, respectively, at node $x \in V$. Model \eqref{model} captures the spatio-temporal dynamics of the susceptible ($S(x,t)$), infected ($I(x,t)$), and recovered ($R(x,t)$) populations across space ($x \in V$) and time ($t$). It combines local disease transmission processes with spatial diffusion, representing the movement or interaction of individuals across space. 
The partial derivatives $\frac{\partial S(x,t)}{\partial t}$, $\frac{\partial I(x,t)}{\partial t}$, and $\frac{\partial R(x,t)}{\partial t}$ represent the temporal rates of change for each compartment at location $x$. The diffusion terms $\sigma \Delta S(x,t)$, $\sigma \Delta I(x,t)$, and $\sigma \Delta R(x,t)$ describe the spatial movement of individuals, where $\sigma$ is the diffusion coefficient or migration parameter, ranging between 0 and 1. Adjusting $\sigma$ governs the rate of population mobility across locations, such that $\sigma \to 1$ indicates all regions exhibit nearly uniform dynamics, while $\sigma \to 0$ leads to decoupled, independent regional behavior \cite{bustamante2021epidemic}. The nonlinear transmission term $\frac{\beta S(x,t) I(x,t)}{1 + \alpha I(x,t)}$ introduces a saturation effect, reflecting the slowing growth of transmission when the infected population becomes large. This captures the realistic scenario where transmission does not increase indefinitely but saturates due to factors such as limited contact opportunities, improved healthcare access, increased public awareness, among many others.} 
% The parameter $\sigma$, which ranges between 0 and 1, characterizes the rate of population mobility. This parameter is often referred to as the migration parameter or diffusion parameter. By adjusting $\sigma$, one can modulate population mobility as required. Notably, when $\sigma$ approaches 1, all regions tend to exhibit similar dynamics, behaving almost uniformly. In contrast, when $\sigma$ is near 0, the solution profiles appear decoupled, with regions behaving independently \cite{bustamante2021epidemic}.

\subsection{Equilibrium points}
Disease-free equilibrium point is $(S_0, 0, 0)$, where $S_0 = \Lambda/\mu$ (notations have been discussed in Section \ref{Sec_Epi_model_Review}) and endemic equilibrium point is $(S^*, I^*, R^*)$, where
\begin{align}\label{steady_state}
    S^* = \dfrac{\Lambda \alpha + (\gamma + \mu)}{\alpha\mu + \beta}, \quad I^* = \dfrac{\mu(\mathcal{R}_0 - 1)}{\alpha\mu + \beta}, \quad R^* = \dfrac{\gamma(\mathcal{R}_0 - 1)}{\alpha\mu + \beta}. 
\end{align}
Thus, the endemic equilibrium exists if  $\mathcal{R}_0 > 1$, where the basic reproduction number  ($\mathcal{R}_0$) is given by $ \mathcal{R}_0 =\dfrac{\beta \Lambda}{\mu(\gamma+\mu)}$. From an epidemiological perspective, $\mathcal{R}_0$ is the expected number of secondary cases generated by a typical infected individual throughout his infectious interval in a fully susceptible population. Mathematically, the progression of a disease is typically characterized by $\mathcal{R}_0$, such that if $\mathcal{R}_0 < 1$, the disease-free equilibrium is stable, and the disease will eventually die out. Conversely, if $\mathcal{R}_0 > 1$, the disease-free equilibrium becomes unstable, and the system transitions towards the endemic equilibrium, indicating the sustained presence of the disease in the population. Therefore, $\mathcal{R}_0$ determines the threshold quantity for investigating the asymptotic stability of the equilibrium states and the prediction value needed for disease eradication. To determine the equilibria states and analyze the stability of these states, we introduce the following Lemma by \cite{tian2020global}, {\color{black} which will be utilized in the proof of stability analysis}:
\begin{lemma}[Green's Formula]\label{Green}
Let $V$ be finite and  any two functions $f, g : V  \to \mathbb{R}$, we have the following result
\begin{align*}
    \sum_{x \in V} f(x) \Delta g(x) = - \frac{1}{2} \sum_{x, y \in V} \left(f(y) - f(x)\right) \left( g(y) - g(x) \right).
\end{align*}
\end{lemma}
%\begin{proof}
\noindent The proof of the lemma is given in the Lemma 2.1 of \cite{tian2020global}.\\[3ex]
\noindent Assume $u(t) \geq 0, $ when $t \in [0, \infty)$, that satisfies the following equation
\begin{align}\label{x_eqn}
    \dot u = \dfrac{a u}{1+\alpha u} - c u, \quad \mbox{ for } t \in [0, \infty),
\end{align}
where $a, \alpha$, and $c$ are positive constants, and $\dot u$ represents the derivative of $u$ with respect to time. Also, note that Eq. \eqref{x_eqn} has a trivial equilibrium at $u_0 = 0$. If $a>c$, then the system in Eq. \eqref{x_eqn} has a unique positive non-trivial equilibrium, $u^* = \dfrac{a-c}{\alpha c}$.
\begin{lemma} \label{lim_glo}
If $a < c$, the trivial equilibrium, $u_0$, of Eq. \eqref{x_eqn} is globally asymptotically stable and if $a > c$, the non-trivial equilibrium, $u^* = \dfrac{a-c}{\alpha c}$, of the Eq. \eqref{x_eqn} is globally asymptotically stable.
\end{lemma}
\vspace{5pt}
\noindent The proof of the lemma is given in the Appendix \ref{proof_lim_glo}.\\
\begin{lemma} \label{supinf}
Consider $\sigma, \kappa,$ and $\nu$ are positive constants. Let for each  $x \in V$, $u(x, \cdot) \in \mathbb{C}([0, \infty))$ is differentiable in $(0, \infty)$, where $\mathbb{C}([0, \infty))$ is the space of continuous functions on $[0, \infty)$. If $u(x, t), \mbox{ where } (x, t) \in V \times (0, \infty),$ satisfies
\begin{equation}
\begin{aligned}
       &\dfrac{\partial u}{\partial t} - \sigma \Delta u \geq ( \leq )\,\, \kappa - \nu u,
\end{aligned}
\end{equation}
with the initial condition $u(x, 0) = u_{0}(x) \geq 0, x \in V$, then $  \liminf\limits_{t \to \infty} u(x, t) \geq \frac{\kappa}{\nu} \, \Big( \limsup\limits_{t \to \infty}u(x, t) \leq  \frac{\kappa}{\nu} \Big)$ uniformly in $x\in V$. Furthermore, for any given sufficiently small positive real number $\epsilon$, there exists a positive $t_\epsilon $, such that 
\begin{equation}\label{rel1}
    u(x, t) > \frac{\kappa}{\nu} - \epsilon\, \Big( u(x, t) < \frac{\kappa}{\nu} + \epsilon \Big), \, \mbox{ for } t \in (t_\epsilon, \infty).
\end{equation}
\end{lemma}
\vspace{5pt}
\noindent The proof of the lemma is given in Lemma 2.4 of  \cite{tian2020global}. 
{\color{black}  We present the above results to facilitate the proof of Lemma~\ref{frac_supinf}.}\\
\begin{lemma} \label{frac_supinf}
Consider  $\alpha,  \sigma, \kappa,$ and $\nu$ are positive constants. Let for each  $x \in V$, $u(x, \cdot) \in \mathbb{
C}([0, \infty))$ is differentiable in $(0, \infty)$. If $u(x, t), \mbox{ where } (x, t) \in V \times (0, \infty),$ satisfies
\begin{equation}
\begin{aligned}
       &\dfrac{\partial u}{\partial t} - \sigma \Delta u \geq ( \leq )\,\, \dfrac{\kappa u}{1+\alpha u}   - \nu u,
\end{aligned}
\end{equation}
with the initial condition $u(x, 0) = u_{0}(x) \geq 0, x \in V$, then 
$ \liminf\limits_{t \to \infty} u(x, t) \geq l\, \Big( \limsup\limits_{t \to \infty}u(x, t) \leq  l \Big)$
uniformly in $x\in V$. Furthermore, for any given sufficiently small positive real number $\epsilon$, there exists a positive $t_\epsilon $, such that 
$$
    u(x, t) > l - \epsilon\, \Big( u(x, t) < l + \epsilon \Big), \, \mbox{ for } t \in (t_\epsilon, \infty), 
$$
where $l = \begin{cases}
     \dfrac{\kappa - \nu}{\alpha \nu}, &\mbox{ if } \kappa > \nu,\\[0.75ex]
     0, &\mbox{ if } \kappa < \nu.
  \end{cases}$
\end{lemma}
\vspace{5pt}
\noindent The proof of the lemma is given in the Appendix \ref{proof_frac_supinf}.
\subsection{Stability Analysis}
Stability analysis plays a pivotal role in epidemic modeling by providing critical insights into the dynamics of disease spread and control. It helps determine whether a disease will die out or persist in a population. %In epidemiology, the progression of a disease is typically characterized by the basic reproduction number, $\mathcal{R}_0$.  If $\mathcal{R}_0 < 1$, the Disease-Free Equilibrium (DFE) is stable, and the disease will eventually die out. Conversely, if  $\mathcal{R}_0 > 1$,  the DFE becomes unstable, and the system transitions towards the Endemic Equilibrium (EE), indicating the sustained presence of the disease in the population. 
Stability analysis can be broadly classified into two types: local and global stability \cite{ma2009dynamical, martcheva2015introduction}. Local stability focuses on the behavior of the system in the vicinity of an equilibrium point. It determines whether small perturbations or deviations from this point will decay over time, allowing the system to return to equilibrium. Global stability, on the other hand, examines the system's behavior across its entire state space. It ensures that, regardless of the initial conditions, the system converges to a specific equilibrium point over time.  Given the complexities and high-dimensional nature of many epidemic models, local stability alone may not suffice to ensure the long-term eradication or persistence of disease. Hence, we focus on establishing global stability, which provides a more comprehensive characterization of the system's dynamics.\\[2ex]
\noindent 
To establish the global stability of an equilibrium point, we analyze the global dynamics of the disease-free equilibrium and the endemic equilibrium of the system in Eq. \eqref{model} using Green's formula and the Comparison principle. To simplify the analysis, we initially perform model dimension reduction. Since the variable  $R$ does not appear in the first two equations of the system, the last equation is omitted to prove the global dynamics of the system in Eq. \eqref{model}. Thus, we focus on the following subsystem:
\begin{subequations}\label{submodel}
    \begin{align}
    \dfrac{\partial S(x, t)}{\partial t} - \sigma\Delta S(x, t) &= \Lambda - \dfrac{\beta S(x, t) I(x, t)}{1 + \alpha I(x, t)} - \mu S(x, t),\quad S(x ,0) = S_0(x) > 0 \mbox{ for } x \in V,    \label{submodel_a}  \\
     \dfrac{\partial I(x, t)}{\partial t} - \sigma \Delta I(x, t) &= \dfrac{\beta S(x, t) I(x, t)}{1 + \alpha I(x, t)} - (\gamma + \mu)I(x, t), \quad I(x ,0) = I_0(x) \geq 0 \mbox{ for } x \in V.   \label{submodel_b} 
    \end{align}
\end{subequations}
{\color{black}For the model to be epidemiologically meaningful, we show the positivity and boundedness of solutions for all time $t$, since the population sizes cannot be negative or unbounded.}
\begin{lemma}\label{positive1}
The region $\mathbb{D}_+ = \big\lbrace (S, I)\,|\, S \geq 0, I\geq 0 \mbox{ for } x \in \textbf{}V  \big\rbrace$ is a positive invariant set for the model in Eq. \eqref{submodel}. 
\end{lemma}
\noindent The proof of the lemma is given in the Appendix \ref{proof_positive1}.

\begin{lemma}\label{bdd}
If $\left(S(x, t), I(x, t)\right)$ satisfies Lemma \ref{positive1} for all $(x, t) \in V \times (0, \infty)$, then there is a positive real number $K$ such that 
$0 \leq S(x, t), I(x, t) \leq K, \quad \mbox{for all } (x, t) \in V \times (0, \infty)$.

\end{lemma}
\noindent The proof of the lemma is given in Appendix \ref{proof_bdd}.

\vspace{0.2cm}

\noindent The disease-free equilibrium point denotes the point at which there are no infections in the population. On contrary, the endemic equilibrium state denotes the state where the disease cannot be completely eradicated but persists in the population. To determine the stability of the disease-free equilibrium state and endemic equilibrium state, we have the following theorems:
\begin{theorem}\label{dfe_glo}
The disease-free equilibrium $(S_0, 0)$  of the model Eq. \eqref{submodel} is globally asymptotically stable when the basic reproduction number is less than 1 ($\mathcal{R}_0 < 1$).
 \end{theorem}
 \noindent The proof is given in Appendix \ref{proof_dfe_glo}.
 \begin{theorem}\label{ee_glo}
The endemic equilibrium $(S^*, I^*)$  of the model Eq. \eqref{submodel} is globally asymptotically stable when the basic reproduction number is greater than 1 ($\mathcal{R}_0 > 1$).
 \end{theorem}
\noindent The proof is given in Appendix \ref{proof_ee_glo}.
 
\vspace{0.2cm}

\noindent The global stability analysis ensures that, regardless of initial conditions, the disease will ultimately converge to one of the equilibrium states, such as the disease-free equilibrium or endemic equilibrium. This concept is central to understanding how epidemic models predict the long-term trajectory of disease spread under different intervention strategies. By examining these equilibria, we can assess the effectiveness of interventions in achieving disease control. Building on these epidemic insights, we integrate the epidemiological principles of the MN-SIR model into deep learning techniques to generate reliable epidemic forecasts.

\section{Epidemic-Guided Deep Learning Framework}\label{Sec_EGDL_Model}
\begin{figure}[!ht]
    \centering
    \includegraphics[width=0.7\textwidth]{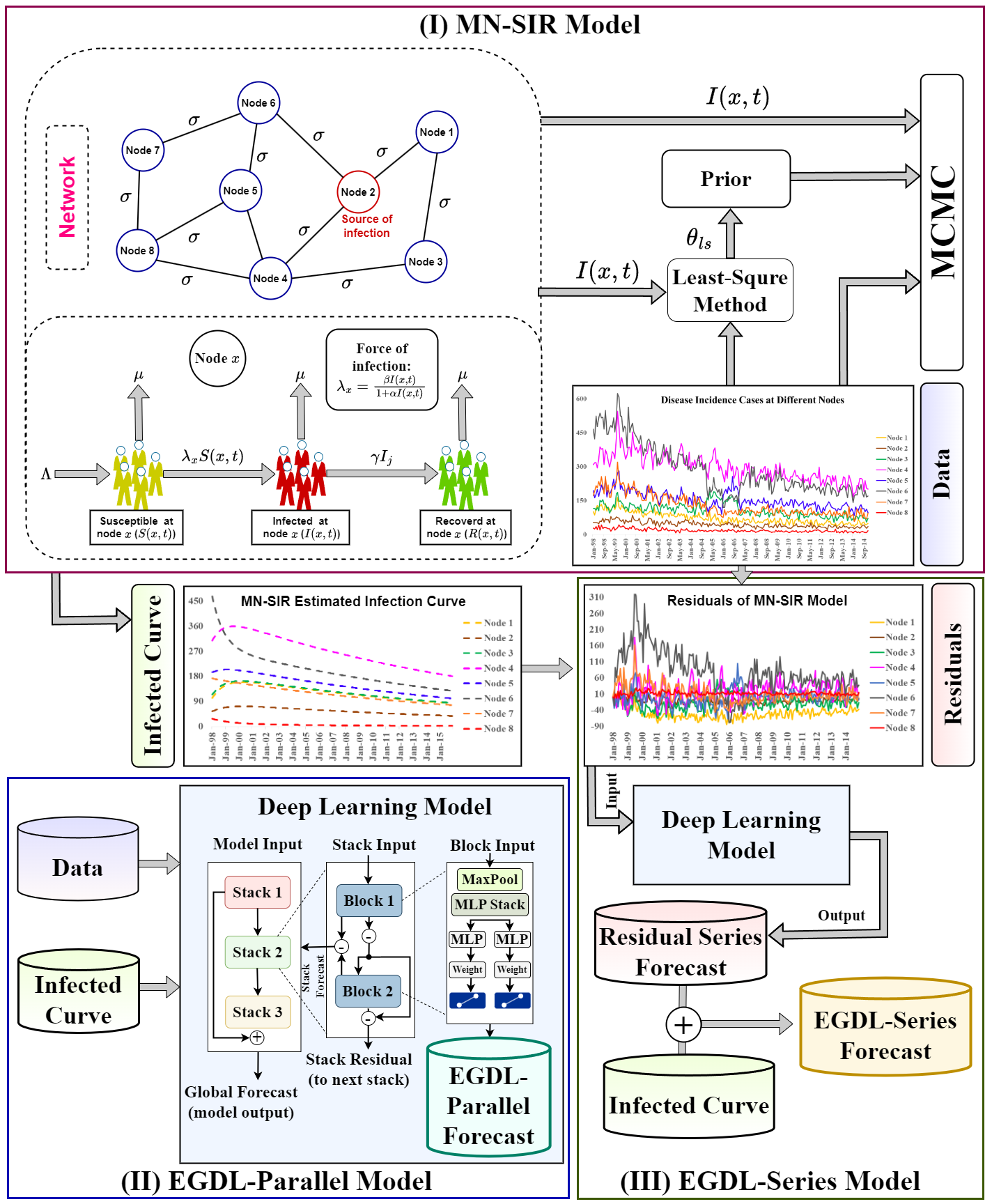}
    \caption{\textbf{Epidemic-Guided Deep Learning (EGDL).} The upper panel of the plot represents a schematic architecture of the modified networked SIR (MN-SIR) model, and the lower panel showcases the workflow of the EGDL-Parallel framework (left) and the EGDL-Series approach (right). The demonstration is given using Japan's TB dataset.}
    \label{Fig_NSIR_EGDL}
\end{figure}

This section outlines the generic architecture of the epidemic-guided deep learning (EGDL) methods, which seamlessly integrate epidemiological principles with spatial interactions modeled through the MN-SIR framework (described in Section \ref{Network_SIR_Model}) with data-driven techniques to forecast the future trajectories of the disease incidence across different geographical locations. Specifically, we introduce two distinct approaches, namely EGDL-Parallel and EGDL-Series, which integrate the scientific knowledge of the disease dynamics and its spatial distribution into deep learning models to forecast TB incidence cases. The parameters of the MN-SIR model, including transmission, recovery, saturation, population mobility, and birth/death rates, guide the deep learning architectures to ensure that the TB incidence forecasts align with real-world epidemiological trends and can effectively capture both the spatial and temporal dependencies. {\color{black}Fig. \ref{Fig_NSIR_EGDL} illustrates the schematic architecture of the EGDL frameworks. A detailed pseudo-code for the proposed approaches is presented in Algorithm \ref{algo_EGDL_Parallel} of Appendix \ref{Appendix_EGDL_Algo}.}

% (a detailed description of these methods is given in Section \ref{Sec_Phenom_Review})
% $$f_{M}: \left[Y(1,t), \ldots ,Y(n,t)\right] \longrightarrow \widehat{Y}(1,{T+v}), \ldots, \widehat{Y}(n,{T+v}); \; t = 1, 2, \ldots, T; \; v = 1, 2, \ldots, q.$$ 

\subsection{EGDL-Parallel Method}
We consider the problem of forecasting epidemic incidence cases based on $T$ historical observations recorded across $n$ vulnerable locations. {\color{black}Following the notations introduced in Section \ref{Prelim_Spatiotemporal_Forecasting}, the objective is to iteratively generate the $q$-step-ahead forecasts of $\{Y(x, t)\}_{t \leq T}$ as $\{\widehat{Y}(x, T + v)\}_{v=1}^q$ for each location $x \in V$. A standard approach is to use a global forecasting model $f_{M} \in \mathcal{F}$, as described in Eq. \eqref{Eq_Fore_Data}, which uses past incidence data to generate future forecasts.} In this study, we choose $\mathcal{F}$ to represent two broad categories of data-driven deep learning architectures: stacked-based frameworks such as NBeats and NHits, and encoder-decoder-based models such as TCN and Transformers. % We denote the number of infected individuals observed at time $t$ in location $x$ as $Y(x,t); \; t = 1, 2, \ldots, T,\; x = 1, 2, \ldots, n$. Our goal is to iteratively generate the $q \; (\geq 1)$ step-ahead forecasts of $Y(x,t)$, to be denoted as $\widehat{Y}(x,{T+1}), \widehat{Y}(x,{T+2}), \ldots, \widehat{Y}(x,{T+q})$. A standard approach to this problem is to use a forecasting model $f_{M} \in \mathcal{F}$, where $\mathcal{F}$ represents the set of possible data-driven global forecasting models. In this study, we choose $\mathcal{F}$ to represent two types of data-driven deep learning architectures: stacked-based frameworks, such as NBeats and NHits, and encoder-decoder-based architectures such as TCN and Transformers. Thus using the historical data $\{Y(1,t), Y(2,t), \ldots Y(n,t); t = 1, 2, \ldots, T\}$ for training $f_{M}$ the future dynamics can be predicted as follows: 
% \begin{equation}\label{Eq_Fore_Data}
%     \widehat{Y}\left(x, \tilde{t} + v\right) = f_{M}\left(\left\{Y\left(x, \tilde{t} - t_0 \right)\right\}_{t_0 = 0}^{t_w - 1}\right),
% \end{equation}
% where $x = 1, 2, \ldots, n, \; \tilde{t} = t_w, t_w+1, \ldots, T, \; v = 1, 2, \ldots, q$, and $t_w$ denotes the number of historical inputs used for predicting the subsequent observations. 
While these models are effective at capturing temporal dependencies and complex nonlinear patterns, they solely rely on the lagged observations of the incidence time series without incorporating the underlying epidemiological mechanisms governing disease spread. As a result, this approach leads to a simple extrapolation of the temporal trends, which may fail to capture the underlying epidemiological underpinnings of the disease. To address this limitation, the EGDL-Parallel approach leverages the MN-SIR model to estimate the infection dynamics curve for the $n$ locations based on various disease drivers such as transmission, recovery, saturation, population mobility, and population death rates. The number of infection cases $I(x,t)$ predicted at timestamp $t$ for location $x$ using the MN-SIR model's infection curve captures the epidemic knowledge and models spatial dependencies among the neighboring locations using the graph Laplacian diffusion. However, these estimates provide an incomplete representation of the target variable $Y(x,t)$ due to their inability to track the lagged temporal dependencies, leading to discrepancies with ground truth observations. 

The EGDL-Parallel approach overcomes this issue by integrating the MN-SIR infection estimates $I(x,t)$ as auxiliary variables (exogenous information) into the data-driven forecasting framework. This hybrid input allows the forecasting model to combine the strengths of both compartmental methods and data-driven techniques to forecast incidence cases. The resulting EGDL-Parallel framework, denoted as $f_{EGDLP}$, is defined as:
% $$ %  $\{Y(1,t), Y(2,t), \ldots, Y(n,t), I(1,t), I(2,t), \ldots, I(n,t); \; t = 1, 2, \ldots, T\}$,
%     f_{EGDLP}: \left[Y(1,t), \ldots, Y(n,t), I(1,t), \ldots, I(n,t)\right] \longrightarrow \widehat{Y}(1,{T+v}), \ldots, \widehat{Y}(n,{T+v}); \; 
% $$
$$
    \widehat{Y}\left(x, \tilde{t}+v\right) = f_{EGDLP}\left(\left\{Y\left(x, \tilde{t} - t_0 \right)\right\}_{t_0 = 0}^{t_w - 1}, \; \left\{I\left(x, \tilde{t} - t_0 \right)\right\}_{t_0 = 0}^{t_w - 1}\right),
$$
where $f_{EGDLP} \in \mathcal{F}$ and notations are consistent with Eq. \eqref{Eq_Fore_Data}.
% where $t = 1, \ldots, T; \; v = 1, \ldots, q,$ and $f_{EGDLP} \in \mathcal{F}$. 
The primary advantage of the $f_{EGDLP}$ model over its data-centric counterpart $f_{M}$ is its ability to integrate disease-specific knowledge, enhancing its generalizability and forecast accuracy. By leveraging the MN-SIR model, the $f_{EGDLP}$ framework is guided by epidemiological principles, enabling the generation of realistic future estimates of disease incidence cases. Moreover, it effectively learns the complex spatiotemporal features from the compartmental models and data-driven methods, making it a robust solution for epidemic forecasting. 

\subsection{EGDL-Series Method}
The EGDL-Series framework employs an integrated deterministic-stochastic approach to model and forecast disease incidence cases by combining epidemiological principles with data-driven techniques. % through a residual remodeling strategy, this hybrid methodology captures the overall variations in disease dynamics. Specifically, the framework leverages the strengths of the epidemic networked SIR model to estimate deterministic patterns while utilizing advanced data-driven forecasting techniques to address stochastic variations. This synergy enables EGDL-Series architectures to leverage the strengths of both paradigms and mitigate their respective limitations. The flexibility of this hybrid framework allows it to better adapt to real-world complexities and dynamic disease behaviors, making it a powerful tool for forecasting infectious disease spread. 
{\color{black}In the EGDL-Series method, a deterministic model (MN-SIR in this case) is combined with stochastic models from set $\mathcal{F}$ using a residual remodeling approach \cite{zhang2003time, chakraborty2019forecasting}.} Mathematically, the disease incidence cases recorded at time $t$ for a location $x$ can be expressed as:
$ Y(x, t) = \mathcal{D}(x, t) + \mathcal{\tilde{S}}(x, t),$ 
where $\mathcal{D}(x, t)$ represent the deterministic component and $\mathcal{\tilde{S}}(x, t)$ accounts for the stochastic variations in $Y(x, t)$. To forecast future disease trajectories, the EGDL-Series framework first utilizes the MN-SIR model to estimate the infection dynamics $I(x,t)$. This curve, determined based on key epidemiological factors, such as disease transmission, recovery, saturation, population mobility, and birth/death rates, approximates the deterministic components $\mathcal{D}(x, t)$ of $Y(x, t)$. The predicted infection trajectory, $\widehat{\mathcal{D}}(x, t) \equiv I(x,t)$, captures structural variations in the disease incidence by modeling epidemic principles and spatial interactions among neighboring locations. The deviations between observed incidences and the predictions of the MN-SIR model, defined as residuals $e(x,t) = Y(x,t) - I(x,t)$, capture the stochastic variations that the deterministic part fails to account for. These residuals store nonlinearities, temporal dependencies, and unexplained stochastic dynamics that the MN-SIR framework cannot model. The residual series is modeled using data-driven forecasting techniques, $f_{M} \in \mathcal{F}$. The $q$-step-ahead forecasts of the residual series generated by the deep learning-based temporal forecasters are combined with the predictions of the MN-SIR model to forecast the disease incidence. The predictions of the EGDL-Series methods can be mathematically represented as:
\begin{equation*}
    \widehat{Y}(x, \tilde{t}+v) = \hat{f}_{EGDLS}(x, \tilde{t}+v) = I(x, \tilde{t}+v) + \hat{e}(x, \tilde{t}+v), 
\end{equation*}
with 
$
    \hat{e}(x, \tilde{t}+v) = f_{M}\left(\left\{e\left(x, \tilde{t} - t_0 \right)\right\}_{t_0 = 0}^{t_w - 1}\right), f_{M} \in \mathcal{F}, 
$
% $$
% f_{EGDLS}: \left[e(1, t), e(2,t), \ldots, e(n,t)\right] \longrightarrow \hat{e}(1,T+v),  \hat{e}(2,T+v), \ldots, \hat{e}(n,T+v);
% $$
% $$
% \widehat{Y}(x, T+v) = I(x, T+v) + \hat{e}(x, T+v);
% $$
% where $\hat{e}(x, T+v)$ is generated using $f_{EGDLS}$ as 
following the same notations as in Eq. \eqref{Eq_Fore_Data}. The proposed formulation ensures that the model aligns with the epidemic principles coupled with spatial interactions while simultaneously addressing the unexplained variations of the compartmental model. The EGDL-Series approach offers enhanced flexibility, greater insight into disease dynamics, and improved forecasting accuracy by integrating compartmental and data-driven components. 

\section{Application to TB study}\label{Sec_forecast}
% In this section, we assess the efficiency of the EGDL-Parallel and EGDL-Series frameworks by comparing their TB incidence forecasting performance across Japan's 47 prefectures with state-of-the-art architectures. To evaluate the robustness and generalizability of the EGDL-based approaches, we employ rolling window forecast horizons of 12 months, 9 months, 6 months, and 3 months, performing multi-step ahead forecasts for each respective horizon. For the 12-month evaluation period, the models are trained using incidence data recorded from January 1998 to December 2014, while the remaining observations are reserved for testing. Similarly, for the 9-month, 6-month, and 3-month horizons, training data spans from January 1998 to March 2015, June 2015, and September 2015, respectively, with the remaining data used for evaluation. Since EGDL is an integrated approach of the networked SIR model with a saturated incidence rate and modern deep learning-based time series forecasting methods, therefore, we first delve into the parameter estimation procedure for the former parts. 

{\color{black}In this section, we assess the efficiency of the EGDL-Parallel and EGDL-Series frameworks across two geographically and epidemiologically distinct settings, Japan and China. For both countries, we compare our models against state-of-the-art forecasting architectures using rolling windows over 12, 9, 6, and 3-month horizons for evaluating the accuracy and generalizability of the EGDL frameworks across diverse temporal contexts and population structures.} For Japan, the experiments are conducted across all 47 prefectures. The models are trained using monthly TB incidence data from January 1998 to December 2014 for the 12-month horizon, while the subsequent months are used for testing. Similarly, for the 9-month, 6-month, and 3-month horizons, the training periods extend until March 2015, June 2015, and September 2015, respectively, with the remaining data reserved for evaluation. {\color{black}In the case of China, we retain a consistent experimental setup to ensure a fair comparison. The models are trained on monthly TB incidence data of 31 provinces from January 2014 to December 2017 for the 12-month horizon, with the remaining months of 2018 used for testing. For the 9-month and 6-month forecasts, training extends up to March 2018 and June 2018, respectively. For the 3-month horizon, data up to September 2018 is used for training, while the final quarter of 2018 is held out for testing.} Since EGDL integrates MN-SIR model with deep learning-based forecasting techniques, we begin by detailing the data characteristics and parameter estimation process of the mechanistic component before analyzing the overall forecasting performance across both countries.

%The following subsections provide a detailed analysis of the global characteristics of the TB incidence data (Section \ref{Sec_Global_Char_Sec}), estimation of the posterior distribution of the networked SIR model parameters (Section \ref{Bayesian_Inf_Sec}), the causal relationship between observed TB incidence cases and the estimated infected curve of the networked SIR model (Section \ref{Sec_Causal_Analysis}), performance indicators used for evaluation (Section \ref{Sec_Performance_measures}), comprehensive summary of the experimental results (Section \ref{Sec_Fore_Results}), statistical significance of the experimental results (Section \ref{Sec_Stat_SIgnif}), and uncertainty quantification of the proposed EGDL approaches (Section \ref{Sec_Uncertain_Quant}).

\subsection{TB Data Characteristics of Japan and China}\label{Sec_Global_Char_Sec}
\begin{figure}[H]
\centering
% \resizebox{0.9\textwidth}{!}{
\begin{tabular}[!ht]{cc}
(A) Laplacian Matrix   & (B)  Correlation Plot \\
  \includegraphics[width=45mm,height=40mm]{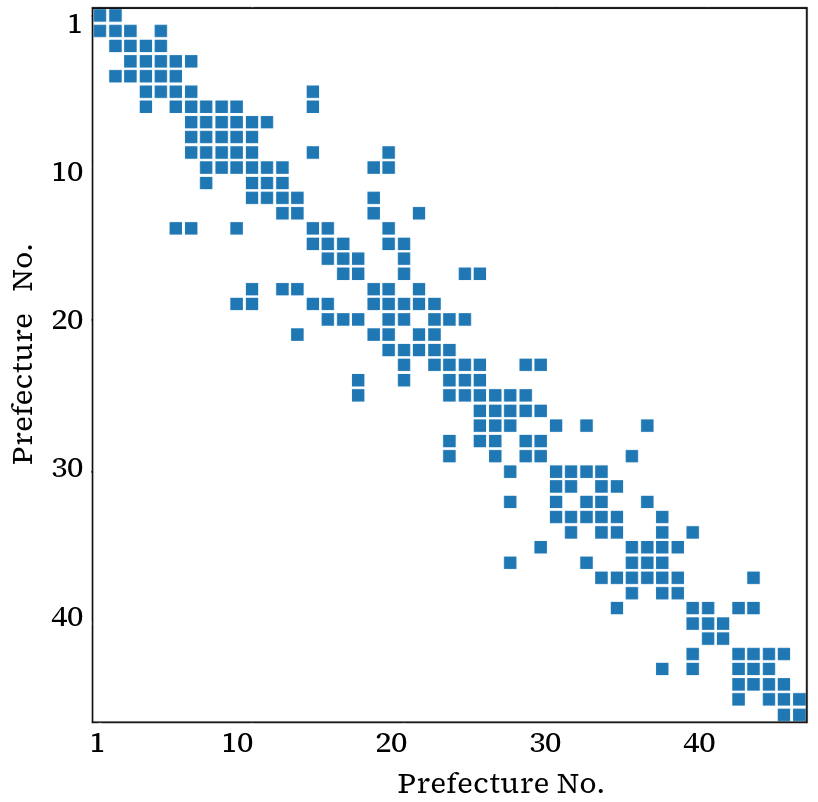} &   
  \includegraphics[width=45mm,height=40mm]{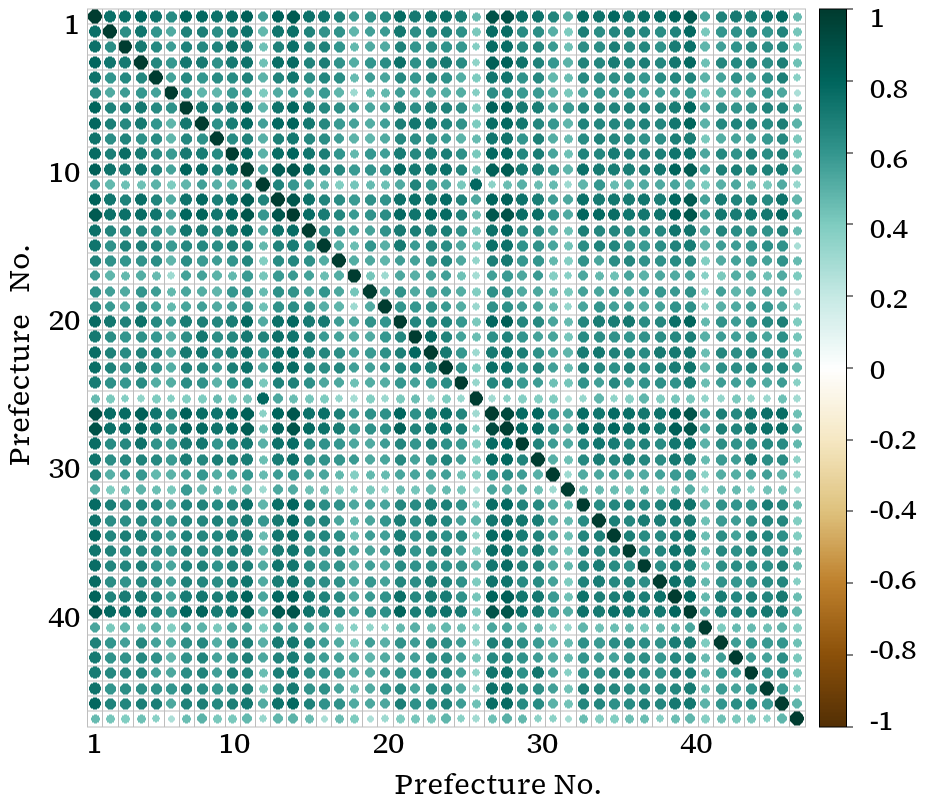} \\
% (c) Time evolution for $S$    & (d) Time evolution for $I$ \\
% \includegraphics[width=85mm,height=65mm]{Endemic_S.png} &   \includegraphics[width=85mm,height=65mm]{Endemic_I.png}   
\end{tabular}
\caption{(A) The Laplacian matrix corresponding to the network depicted in Fig. \ref{fig:TB47}(A), representing the connectivity and structure of the nodes in the system. This matrix captures the relationships between neighboring nodes and is essential for analyzing diffusion processes and other network dynamics. (B) Pairwise correlation of TB incidence cases recorded at 47 prefectures of Japan.}
\label{Fig_Correlation_LAP47}
\end{figure}

\begin{figure}
    \centering
    \includegraphics[width=0.7\linewidth]{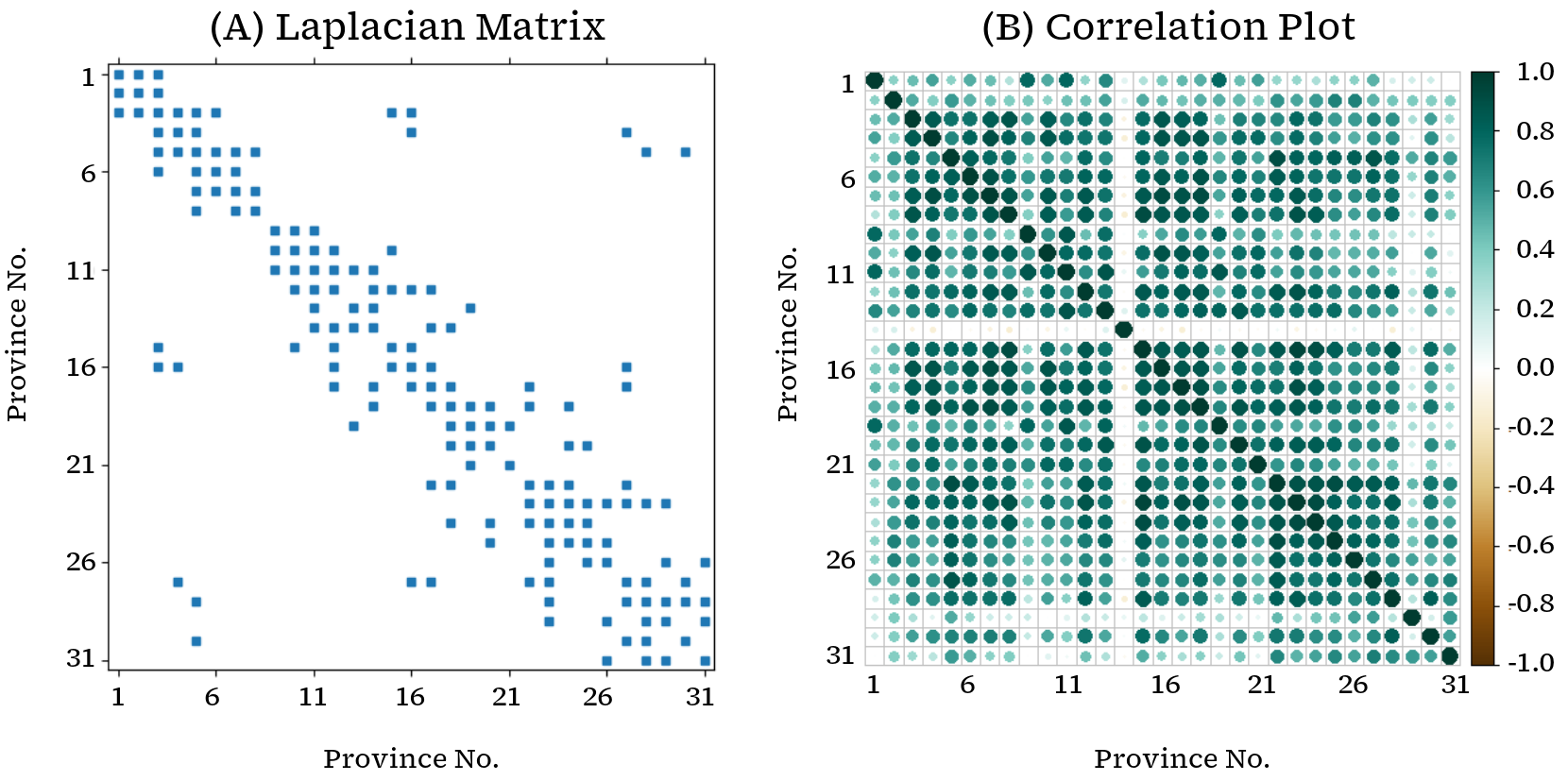}
    \caption{{\color{black}(A) The Laplacian matrix corresponding to the network depicted in Fig. \ref{Fig_China_MAP}(A), representing the connectivity and structure of the nodes in the system. This matrix captures the relationships between neighboring provinces in mainland China and is essential for analyzing diffusion processes and other network dynamics. (B) Pairwise correlation of TB incidence cases recorded at 31 provinces of mainland China.}}
    \label{Fig_Correlation_LAP31_China}
\end{figure}

{\color{black}The TB incidence datasets for Japan and China, though collected from distinct geographical and epidemiological settings, both exhibit strong spatiotemporal dependencies that are crucial for accurate forecasting.} In Japan, monthly TB case counts across 47 prefectures are devoid of missing observations and exhibit rapid variations, with average monthly incidences ranging from 9.49 to 294.84. We analyzed several global characteristics for spatiotemporal TB datasets. The results of the statistical analysis, reported in Table \ref{Table_Global_Features} of Appendix \ref{Appendix_Global_Char}, reveal that the TB incidence time series for all 47 prefectures of Japan are non-stationary and exhibit long-term dependency, as indicated by Hurst exponents greater than 0.50. Additionally, most series display non-linear patterns with quarterly and annual seasonality. Furthermore, we explore the spatial relationships among prefectures by analyzing pairwise correlations in TB incidence. Fig. \ref{Fig_Correlation_LAP47}(B) presents a heatmap of these correlations, showing stronger associations between geographically proximate prefectures, for instance prefecture 10 (Gunma) and its neighboring regions, while isolated prefectures like 47 (Okinawa) exhibit weaker correlations. Diagonal entries in the heatmap represent self-correlations, which are equal to 1. {\color{black}A similar spatiotemporal structure is also observed in mainland China’s TB incidence cases. This dataset, free from missing observations, also exhibits considerable regional variations, with the average monthly cases ranging from 243.67 to 4,877.45. Statistical analysis of mainland China’s TB incidence datasets, summarized in Table \ref{Table_Features_CHina} of Appendix \ref{Appendix_Global_Char}, reveals that all series exhibit long-term memory, are predominantly non-stationary, follow approximately linear trends, and demonstrate annual seasonality. Correlation heatmaps, Fig. \ref{Fig_Correlation_LAP31_China}(B), indicate that geographically adjacent provinces, such as 3 (Hebei), 4 (Shanxi), and 5 (Inner Mongolia), tend to have stronger correlations, while remote provinces such as 1 (Beijing) and 26 (Tibet) show weaker associations. Notably, both TB incidence data from Japan and China reveal moderate to strong correlations even between non-adjacent regions, likely due to shared non-local drivers. These findings collectively validate the presence of both spatial and temporal dependencies in TB incidence across Japan and China, emphasizing the importance of jointly modeling these dynamics to improve forecasting accuracy.}

\subsection{Results on Japan's Prefecture-specific Active TB Cases}
In this section, we conduct an in-depth analysis of the EGDL frameworks for forecasting monthly active TB cases across 47 prefectures of Japan. We begin by estimating the key epidemiological parameters that govern infection dynamics within the MN-SIR model, followed by a comprehensive evaluation of the forecasting performance of the proposed frameworks. 

\subsubsection{Parameter Calibration}\label{numerical}
\noindent To numerically solve the MN-SIR model, as in Eq. \eqref{submodel}, we transform it into a system of ODEs \eqref{num_model} using the definition \eqref{hfun}. This process requires the Laplacian matrix used in our study to solve the model numerically. The Laplacian matrix \( L \) is defined by $ L = D - A,$
where \( D \) is the degree matrix and \( A \) is the adjacency matrix. The degree matrix \( D \) is a diagonal matrix with the entries \( D_{ii} \), the degree of node \( i \), representing the number of edges incident to node \( i \), i.e., 
\(D_{ii} = \text{deg}(v_i)\). The adjacency matrix \( A \) with entries \( A_{ij} \) indicates the link between nodes \( i \) and \( j \) as follows:
\begin{equation}\label{eq_Aij}
   A_{ij} =
  \begin{cases}
    1, & \text{if there is a link between nodes } i \text{ and } j \\
    0, & \text{otherwise.}
  \end{cases} 
\end{equation}
Thus, the diagonal elements of \( L \) 
%in Eq. \eqref{eq_L} 
are the degrees of the nodes, while the off-diagonal elements \( L_{ij} \) are 1 if nodes \( i \) and \( j \) are connected by an edge, and 0 otherwise:
\begin{equation}\label{eq_Lij}
    L_{ij} =
  \begin{cases}
    -\text{deg}(v_i), & \text{if } i = j \\
    1, & \text{if } i \neq j \text{ and } (i, j) \in \mathcal{E} \\
    0, & \text{if } i \neq j \text{ and } (i, j) \notin \mathcal{E}.
  \end{cases}
\end{equation}
To construct the Laplacian matrix across different prefectures of Japan, we assume a connection between two prefectures if they share a common boundary. For the prefectures of Hokkaido, Tokushima, Kagawa, Ehime, and Okinawa, which are geographically separated from the mainland (see Fig. \ref{fig:TB47}(A)), we establish connections to the nearest mainland prefecture. The structure of these connections is illustrated in Fig. \ref{Fig_Correlation_LAP47}(A). Now, for each node $k \in V$, we rewrite the model in Eq. \eqref{submodel} as follows:
\begin{equation}\label{num_model}
    \begin{aligned}
            \dfrac{d S_k}{d t} =  \sigma\sum_{j=1}^n {L}_{k\,j}S_j + \Lambda - \dfrac{\beta S_k I_k}{1 + \alpha I_k} - \mu S_k, \quad\, S(k, t) = S_k(t), S(k, 0) = S_{k}(0) \mbox{ for } k \in V,\\ 
    \dfrac{d I_k}{d t} =  \sigma \sum_{j=1}^n {L}_{k\,j}I_j + \dfrac{\beta S_k I_k}{1 + \alpha I_k} - (\gamma + \mu)I_k,\quad  I(k, t) = I_{k}(t), I(k, 0) = I_{k}(0) \mbox{ for } k \in V.
\end{aligned}
\end{equation}

\noindent Using the Laplacian matrix (see Fig. \ref{Fig_Correlation_LAP47}(A)), we numerically solve the system Eq. \eqref{num_model}, assuming that prefecture 20 (Nagano) serves as the source of infection. Additionally, to support theoretical findings, we consider an initial total population of 47,000 distributed evenly across the 47 prefectures, with each prefecture having a population of 1,000. The initial conditions are defined as follows:
\begin{equation}\label{init_p}
\left\lbrace
    \begin{aligned}
    I_{20}(0) & = 10.0, \quad I_1(0) = \cdots = I_{19}(0) = I_{21}(0) = \cdots = I_{47}(0) = 0.0\\
        S_{20}(0) &= 990.0, \quad   S_1(0) = \cdots = S_{19}(0) = S_{21}(0) = \cdots = S_{47}(0) = 1000.0.
    \end{aligned}
    \right.
\end{equation}
Further, we set the model parameters as follows: 
\begin{equation}\label{dis-para}
\left\lbrace
   \begin{aligned}
    \Lambda &= 10.0 \mbox{ day}^{-1},\, \mu = 0.01 \mbox{ day}^{-1},\, \beta = 0.0001\mbox{ day}^{-1},\\
    \gamma &= 0.25\mbox{ day}^{-1},\,\alpha = 0.5,\, \sigma = 0.75. 
\end{aligned} 
\right.
\end{equation}
Now, we calculate the basic reproduction number, $\mathcal{R}_0  = 0.385$, that is less than unity, according to Theorem \ref{dfe_glo} and the disease-free equilibrium $(\Lambda/\mu, 0) = (1000.0,0)$ point is globally asymptotically stable (see  Fig \ref{fig:stab}(a-b) of Appendix \ref{App_Param_Calli}). In Fig \ref{fig:stab}(a), all $S$ solutions converge to $\Lambda/\mu$ regardless of the node, similarly, in Fig \ref{fig:stab}(b), all  $I$ solutions converge to $0$ regardless of the node.  These results can be similarly demonstrated, independent of the initial conditions. For the endemic equilibrium, we use all the parameters from Eq. \eqref{dis-para} except $\beta$ and $\sigma$ which are set to $\beta = 0.001$ and $\sigma = 10^{-5}$. Direct calculation gives: $\mathcal{R}_0 = 3.85, S^* = 876.67, \text{ and } I^* = 4.74$. Since $\mathcal{R}_0  = 3.85$ is greater than unity, then by Theorem \ref{ee_glo} endemic equilibrium $(S^*, I^*) = (876.67, 7.74)$  point is globally asymptotically stable as shown in Fig \ref{fig:stab}(c-d). In Fig \ref{fig:stab}(c), all $S$ solutions converge to $S^*$ regardless of the node; similarly, in Fig \ref{fig:stab}(d), all $I$ solutions converge to $I^*$ regardless of the node. The saturation parameter \(\alpha\) significantly influences the solution of the system, as illustrated in Fig. \ref{fig:alpha_var} of Appendix \ref{App_Param_Calli}. For this analysis, we use all the parameters associated with the endemic equilibrium case, except for \(\alpha\), which is varied between 0.1 and 1.0. At lower values of \(\alpha\), the contact rate increases, resulting in a higher peak of infections. As \(\alpha\) approaches 1.0, the contact rate saturates, and the infection profile becomes more uniform over time.

\subsubsection{Sensitivity Analsyis}\label{Bayesian_Inf_Sec}
In this part, we estimate the posterior distributions of the MN-SIR model parameters using the Markov-Chain Monte-Carlo (MCMC) approach on the Laplacian network (depicted in Fig. \ref{Fig_Correlation_LAP47}(A)) via the \texttt{PyMC} Python library \cite{abril2023pymc}. To get posterior distributions of the unknown parameters, the state variables $\hat S_i(t)$ and $\hat I_i(t)$ are observed at $T$ certain times $(t_1, t_2,\dots,t_T)$.
%  such that
% \begin{align*}
%     \hat S_i(t_j) &= S_i(t_j) + \epsilon_j^{S_i},\\
%     \hat I_i(t_j) &= I_i(t_j) + \epsilon_j^{I_i}, \quad \mbox{ where } j = 1,2,\dots, T;\,\, i=1,2,\dots,M,
% \end{align*}
% and $\epsilon_j^{S_i}$ and $\epsilon_j^{I_i}$ are noises related to  susceptible individual and i-th infected individual at node i.
Let, $\theta = \lbrace \alpha, \beta, \gamma, \sigma, \mu \rbrace$ be the unknown parameter and $D_i = \lbrace \hat S_i, \hat I_i \rbrace$ be observed data at node $i$.
% then by Bayes' theorem:
% \begin{align}\label{eq_post}
%   \textbf{P}(\theta\,|\, D_i)  = \dfrac{\textbf{P}(D_i|\,\theta\ ) \textbf{P}(\theta)}{\textbf{P}(D_i)} \propto\textbf{P}(D_i|\,\theta) \textbf{P}(\theta) \,\, \mbox{ for each node }i, 
% \end{align}
% where $\textbf{P}(\theta\,|\, D_i)$ posterior distribution of $\theta$ given observed data, $\textbf{P}(D_i|\,\theta)$ is the likelihood of observing data at $i^{th}$ node given $\theta$, and $\textbf{P}(\theta)$ describes the prior distribution of $\theta$. %and $\textbf{P}(D_i)$ is the prior predictive distribution equation which is also the normalizing constant of the posterior distribution for node $i$ such that 
% \begin{equation}\label{eq_post_data}
%     \textbf{P}(D_i) = \int\limits_{\mathcal{G}_i} \textbf{P}(D_i\,|\, \theta) \textbf{P}(\theta)\, d\theta,
% \end{equation}
%where $\mathcal{G}_i$ is the domain of the parameter $\theta$. 
It is not always possible to find sample $\theta$ from the posterior probability distribution $\textbf{P}(\theta\,|\, D_i)$. This is where the iterative MCMC algorithm plays a crucial role. It generates a new vector parameter at step $\hat{n}$ ($\theta^{\hat{n}}$), from the posterior distribution, using the previous vector parameter ($\theta^{\hat{n} -1}$), with initial guess $\theta^0$. The chain or sample-path run $\hat{n}$ steps until it reaches its stationary distribution, where $\hat{n}$ is a sufficiently large number, and fills the space of the target un-normalized posterior distribution. % For implementing Bayesian inference with MCMC, we used the No-U-Turn Sampler (NUTS) \cite{hoffman2014no}. {\color{black} In this study, the NUTS sampler was configured with 2 chains, 3,000 tuning (warm-up) and  3,000 sampling iterations  per chain to ensure convergence to a stationary distribution and to obtain an efficient parameter estimates. On a single core of an Intel i7 9th generation processor, each chain took approximately 2.5 minutes to complete. Although the iterative MCMC procedure is computationally intensive, this runtime indicates that the method is feasible for moderate-scale applications on standard hardware. For larger datasets or real-time deployment, computational efficiency could be improved through parallel processing, hardware acceleration, or approximate inference techniques.} The NUTS approach automatically selects these parameters, making it a tuning-free sampling algorithm that performs similar or better than other MCMC algorithms \cite{hoffman2014no}. Before performing inference, we estimate the parameter set $\theta$, denoted as $\theta_{ls} \equiv \lbrace \alpha_{ls}, \beta_{ls}, \gamma_{ls}, \sigma_{ls}, \mu_{ls} \rbrace$, using the ordinary least squares (OLS) method with TB dataset from 47 prefectures in Japan. Based on these estimates, we define the prior distribution information for the parameters as follows:
For implementing Bayesian inference with MCMC, we utilized the No-U-Turn Sampler (NUTS) that adaptively tunes its parameters and performs competitively with other sampling methods. \cite{hoffman2014no}. {\color{black} In this study, the NUTS sampler was configured with two chains, each comprising 3,000 tuning (warm-up) and  3,000 sampling iterations, ensuring convergence to a stationary distribution and accurate parameter estimates. On a single core of an Intel i7 9th generation processor, each chain completed in approximately 2.5 minutes, demonstrating the feasibility of our approach for moderate-scale applications on standard hardware. For larger datasets or real-time deployment, this process could be accelerated using parallel computing, hardware acceleration, or approximate inference techniques.} 

To initialize the Bayesian inference process, we first estimate the parameter set $\theta$, denoted as $\theta_{ls} \equiv \lbrace \alpha_{ls}, \beta_{ls}, \gamma_{ls}, \sigma_{ls}, \mu_{ls} \rbrace$, using the ordinary least squares (OLS) method with TB dataset from 47 prefectures in Japan. Based on these estimates, we define the prior distribution information for the parameters as follows:
%\subsection{No-U-Turn Sampler(NUTS)}
% Hamiltonian Monte Carlo (HMC) is one of the most important MCMC algorithms, extending the Metropolis algorithm by overcoming random walk behavior and sensitivity to the correlated parameters. This advantage is particularly beneficial in high dimensional parameter target distributions, where HMC, a gradient-based method, walks faster in the parameter domain by including momentum variables, leading to faster convergence and improved mixing. However, HMC has two highly sensitive tuning parameters, namely the discretization step size ($\epsilon$) and the number of steps ($\mathcal{L}$). Selecting the optimal values for these parameters is crucial, since larger values of $\mathcal{L}$ increase the computational cost, whereas a smaller value of $\mathcal{L}$ leads to undesirable random walk behavior with the parameter space. To address this challenge, 
\begin{equation}\label{prior}
\left\lbrace
    \begin{aligned}
    \alpha &\sim \mbox{TruncatedNormal}(  \texttt{mean} = \alpha_{ls}, \texttt{SD} =   0.1,  \texttt{lower}=0, \texttt{initval}=\alpha_{ls})\\
    \beta &\sim \mbox{TruncatedNormal}( \texttt{mean} = \beta_{ls}, \texttt{SD} = 0.01, \texttt{lower}=0, \texttt{initval}=\beta_{ls})\\
    \gamma &\sim \mbox{TruncatedNormal}(\texttt{mean} = \gamma_{ls}, \texttt{SD} = 0.1, \texttt{lower}=0, \texttt{initval}=\gamma_{ls})\\
     \sigma &\sim \mbox{TruncatedNormal}(\texttt{mean} = \sigma_{ls}, \texttt{SD} = 0.3, \texttt{lower}=0, \texttt{initval}=\sigma_{ls})\\
     \mu &\sim \mbox{TruncatedNormal}(\texttt{mean} = \mu_{ls}, \texttt{SD} = 0.01, \texttt{lower}=0, \texttt{initval}=\mu_{ls})\\
    \eta &\sim \mbox{HalfNormal}(10),
\end{aligned}
\right.
\end{equation}
where $\eta$ is the likelihood distribution. We selected the `TruncatedNormal' distribution to ensure the non-negativity of the parameters, aligning with epidemiological considerations.  For each MCMC step $\hat{n}$, NUTS chooses the new parameter $\theta^{\hat{n}}$  in such a way that the set of parameters minimizes the previous deviation. At each time $t$, integration with its model parameters produces one Monte Carlo sample, and iteratively, MCMC will create the full set of required samples. %The summary of the MCMC results is provided in Table \ref{tab:mcmc}.
{\color{black} Table~\ref{tab:mcmc} summarizes the results of the MCMC sampling, conducted using the NUTS approach with 3000 samples across 2 chains, based on the prior specifications given in Eq.~\eqref{prior}. The reported statistics include the posterior mean (Mean), standard deviation (SD), highest density intervals (HDI) at 3\% and 97\%, Monte Carlo standard errors (MCSE) for both the mean and SD, effective sample sizes (ESS) for both bulk and tail estimates, and the potential scale reduction factor (R\_hat), which assesses convergence across chains. The parameter, $\alpha$, has a posterior mean of 0.04112 with a narrow standard deviation SD=0.00116, indicating good confidence in its estimate. The 94\% HDI range, given by HDI = [0.03888, 0.04324], also confirms this precision. The convergence diagnostics R\_hat =1.00793 and bulk ESS = 64.44 suggest moderate sampling efficiency and acceptable convergence. The parameters $\beta, \gamma$ can be interpreted similarly. The parameter $\sigma$ and the parameter $\mu$ are both small in magnitude. They are estimated with high precision and sampling efficiency, as indicated by large ESS values (bulk ESS) and R\_hat $\approx 1$, confirming excellent convergence. Lastly, $\eta$ also exhibits very good convergence.} Now, using the values of the estimated parameters for the MN-SIR model, we generate the predictions for the infected curve, which will be further used in the EGDL framework.
\begin{table}[!ht]
    \caption{Summary of the MCMC results with the prior information Eq. \eqref{prior}. Here, Standard Deviation (SD), Highest Density Interval (HDI), Monte Carlo Standard Error (MCSE), Effective Sample Size (ESS), and Potential Scale Reduction Factor (R\_hat) are reported.}
    \label{tab:mcmc}
    \centering
    \setlength\tabcolsep{2pt}   
    \scriptsize{
    % \resizebox{0.99\textwidth}{!}{
    \begin{tabular}{|l|l|l|l|l|l|l|l|l|l|}
    \hline
        &Mean & SD& HDI-3\% & HDI-97\% & MCSE Mean & MCSE SD & ESS Bulk & ESS Tail & R\_hat  \\ \hline
        $\alpha$ & 0.04112 & 0.00116 & 0.03888 & 0.04324 & 0.00014 & 0.00010 & 64.43525& 138.42481& 1.00793 \\ \hline
        $\beta$ & 0.00016 & 5.66E-06 & 0.00014 & 0.00017 & 8.41E-07 & 5.99E-07 & 45.53547 & 52.76396 & 1.01371 \\ \hline
       $ \gamma$ & 0.10632 & 0.00179& 0.10320 & 0.10973 & 0.00023 & 0.00016 & 62.12841 & 219.92519 & 1.01791 \\ \hline
        $\sigma$ & 3.33E-06 & 3.21E-06 & 1.10E-09 & 8.95E-06 & 6.01E-08 & 4.38E-08 & 3487.84068 & 3684.22854 & 1.00098 \\ \hline
        $\mu$ & 5.17E-05 & 5.24E-05 & 6.80E-09 & 0.00015 & 2.11E-06 & 1.49E-06 & 726.51966 & 1243.31641 & 1.00087 \\ \hline
        $\eta$ & 23.78525 & 0.17405 & 23.47343 & 24.11068 & 0.00362 & 0.00256 & 2321.03203 & 3599.16818 & 1.00046 \\ \hline
    \end{tabular}}
\end{table}

\subsubsection{Forecasting Performance Evaluation}\label{Sec_Fore_Results}
% This section first examines the causal relationship between TB incidence cases and the infected curve of the MN-SIR model (see Appendix \ref{Appendix_Causal} for details). The results show a strong causal relationship between TB incidence cases and the infection dynamic curve. This finding supports the integration of historical incidence data and the MN-SIR model’s infection curve in the EGDL architectures, leveraging the complementary strengths of both approaches to improve TB incidence forecasting. Furthermore, t
To evaluate the performance of the forecasting approaches, we utilize four key performance indicators, namely Symmetric Mean Absolute Percent Error (SMAPE), Mean Absolute Error (MAE), Mean Absolute Scaled Error (MASE), and Root Mean Squared Error (RMSE) \cite{hyndman2018forecasting} (details in Appendix \ref{Sec_Performance_measures}). {\color{black}By definition, the lower values of these metrics indicate better performance. In this evaluation, we report the average performance metrics and corresponding standard deviations (SDs) computed across 47 prefectures of Japan for various forecast horizons. To ensure robustness and consistency, the forecasting models are trained over 5 independent runs, and the reported results reflect the aggregated performance across these repetitions. While the EGDL-Parallel and EGDL-Series frameworks can be flexibly integrated with any forecasting model, our analysis employs state-of-the-art global architectures, including Transformers, NBeats, NHits, and TCN. These models are specifically designed for time series forecasting and are well-suited to multivariate setups, which is a key challenge for many traditional statistical methods \cite{lim2021time}.} %Furthermore, their performance remains consistent across varying sample sizes, demonstrating robustness to both large and small datasets. 
By integrating the EGDL-Parallel and EGDL-Series frameworks with the Transformers architecture, we develop the epidemic-guided parallel Transformers (EGP-Transformers) and epidemic-guided series Transformers (EGS-Transformers), respectively. Similarly, integrating NBeats with these frameworks results in the epidemic-guided parallel NBeats (EGP-NBeats) and epidemic-guided series NBeats (EGS-NBeats), respectively. Following the same methodology, combining EGDL-Parallel and EGDL-Series with NHits and TCN generates EGP-NHits, EGS-NHits, EGP-TCN, and EGS-TCN models, respectively. These newly developed forecasting models effectively learn temporal patterns from lagged values while incorporating epidemic dynamics via the infection curves estimated from the MN-SIR model, using the mean of the parameters reported in Table \ref{tab:mcmc}. This integration ensures that epidemic principles, including spatial interactions, guide the forecasters to predict realistic future disease trends. To assess the impact of incorporating these epidemic principles, we compare the performance of the epidemic-guided models with their corresponding baseline architectures. Additionally, we include three spatiotemporal forecasting techniques, namely GSTAR \cite{imro2023determination}, GpGp \cite{senanayake2016predicting}, and STGCN \cite{wang2022causalgnn}, which can inherently model spatial and temporal dependencies in the disease incidence cases. 

% To ensure the robustness and consistency of the empirical findings, each model is trained for 5 independent runs. The reported results represent the average performance along with the corresponding standard deviations (SDs), computed across all 47 prefectures of Japan.}

\begin{table}[!ht]
\centering
\caption{{\color{black}Mean forecast performance along with standard deviations (SDs) of spatiotemporal models, temporal deep learners, and Epidemic-Guided Deep Learning (EGDL)-Parallel approach for different forecast horizons on TB incidence data of Japan. The best results are \underline{\textbf{highlighted}}.}}
\label{Table_infected_exogenous}
\setlength\tabcolsep{1.5pt}
\tiny{
% \resizebox{0.99\textwidth}{!}{
\begin{tabular}{|c|c|cccc|cccc|cccc|}
   \hline \multirow{3}{*}{Horizon}&  \multirow{3}{*}{Metric}     & \multicolumn{4}{c|}{Spatiotemporal Models}& \multicolumn{4}{c|}{Temporal Models} & \multicolumn{4}{c|}{EGDL-Parallel Models}\\
   &       & \multirow{2}{*}{GSTAR} & \multirow{2}{*}{GpGp}   & \multirow{2}{*}{STGCN}  & \multirow{2}{*}{MN-SIR}  & Trans- & \multirow{2}{*}{NBeats} & \multirow{2}{*}{NHits}  & \multirow{2}{*}{TCN}    & EGP-Tran- & EGP- & EGP-  & EGP-    \\
   &  & & & & & formers & & & & sformers & NBeats & NHits & TCN \\ \hline
   
\multirow{8}{*}{12-month} & \multirow{2}{*}{SMAPE} & 24.746         & 29.875 & 39.432 & 45.805 & 49.296       & 24.631 & 23.339 & 32.073 & 49.170       & 24.330 & \underline{\textbf{23.337}} & 30.896 \\
    & & {\color{black}(9.97)} & {\color{black}(18.92)} & {\color{black}(8.18)} & {\color{black}36.75)} & {\color{black}(23.71)} & {\color{black}(10.45)} & {\color{black}(9.92)} & {\color{black}(23.43)} & {\color{black}(23.75)} & {\color{black}(11.13)} & {\color{black}(9.84)} & {\color{black}(21.08)} \\

   & \multirow{2}{*}{MAE}   & 5.640          & 7.428  & 13.751 & 8.920  & 17.723       & 5.302  & 5.116  & 6.994  & 17.678       & 5.246  & \underline{\textbf{5.032}}  & 5.920  \\
   & & {\color{black}(4.43)} & {\color{black}(7.34)} & {\color{black}(16.50)} & {\color{black}(8.17)} & {\color{black}(0.21)} & {\color{black}(3.52)} & {\color{black}(3.46)} & {\color{black}(8.57)} & {\color{black}(30.18)} & {\color{black}(3.62)} & {\color{black}(3.57)} & {\color{black}(3.43)} \\
  
   & \multirow{2}{*}{MASE}  & 0.979          & 1.484  & 2.067  & 1.608  & 2.316        & 0.947  & 0.901  & 1.207  & 2.309        & 0.931  & \underline{\textbf{0.878}}  & 1.132  \\
   & & {\color{black}(0.41)} & {\color{black}(2.38)} & {\color{black}(1.06)} & {\color{black}(1.08)} & {\color{black}(1.68)} & {\color{black}(0.32)} & {\color{black}(0.33)} & {\color{black}(0.79)} & {\color{black}(1.68)} & {\color{black}(0.36)} & {\color{black}(0.28)} & {\color{black}(0.63)} \\
   
   & \multirow{2}{*}{RMSE}  & 6.835          & 8.854  & 15.351 & 10.215 & 18.674       & 6.565  & 6.284  & 8.224  & 18.636       & 6.521  & \underline{\textbf{6.137}}  & 7.067  \\

   & & {\color{black}(5.11)} & {\color{black}(8.44)} & {\color{black}(17.76)} & {\color{black}(8.81)} & {\color{black}(30.40)} & {\color{black}(4.31)} & {\color{black}(4.26)} & {\color{black}(8.87)} & {\color{black}(30.37)} & {\color{black}(4.33)} & {\color{black}(4.27)} & {\color{black}(4.01)} \\    \hline

\multirow{8}{*}{9-month}  & \multirow{2}{*}{SMAPE} & 25.719         & 27.747 & 43.612 & 45.359 & 48.903       & 26.962 & 22.859 & 33.656 & 48.893       & 24.819 & \underline{\textbf{22.077}} & 26.790 \\
    & & {\color{black}(12.20)} & {\color{black}(13.54)} & {\color{black}(11.25)} & {\color{black}(36.63)} & {\color{black}(25.32)} & {\color{black}(12.78)} & {\color{black}(10.42)} & {\color{black}(27.90)} & {\color{black}(25.11)} & {\color{black}(13.80)} & {\color{black}(10.49)} & {\color{black}(14.62)} \\
   & \multirow{2}{*}{MAE}   & 5.687          & 6.268  & 15.942 & 8.835  & 17.961       & 5.592  & 5.084  & 7.173  & 17.946       & 5.286  & \underline{\textbf{5.023}}  & 5.726  \\
    & & {\color{black}(4.46)} & {\color{black}(4.55)} & {\color{black}(19.69)} & {\color{black}(7.71)} & {\color{black}(30.88)} & {\color{black}(3.60)} & {\color{black}(4.08)} & {\color{black}(8.98)} & {\color{black}(30.78)} & {\color{black}(3.65)} & {\color{black}(3.70)} & {\color{black}(3.58)} \\
   
   & \multirow{2}{*}{MASE}  & 1.019          & 1.240  & 2.616  & 1.753  & 2.524        & 1.065  & 0.931  & 1.325  & 2.505        & 0.994  & \underline{\textbf{0.892}}  & 1.138  \\
   & & {\color{black}(0.32)} & {\color{black}(1.02)} & {\color{black}(1.65)} & {\color{black}(1.47)} & {\color{black}(2.16)} & {\color{black}(0.43)} & {\color{black}(0.32)} & {\color{black}(0.87)} & {\color{black}(2.15)} & {\color{black}(0.41)} & {\color{black}(0.36)} & {\color{black}(0.68)} \\
   
   & \multirow{2}{*}{RMSE}  & 6.954          & 7.617  & 16.920 & 9.963  & 18.877       & 6.686  & 6.198  & 8.338  & 18.851       & 6.399  & \underline{\textbf{6.080}}  & 6.987  \\ 
   & & {\color{black}(5.61)} & {\color{black}(5.90)} & {\color{black}(20.10)} & {\color{black}(8.25)} & {\color{black}(31.07)} & {\color{black}(4.35)} & {\color{black}(4.74)} & {\color{black}(9.17)} & {\color{black}(30.98)} & {\color{black}(4.35)} & {\color{black}(4.60)} & {\color{black}(4.71)} \\ \hline

\multirow{8}{*}{6-month}  & \multirow{2}{*}{SMAPE} & 25.610         & 27.584 & 42.650 & 45.841 & 48.062       & 25.203 & 22.682 & 32.459 & 46.737       & 24.857 & \underline{\textbf{22.464}} & 25.644 \\
    & & {\color{black}(13.97)} & {\color{black}(16.29)} & {\color{black}(12.38)} & {\color{black}(39.75)} & {\color{black}(25.80)} & {\color{black}(17.03)} & {\color{black}(9.86)} & {\color{black}(26.38)} & {\color{black}(25.74)} & {\color{black}(12.91)} & {\color{black}(10.83)} & {\color{black}(13.69)} \\

   &  \multirow{2}{*}{MAE}   & 6.379          & 6.499  & 15.733 & 8.383  & 17.633       & 5.246  & 5.486  & 7.171  & 17.319       & 5.207  & \underline{\textbf{5.105}}  & 5.413  \\
   & & {\color{black}(7.80)} & {\color{black}(6.95)} & {\color{black}(20.24)} & {\color{black}(7.46)} & {\color{black}(29.88)} & {\color{black}(3.67)} & {\color{black}(5.45)} & {\color{black}(9.52)} & {\color{black}(29.86)} & {\color{black}(4.12)} & {\color{black}(3.95)} & {\color{black}(3.58)} \\

   &  \multirow{2}{*}{MASE}  & 1.178          & 1.283  & 2.923  & 1.915  & 2.783        & 1.119  & 0.989  & 1.406  & 2.687        & 1.096  & \underline{\textbf{0.987}}  & 1.105  \\
   & & {\color{black}(0.69)} & {\color{black}(0.85)} & {\color{black}(2.28)} & {\color{black}(1.85)} & {\color{black}(2.75)} & {\color{black}(0.59)} & {\color{black}(0.44)} & {\color{black}(0.94)} & {\color{black}(2.74)} & {\color{black}(0.49)} & {\color{black}(0.39)} & {\color{black}(0.44)} \\
   &  \multirow{2}{*}{RMSE}  & 7.375          & 7.609  & 16.705 & 9.347  & 18.400       & 6.157  & 6.386  & 8.371  & 18.080       & 6.338  & \underline{\textbf{6.012}}  & 6.287  \\
   & & {\color{black}(8.22)} & {\color{black}(7.44)} & {\color{black}(20.64)} & {\color{black}(7.80)} & {\color{black}(29.92)} & {\color{black}(4.01)} & {\color{black}(5.99)} & {\color{black}(9.75)} & {\color{black}(29.91)} & {\color{black}(4.41)} & {\color{black}(4.56)} & {\color{black}(4.00)} \\ \hline

\multirow{8}{*}{3-month}  & \multirow{2}{*}{SMAPE} & 24.533         & 25.467 & 36.300 & 45.835 & 50.266       & 28.612 & 24.292 & 31.572 & 48.540       & 26.716 & \underline{\textbf{24.070}} & 27.403 \\
    & & {\color{black}(16.26)} & {\color{black}(17.49)} & {\color{black}(17.44)} & {\color{black}(38.52)} & {\color{black}(27.49)} & {\color{black}(20.15)} & {\color{black}(16.46)} & {\color{black}(21.76)} & {\color{black}(27.55)} & {\color{black}(17.98)} & {\color{black}(15.01)} & {\color{black}(16.13)} \\

   & \multirow{2}{*}{MAE}   & 5.012          & \underline{\textbf{4.857}}  & 11.838 & 8.032  & 17.511       & 5.629  & 5.167  & 6.983  & 17.140       & 5.417  & \underline{\textbf{4.857}}  & 5.758  \\
    & & {\color{black}(3.80)} & {\color{black}(3.16)} & {\color{black}(16.27)} & {\color{black}(6.67)} & {\color{black}(27.98)} & {\color{black}(3.96)} & {\color{black}(4.18)} & {\color{black}(9.19)} & {\color{black}(28.14)} & {\color{black}(4.14)} & {\color{black}(3.74)} & {\color{black}(4.23)} \\

   & \multirow{2}{*}{MASE}  & 1.329          & 1.485  & 3.068  & 2.562  & 3.757        & 1.672  & 1.301  & 1.651  & 3.577        & 1.361  & \underline{\textbf{1.083}}  & 1.546  \\
    & & {\color{black}(1.71)} & {\color{black}(2.13)} & {\color{black}(4.55)} & {\color{black}(3.84)} & {\color{black}(4.76)} & {\color{black}(2.28)} & {\color{black}(1.36)} & {\color{black}(1.85)} & {\color{black}(4.62)} & {\color{black}(1.42)} & {\color{black}(0.75)} & {\color{black}(1.65)} \\   
   & \multirow{2}{*}{RMSE}  & 5.558          & 5.576  & 12.664 & 8.703  & 18.158       & 6.328  & 5.854  & 7.885  & 17.785       & 6.077  & \underline{\textbf{5.511}}  & 6.416 \\
   & & {\color{black}(3.88)} & {\color{black}(3.47)} & {\color{black}(16.51)} & {\color{black}(6.75)} & {\color{black}(27.88)} & {\color{black}(4.40)} & {\color{black}(4.51)} & {\color{black}(9.48)} & {\color{black}(28.04)} & {\color{black}(4.30)} & {\color{black}(4.03)} & {\color{black}(4.56)} \\

   \hline
\end{tabular}}
\end{table}

\begin{table}[]
\centering
\caption{{\color{black}Mean forecast performance along with standard deviations (SDs) of spatiotemporal models, temporal deep learners, and Epidemic-Guided Deep Learning (EGDL)-Series approach for different forecast horizons on TB incidence data of Japan. The best results are \underline{\textbf{highlighted}}.}}
\label{Table_error_remodel}
\setlength\tabcolsep{1.5pt}
\tiny{
% \resizebox{0.99\textwidth}{!}{
\begin{tabular}{|c|c|cccc|cccc|cccc|}
   \hline \multirow{3}{*}{Horizon}&  \multirow{3}{*}{Metric}     & \multicolumn{4}{c|}{Spatiotemporal Models}& \multicolumn{4}{c|}{Temporal Models} & \multicolumn{4}{c|}{EGDL-Series Models}\\
   &       & \multirow{2}{*}{GSTAR} & \multirow{2}{*}{GpGp}   & \multirow{2}{*}{STGCN}  & \multirow{2}{*}{MN-SIR}  & Trans- & \multirow{2}{*}{NBeats} & \multirow{2}{*}{NHits}  & \multirow{2}{*}{TCN}    & EGS-Tran- & EGS- & EGS-  & EGS-\\
   &  & & & & & formers & & & & sformers & NBeats & NHits & TCN \\ \hline
   
\multirow{8}{*}{12-month} & \multirow{2}{*}{SMAPE} & 24.746         & 29.875 & 39.432 & 45.805 & 49.296       & 24.631 & 23.339 & 32.073 & 24.560           & 23.464 & 23.233 & \underline{\textbf{22.931}} \\
    & & {\color{black}(9.97)} & {\color{black}(18.92)} & {\color{black}(8.18)} & {\color{black}(36.75)} & {\color{black}(23.71)} & {\color{black}(10.45)} & {\color{black}(9.92)} & {\color{black}(23.43)} & {\color{black}(10.47)} & {\color{black}(10.89)} & {\color{black}(9.26)} & {\color{black}(8.85)} \\
   & \multirow{2}{*}{MAE}   & 5.640          & 7.428  & 13.751 & 8.920  & 17.723       & 5.302  & 5.116  & 6.994  & 5.508            & \underline{\textbf{5.000}}  & 5.159  & 5.149  \\
    & & {\color{black}(4.43)} & {\color{black}(7.34)} & {\color{black}(16.50)} & {\color{black}(8.17)} & {\color{black}(30.21)} & {\color{black}(3.52)} & {\color{black}(3.46)} & {\color{black}(8.57)} & {\color{black}(4.06)} & {\color{black}(3.21)} & {\color{black}(3.71)} & {\color{black}(3.69)} \\
   
   & \multirow{2}{*}{MASE}  & 0.979          & 1.484  & 2.067  & 1.608  & 2.316        & 0.947  & 0.901  & 1.207  & 0.954            & \underline{\textbf{0.872}}  & 0.899  & 0.880  \\

   & & {\color{black}(0.41)} & {\color{black}(2.38)} & {\color{black}(1.06)} & {\color{black}(1.08)} & {\color{black}(1.68)} & {\color{black}(0.32)} & {\color{black}(0.33)} & {\color{black}(0.79)} & {\color{black}(0.33)} & {\color{black}(0.23)} & {\color{black}(0.31)} & {\color{black}(0.24)} \\
   
   & \multirow{2}{*}{RMSE}  & 6.835          & 8.854  & 15.351 & 10.215 & 18.674       & 6.565  & 6.284  & 8.224  & 6.770            & 6.219  & \underline{\textbf{6.190}}  & 6.369  \\ 
   & & {\color{black}(5.11)} & {\color{black}(8.44)} & {\color{black}(17.76)} & {\color{black}(8.81)} & {\color{black}(30.40)} & {\color{black}(4.31)} & {\color{black}(4.26)} & {\color{black}(8.87)} & {\color{black}(4.91)} & {\color{black}(4.08)} & {\color{black}(4.26)} & {\color{black}(4.43)} \\ \hline
   
\multirow{8}{*}{9-month}  & \multirow{2}{*}{SMAPE} & 25.719         & 27.747 & 43.612 & 45.359 & 48.903       & 26.962 & 22.859 & 33.656 & 24.728           & 23.354 & 23.172 & \underline{\textbf{22.765}} \\
    & & {\color{black}(12.20)} & {\color{black}(13.54)} & {\color{black}(11.25)} & {\color{black}(36.63)} & {\color{black}(25.32)} & {\color{black}(12.78)} & {\color{black}(10.42)} & {\color{black}(27.90)} & {\color{black}(11.93)} & {\color{black}(10.93)} & {\color{black}(10.16)} & {\color{black}(10.53)} \\

   & \multirow{2}{*}{MAE}   & 5.687          & 6.268  & 15.942 & 8.835  & 17.961       & 5.592  & 5.084  & 7.173  & 5.476            & 5.104  & 5.061  & \underline{\textbf{5.030}}  \\
   & & {\color{black}(4.46)} & {\color{black}(4.55)} & {\color{black}(19.69)} & {\color{black}(7.71)} & {\color{black}(30.88)} & {\color{black}(3.60)} & {\color{black}(4.08)} & {\color{black}(8.98)} & {\color{black}(3.93)} & {\color{black}(3.53)} & {\color{black}(3.38)} & {\color{black}(3.60)} \\

   & \multirow{2}{*}{MASE}  & 1.019          & 1.240  & 2.616  & 1.753  & 2.524        & 1.065  & 0.931  & 1.325  & 1.029            & 0.949  & 0.957  & \underline{\textbf{0.910}}  \\
   & & {\color{black}(0.32)} & {\color{black}(1.02)} & {\color{black}(1.65)} & {\color{black}(1.47)} & {\color{black}(2.16)} & {\color{black}(0.43)} & {\color{black}(0.32)} & {\color{black}(0.87)} & {\color{black}(0.46)} & {\color{black}(0.35)} & {\color{black}(0.37)} & {\color{black}(0.29)} \\
   
   & \multirow{2}{*}{RMSE}  & 6.954          & 7.617  & 16.920 & 9.963  & 18.877       & 6.686  & 6.198  & 8.338  & 6.673            & 6.164  & \underline{\textbf{6.145}}  & 6.178  \\ 
   & & {\color{black}(5.61)} & {\color{black}(5.90)} & {\color{black}(20.10)} & {\color{black}(8.25)} & {\color{black}(31.07)} & {\color{black}(4.35)} & {\color{black}(4.74)} & {\color{black}(9.17)} & {\color{black}(4.79)} & {\color{black}(4.24)} & {\color{black}(4.06)} & {\color{black}(4.44)} \\ \hline

\multirow{8}{*}{6-month}  & \multirow{2}{*}{SMAPE} & 25.610         & 27.584 & 42.650 & 45.841 & 48.062       & 25.203 & 22.682 & 32.459 & 24.057           & 22.786 & \underline{\textbf{21.916}} & 22.214 \\

    & & {\color{black}(13.97)} & {\color{black}(16.29)} & {\color{black}(12.38)} & {\color{black}(39.75)} & {\color{black}(25.80)} & {\color{black}(17.03)} & {\color{black}(9.85)} & {\color{black}(26.38)} & {\color{black}(12.03)} & {\color{black}(11.14)} & {\color{black}(10.55)} & {\color{black}(12.22)} \\
   & \multirow{2}{*}{MAE}   & 6.379          & 6.499  & 15.733 & 8.383  & 17.633       & 5.246  & 5.486  & 7.171  & 5.544            & 5.189  & \underline{\textbf{4.911}}  & 5.024  \\
    & & {\color{black}(7.80)} & {\color{black}(6.95)} & {\color{black}(20.24)} & {\color{black}(7.46)} & {\color{black}(29.88)} & {\color{black}(3.99)} & {\color{black}(5.45)} & {\color{black}(9.52)} & {\color{black}(4.57)} & {\color{black}(4.24)} & {\color{black}(4.02)} & {\color{black}(3.98)} \\
   
   & \multirow{2}{*}{MASE}  & 1.178          & 1.283  & 2.923  & 1.915  & 2.783        & 1.119  & 0.989  & 1.406  & 1.088            & 1.013  & 0.965  & \underline{\textbf{0.953}}  \\

   & & {\color{black}(0.69)} & {\color{black}(0.85)} & {\color{black}(2.28)} & {\color{black}(1.85)} & {\color{black}(2.75)} & {\color{black}(0.59)} & {\color{black}(0.44)} & {\color{black}(0.94)} & {\color{black}(0.50)} & {\color{black}(0.38)} & {\color{black}(0.35)} & {\color{black}(0.38)} \\
   
   & \multirow{2}{*}{RMSE}  & 7.375          & 7.609  & 16.705 & 9.347  & 18.400       & 6.157  & 6.386  & 8.371  & 6.570            & 6.234  & \underline{\textbf{5.946}}  & 6.001  \\ 
   & & {\color{black}(8.22)} & {\color{black}(7.44)} & {\color{black}(20.64)} & {\color{black}(7.80)} & {\color{black}(29.92)} & {\color{black}(3.99)} & {\color{black}(5.99)} & {\color{black}(9.75)} & {\color{black}(5.03)} & {\color{black}(4.81)} & {\color{black}(4.48)} & {\color{black}(4.47)} \\ \hline

\multirow{8}{*}{3-month}  & \multirow{2}{*}{SMAPE} & 24.533         & 25.467 & 36.300 & 45.835 & 50.266       & 28.612 & \underline{\textbf{24.292}} & 31.572 & 24.860           & 27.443 & 24.381 & 25.188 \\
    & & {\color{black}(16.26)} & {\color{black}(17.49)} & {\color{black}(17.44)} & {\color{black}(38.52)} & {\color{black}(27.49)} & {\color{black}(20.15)} & {\color{black}(16.46)} & {\color{black}(21.76)} & {\color{black}(16.47)} & {\color{black}(18.19)} & {\color{black}(16.16)} & {\color{black}(17.41)} \\
   & \multirow{2}{*}{MAE}   & 5.012          & \underline{\textbf{4.857}}  & 11.838 & 8.032  & 17.511       & 5.629  & 5.167  & 6.983  & 5.400            & 5.591  & 5.446  & 5.275  \\
   & & {\color{black}(3.80)} & {\color{black}(3.16)} & {\color{black}(16.27)} & {\color{black}(6.67)} & {\color{black}(27.98)} & {\color{black}(3.96)} & {\color{black}(4.18)} & {\color{black}(9.19)} & {\color{black}(4.81)} & {\color{black}(4.21)} & {\color{black}(5.20)} & {\color{black}(4.20)} \\
   & \multirow{2}{*}{MASE}  & 1.329          & 1.485  & 3.068  & 2.562  & 3.757        & 1.672  & 1.301  & 1.651  & \underline{\textbf{1.154}}            & 1.274  & 1.346  & 1.339  \\

    & & {\color{black}(1.71)} & {\color{black}(2.13)} & {\color{black}(4.55)} & {\color{black}(3.84)} & {\color{black}(4.76)} & {\color{black}(2.28)} & {\color{black}(1.36)} & {\color{black}(1.85)} & {\color{black}(1.10)} & {\color{black}(0.80)} & {\color{black}(1.54)} & {\color{black}(1.49)} \\

   & \multirow{2}{*}{RMSE}  & \underline{\textbf{5.558}}          & 5.576  & 12.664 & 8.703  & 18.158       & 6.328  & 5.854  & 7.885  & 6.045            & 6.275  & 6.062  & 5.960  \\ 

   & & {\color{black}(3.88)} & {\color{black}(3.47)} & {\color{black}(16.51)} & {\color{black}(6.75)} & {\color{black}(27.88)} & {\color{black}(4.40)} & {\color{black}(4.51)} & {\color{black}(9.48)} & {\color{black}(5.05)} & {\color{black}(4.53)} & {\color{black}(5.37)} & {\color{black}(4.50)} \\   
   \hline
\end{tabular}}
\end{table}

\begin{figure}[H]
    \centering
    \includegraphics[width=0.85\linewidth]{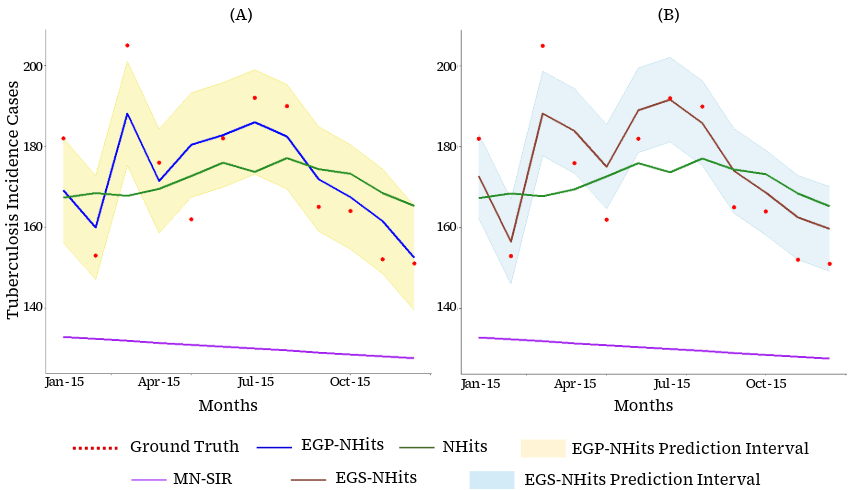}
    \caption{Visualization of the observed TB incidence cases (red dots) for prefecture 27 (Osaka) from January to December 2015, with the corresponding point forecasts from (A) EGP-NHits (blue line), (B) EGS-NHits (brown line), NHits (green line), and the infection curve generated by the MN-SIR model (violet line). The shaded regions represent conformal prediction intervals: yellow for EGP-NHits and light blue for EGS-NHits.}
    \label{Fig_Conformal_prediction_EGDLP}
\end{figure}

Tables \ref{Table_infected_exogenous} and \ref{Table_error_remodel} summarize the forecasting performance of the EGDL-Parallel and EGDL-Series frameworks, alongside several baseline models, for predicting TB incidence in Japan across multiple forecast horizons.  As reported in Table \ref{Table_infected_exogenous}, the EGP-NHits model consistently delivers superior performance in most forecasting tasks. The integration of epidemic knowledge, specifically the MN-SIR model's estimated infection curve, enables the EGP-NHits framework to produce reliable forecasts, significantly enhancing its accuracy compared to its baseline architecture. Similarly, the EGP-Transformers demonstrates superior performance than its components, Transformers and MN-SIR model. The EGP-NBeats and EGP-TCN architectures also show marked improvements over their respective baselines, further justifying the advantage of the combined forecasting techniques. {\color{black}Alongside the improved forecasting performance, the EGP-NHits and EGP-TCN architectures exhibit the lowest SDs across multiple evaluation metrics, highlighting their ability to produce reliable and consistent forecasts across various horizons for different prefectures of Japan.} Additionally, Table \ref{Table_error_remodel} reveals the superiority of the EGDL-Series frameworks over competing models for diverse temporal settings. For the 12-month forecast period, the EGS-NBeats, EGS-TCN, and EGS-NHits frameworks demonstrate the best performance across multiple indicators. For the 9-month horizon, the EGS-TCN model provides the most accurate out-of-sample predictions, followed closely by the EGS-NHits architecture. Similar trends are evident for the 6-month forecast horizon, where EGS-NHits and EGS-TCN outperform all competing models. In short-term forecasting over 3 months, the spatiotemporal GSTAR and GpGp models offer competitive performance in terms of MAE and RMSE, while the EGS-Transformers and NHits models deliver the lowest MASE and SMAPE values, respectively. {\color{black}The SDs of the performance indicators, reported in Table \ref{Table_error_remodel}, further highlight that for the 12-month horizon, EGS-NBeats and STGCN exhibit the least variability. In contrast, for the 9-month horizon, EGS-NHits and EGS-TCN architectures demonstrate greater consistency compared to competing approaches. For the 6-month and 3-month horizons, EGDL-Series frameworks generate robust forecasts alongside baseline architectures such as GSTAR, GpGp, and NHits.} Overall, the combination of deterministic and stochastic components, achieved through residual remodeling in the EGDL-Series frameworks, significantly improves the performance of individual deep learning architectures and the MN-SIR model across most forecast horizons. Furthermore, we employ the non-parametric multiple comparisons with the best (MCB) test to validate the statistical significance of the performance improvements \cite{edwards1983multiple}. {\color{black}The results of the MCB test (see Appendix \ref{Sec_Stat_Signif}) demonstrate that all variants of the EGDL architecture significantly outperform their individual components, owing to their ability to incorporate core epidemic mechanisms and spatiotemporal patterns in a unified framework.} More particularly, the performance of the data-centric approaches improved by employing the residual remodeling mechanism of EGDL-Series frameworks. Overall, the EGDL-Series architectures are particularly well suited for medium-term forecasting tasks (i.e., 9-month and 12-month horizons), whereas the EGDL-Parallel framework is more accurate for short-term (up to 6-month) forecasts of TB incidence. {\color{black}In addition to improved forecasting performance, we also evaluated the computational feasibility and scalability of the proposed frameworks by recording the average training and inference time for each model. Table \ref{computational_time} reports the total time (in seconds) required to train the models and generate 12-month-ahead forecasts for all 47 prefectures of Japan using a single core of an Intel i7 12th generation processor. Among the deep learning models, the convolution-based architectures exhibited the lowest training times due to their efficient design. The hierarchical models as well as the attention-based frameworks incurred moderate training durations owing to its increased structural complexity. The EGDL frameworks, although competitive in forecasting performance, exhibited the high computational cost, primarily due to the parameter estimation of the MN-SIR model using Bayesian inference. This estimation step, which precedes the training of the deep learning layers, significantly increases total runtime but is essential for incorporating domain-specific epidemiological knowledge into the learning process. Despite these differences, inference remains fast across all models, and 12-month forecasts for 47 prefectures of Japan could be generated within 10–20 seconds. Among the statistical baselines, GSTAR was comparatively faster to train, whereas the GpGp model was significantly more computationally intensive due to the modeling of complex spatiotemporal dependencies. This runtime analysis underscores the computational demands of the EGDL frameworks, particularly in large-scale settings; however, their strong predictive capabilities justify this cost in accuracy-critical applications. For real-time implementations, hardware acceleration and alternative estimation techniques can be employed to overcome the computational challenges.} To further assess the robustness of our models, we quantify the uncertainties associated with the EGDL frameworks using the conformal prediction approach \cite{vovk2005algorithmic}. This distribution-free method constructs prediction intervals at a pre-specified confidence level, providing robust uncertainty estimates. The prediction intervals of the EGDL-based forecasters make it a suitable choice for real-world decision-making. The conformal prediction process begins by fitting an uncertainty model to the input data and calculating conformal scores, which are derived from the residuals of the EGDL frameworks scaled by the uncertainty model's predictions. Using these scores, conformal quantiles are computed via a weighted aggregation technique. The prediction intervals are then determined by adding (subtracting) the EGDL forecasts with the upper (lower) quantiles scaled by the uncertainty model predictions. Fig. \ref{Fig_Conformal_prediction_EGDLP} presents the conformal prediction intervals for the EGP-NHits and EGS-NHits model for a selected prefecture of Japan, alongside the forecasts of baseline models. As the plots highlight, the prediction intervals effectively capture the TB incidence trends in most scenarios, demonstrating superior generalization compared to the baseline and MN-SIR models. These empirical results and prediction intervals demonstrate the forecast's uncertainties and provide significant evidence for the practical applicability of EGDL architectures in designing effective disease intervention strategies.

{\color{black}
\begin{table}[]
    \centering
    \caption{{\color{black}Training and inference time (in seconds) for forecasting models across 47 prefectures of Japan}}
    \tiny{
    \begin{tabular}{|c|cccc|cccc|}
    \hline
        {\color{black}Model} & {\color{black}GSTAR} & {\color{black}GpGp} & {\color{black}STGCN} & {\color{black}MN-SIR} & {\color{black}Transformers} & {\color{black}NBeats} & {\color{black}NHits} & {\color{black}TCN} \\ \hline
        {\color{black}Training Time} & {\color{black}0.78} & {\color{black}419.38} & {\color{black}17.46} & {\color{black}306.25} & {\color{black}49.11} & {\color{black}144.80} & {\color{black}28.85} & {\color{black}9.87} \\ 
        {\color{black}Inference Time} & {\color{black}0.06} & {\color{black}112.46} & {\color{black}0.37} & {\color{black}0.30} & {\color{black}0.11} & {\color{black}0.18} & {\color{black}0.10} & {\color{black}0.09} \\ \hline
        \multirow{2}{*}{{\color{black}Model}} & {\color{black}EGP-Tran-} & {\color{black}EGP-} & {\color{black}EGP-} & \multirow{2}{*}{{\color{black}EGP-TCN}} & {\color{black}EGS-Tran-} & {\color{black}EGS-} & {\color{black}EGS-} & \multirow{2}{*}{{\color{black}EGS-TCN}} \\
        & {\color{black}sformers} & {\color{black}NBeats} & {\color{black}NHits} &  & {\color{black}sformers} & {\color{black}NBeats} & {\color{black}NHits} &  \\ \hline
        {\color{black}Training Time} & {\color{black}349.55} & {\color{black}540.86} & {\color{black}340.61} & {\color{black}309.64} & {\color{black}348.64} & {\color{black}464.46} & {\color{black}327.20} & {\color{black}309.50} \\
        {\color{black}Inference Time} & {\color{black}0.12} & {\color{black}0.18} & {\color{black}0.11} & {\color{black}0.16} & {\color{black}0.11} & {\color{black}0.20} & {\color{black}0.12} & {\color{black}0.11} \\ 
        \hline
    \end{tabular}}
    \label{computational_time}
\end{table}
}

% Cross-Country Evaluation on China TB Dataset
{\color{black}\subsection{Results on Cross-Country Evaluation of China TB Dataset} %Application to China TB dataset
To assess the generalizability of the proposed EGDL frameworks beyond the Japan case study, we conducted a second set of experiments using publicly available TB incidence data from 31 provinces of mainland China \cite{ma2023influential}. Adopting the same methodological pipeline as used for Japan, we first preprocessed the monthly TB cases of China, ensuring that the spatial adjacency matrices accurately captured inter-provincial boundaries and transportation links. The MN-SIR model was then employed to simulate the underlying infection dynamics based on these spatial dependencies and epidemiological drivers. To validate the theoretical behavior of the model, we performed numerical simulations over the network of 31 Chinese provinces using the Laplacian matrix depicted in Fig. \ref{Fig_Correlation_LAP31_China}(A), with initial conditions consistent with Eq. \eqref{init_p} and parameters chosen according to Eq. \eqref{dis-para}. Theoretical analysis revealed convergence patterns similar to those observed in the Japan case (see Section \ref{numerical}), since we used the same set of parameter assumptions and initializations for China as for Japan. This is similar to the idea of transfer learning where we can use the learned dynamics of Japan TB cases for understanding the epidemic dynamics of China and further calibrated the model to obtain the values of the parameters reported in Table \ref{tab:mcmc_chin}. Specifically, the solutions of the entire network converged to the disease-free equilibrium point $(1000, 0)$ when the basic reproduction number satisfies $\mathcal{R}_0 = 0.385 < 1$ (Fig. \ref{fig:stab_china}(a--b) of Appendix \ref{App_Param_Calli}) and to the endemic equilibrium point $(876.67, 4.74)$ when $\mathcal{R}_0 = 3.85 > 1$ (Fig. \ref{fig:stab_china}(c--d) of Appendix \ref{App_Param_Calli}). The temporal evolution of the system also exhibited qualitatively similar behavior to that observed in the Japanese network (Fig. \ref{fig:alpha_var_china} of Appendix \ref{App_Param_Calli}). Following this theoretical validation, the parameters of the MN-SIR model were recalibrated using Bayesian inference via NUTS sampler, configured with 3,000 samples over 2 chains and using the same prior specifications as in Eq. \eqref{prior}. The MCMC results are summarized in Table \ref{tab:mcmc_chin} and can be interpreted similarly as those presented for Japan in Table \ref{tab:mcmc}. The posterior mean estimates of the MN-SIR parameters were then used to generate the infected curve, which was subsequently integrated into both the EGDL-Parallel and EGDL-Series frameworks. These architectures were implemented using global forecasting models to evaluate their performance on the TB dataset of China, maintaining consistency with the approach used for Japan.

\begin{table}[ht!]
    \caption{\color{black} Summary of the MCMC results for the China TB dataset with the prior information Eq. \eqref{prior}.}
    \label{tab:mcmc_chin}
    \centering
    \setlength\tabcolsep{1.2pt}   
    \scriptsize{
\begin{tabular}{|c|c|c|c|c|c|c|c|c|c|}
\hline
      & Mean     & SD       & HDI-3\% & HDI-97\% & MCSE Mean & MCSE SD & ESS Bulk & ESS tail & R\_hat   \\ \hline
$\alpha$ & 0.682375 & 0.097011 & 0.497088 & 0.859222 & 0.003211 & 0.002271 & 913.8217 & 1815.563 & 1.003    \\ \hline
$\beta$  & 0.000113 & 2.76E-05 & 6.52E-05 & 0.000166 & 1.17E-06 & 8.29E-07 & 563.9848 & 1069.04  & 1.00329  \\ \hline
$\gamma$ & 0.004561 & 0.002525 & 2.04E-05 & 0.009026 & 0.000428 & 0.000305 & 36.21719 & 102.8017 & 1.071449 \\ \hline
$\sigma$   & 3.08E-05 & 2.79E-05 & 3.10E-09 & 8.29E-05 & 2.65E-07 & 1.88E-07 & 8790.387 & 6415.112 & 1.000185 \\ \hline
$\mu$    & 0.005958 & 0.002559 & 0.001025 & 0.010417 & 0.000431 & 0.000307 & 36.72529 & 104.8921 & 1.068247 \\ \hline
$\eta$ & 475.6355 & 4.331996 & 467.2253 & 483.4919 & 0.064314 & 0.045489 & 4540.91  & 8069.181 & 1.000979 \\ \hline
\end{tabular}}
\end{table}
Now, we evaluate the performance of the EGDL frameworks for forecasting TB incidence cases from 31 provinces of China across different temporal settings. To maintain consistency we adopt the same evaluation setup described in Section \ref{Sec_Fore_Results}, utilizing four standard performance metrics to compare the forecasting ability of EGDL models against a suite of temporal and spatiotemporal forecasting baselines. Tables \ref{Table_infected_exogenous_China} and \ref{Table_error_remodel_China} summarize the average forecasting performance along with standard deviations of the EGDL-Parallel and EGDL-Series approaches alongside baseline models across all 31 provinces of China. As shown in Table \ref{Table_infected_exogenous_China}, the EGP-NBeats framework demonstrates superior performance in predicting 12-month ahead TB incidence dynamics. For the 9-month horizon, EGP-NBeats and EGP-NHits architectures provide competitive forecasts with the GpGp model across various metrics. For the short-term horizons of 6 months and 3 months, the EGP-NBeats and EGP-NHits showcase superior performance alongside GpGp and GSTAR models. The empirical results also highlight that integrating epidemic insights, particularly through the infection curve estimated by the MN-SIR model, enhances the predictive power of the EGDL-Parallel frameworks compared to their baseline counterparts. Additionally, Table \ref{Table_error_remodel_China} demonstrates the superiority of the EGDL-Series frameworks, especially for forecasting long-term dynamics of TB incidence cases. This improvement is largely attributed to the residual remodeling mechanism, which effectively combines deterministic and stochastic components, enhancing the predictive capabilities of both deep learning and epidemiological models. In particular, for the 12-month horizon, the EGS-NBeats framework consistently provides the most accurate forecast, while for 9-month ahead forecasting, both EGS-NBeats and EGS-Transformers architectures show competitive performance. For the 6-month forecast, EGS-Transformers outperform baseline models, whereas for the 3-month horizon, the GSTAR framework produces the most accurate forecasts, likely due to the linear characteristics of the dataset. It is important to note that, although the TB incidence dataset spans 31 provinces, the number of historical observations available for each province is relatively limited. This constraint posed significant challenges for training deep learning architectures, particularly for short-term forecasting horizons. As a result, simpler statistical models often outperformed deep learning architectures, owing to their robustness in low-data regimes. Overall, the empirical evaluation suggests that both EGDL-Parallel and EGDL-Series architectures are well-suited for capturing long-range dependencies and providing reliable forecasts over medium horizons. However, for short-term (3-month) forecasting, statistical models such as GpGp and GSTAR remain highly effective in capturing the spatiotemporal dynamics of the relatively linear TB incidence dataset. Furthermore, to validate the statistical significance of performance improvements across different horizons, we perform the non-parametric MCB test. The results of these tests, depicted in Fig. \ref{MCB_Plot_China} of Appendix \ref{Sec_Stat_Signif}, validate that all the EGDL variants significantly outperform the baseline models, demonstrating statistical robustness of their forecasting improvements. Furthermore, to quantify the uncertainties associated with the EGDL frameworks, we employ the conformal prediction approach. Fig. \ref{Fig_Conformal_prediction_China} depicts the 12-month ahead point forecasts and corresponding prediction intervals for TB incidence cases in Sichuan province, generated by the EGP-NBeats and EGS-NBeats architectures. The visualization demonstrates that both EGDL approaches effectively capture the overall trend of TB dynamics, while the conformal prediction intervals provide meaningful uncertainty estimates. These results support the practical utility and robustness of EGDL techniques across distinct epidemiological and demographic contexts and confirm its adaptability as a generalized epidemic forecasting framework.

% Table~X presents the averaged forecasting results over 5 independent runs. The EGDL architectures again outperformed baseline statistical and deep learning models across all performance metrics (SMAPE, MAE, MASE, RMSE), confirms the adaptability of the approach to a distinct epidemiological and demographic context.

\begin{table}[!ht]
\centering
\caption{\color{black}Mean forecast performance along with standard deviations (SDs) of spatiotemporal models, temporal deep learners, and Epidemic-Guided Deep Learning (EGDL)-Parallel approach in forecasting TB incidence cases of China for different forecast horizons. The best results are \underline{\textbf{highlighted}}.}
\label{Table_infected_exogenous_China}
\setlength\tabcolsep{0.6pt}
\tiny{
% \resizebox{0.99\textwidth}{!}{
\begin{tabular}{|c|c|cccc|cccc|cccc|}
   \hline \multirow{3}{*}{Horizon}&  \multirow{3}{*}{Metric}     & \multicolumn{4}{c|}{Spatiotemporal Models}& \multicolumn{4}{c|}{Temporal Models} & \multicolumn{4}{c|}{EGDL-Parallel Models}\\
   &       & \multirow{2}{*}{GSTAR} & \multirow{2}{*}{GpGp}   & \multirow{2}{*}{STGCN}  & \multirow{2}{*}{MN-SIR}    & Trans- & \multirow{2}{*}{NBeats} & \multirow{2}{*}{NHits}  & \multirow{2}{*}{TCN}    & EGP-Tran- & EGP- & EGP-  & EGP-    \\
   &  & & & & & formers & & & & sformers & NBeats & NHits & TCN \\ \hline

   \multirow{8}{*}{12-month}	&	\multirow{2}{*}{SMAPE}	&	21.84	&	16.44	&	18.94	&	22.17	&	198.83	&	19.55	&	26.47	&	143.17	&	198.66	&	\underline{\textbf{15.26}}	&	18.77	&	125.05	\\ 
	&		&	(12.11)	&	(8.61)	&	(10.14)	&	(13.60)	&	(1.29)	&	(11.54)	&	(16.49)	&	(48.75)	&	(1.51)	&	(10.81)	&	(12.23)	&	(62.92)	\\ 
	&	\multirow{2}{*}{MAE}	&	453.05	&	353.22	&	442.08	&	492.10	&	2212.54	&	370.00	&	501.15	&	2010.94	&	2212.12	&	\underline{\textbf{331.89}}	&	371.67	&	1824.08	\\ 
	&		&	(587.06)	&	(442.97)	&	(502.33)	&	(518.47)	&	(1583.32)	&	(398.14)	& (391.44)	&	(1657.91)	&	(1583.35)	&	(517.59)	&	(422.77)	&	(1213.63) \\ 
	&	\multirow{2}{*}{MASE}	&	2.08	&	1.68	&	2.04	&	12.06	&	10.91	&	1.90	&	3.21	&	12.35	&	10.90	&	\underline{\textbf{1.66}}	&	1.91	&	11.59	\\ 
	&		&	(0.79)	&	(1.13)	&	(0.99)	&	(6.94)	&	(3.19)	&	(0.79)	&	(2.32)	&	(4.52)	&	(3.20)	&	(1.61)	&	(1.06)	&	(10.58)	\\ 
	&	\multirow{2}{*}{RMSE}	&	518.01	&	\underline{\textbf{434.93}}	&	517.70	&	583.72	&	2237.18	&	456.66	&	579.61	&	2059.18	&	2236.77	&	529.27	&	448.83	&	1910.77	\\ 
	&		&	(643.64)	&	(528.77)	&	(562.32)	&	(594.01)	&	(1604.79)	&	(479.04)	&	(473.66)	&	(1669.06)	&	(1604.81)	&	(1155.41)	&	(496.59)	&	(1196.61)	\\  \hline

    \multirow{8}{*}{9-month}	&	\multirow{2}{*}{SMAPE}	&	19.05	&	19.92	&	18.93	&	23.72	&	198.72	&	22.75	&	27.40	&	103.18	&	198.62	&	\underline{\textbf{18.72}}	&	\underline{\textbf{18.72}}	&	155.11	\\ 
	&		&	(7.48)	&	(9.77)	&	(12.85)	&	(15.80)	&	(1.43)	&	(19.62)	&	(18.22)	&	(66.02)	&	(1.53)	&	(14.85)	&	(14.18)	&	(46.30)	\\ 
	&	\multirow{2}{*}{MAE}	&	452.79	&	\underline{\textbf{336.51}}	&	465.38	&	542.65	&	2148.93	&	400.40	&	535.43	&	1589.59	&	2148.67	&	345.74	&	367.95	&	2119.73	\\ 
	&		&	(492.58)	&	(461.71)	&	(603.32)	&	(611.25)	&	(1597.28)	&	(501.76)	&	(538.17)	&	(951.47)	&	(1597.30)	&	(543.58)	&	(527.72)	&	(1653.80)	\\ 
	&	\multirow{2}{*}{MASE}	&	3.08	&	2.55	&	3.08	&	26.39	&	15.86	&	3.13	&	5.03	&	20.31	&	15.86	&	\underline{\textbf{2.53}}	&	2.76	&	16.05	\\ 
	&		&	(1.24)	&	(1.21)	&	(2.00)	&	(17.88)	&	(4.78)	&	(1.73)	&	(3.69)	&	(24.54)	&	(4.78)	&	(1.40)	&	(1.68)	&	(6.17)	\\ 
	&	\multirow{2}{*}{RMSE}	&	511.39	&	\underline{\textbf{397.94}}	&	520.56	&	627.84	&	2165.74	&	471.71	&	590.48	&	1668.56	&	2165.48	&	411.45	&	429.56	&	2184.71	\\ 
	&		&	(540.53)	&	(507.19)	&	(643.19)	&	(663.77)	&	(1612.70)	&	(548.37)	&	(575.84)	&	(950.27)	&	(1612.71)	&	(586.62)	&	(573.34)	&	(1646.74)	\\  \hline

    \multirow{8}{*}{6-month}	&	\multirow{2}{*}{SMAPE}	&	18.48	&	20.20	&	21.05	&	26.44	&	198.61	&	29.77	&	32.77	&	103.50	&	198.53	&	22.84	&	\underline{\textbf{18.44}}	&	153.35	\\ 
	&		&	(7.69)	&	(9.64)	&	(14.42)	&	(18.31)	&	(1.56)	&	(23.99)	&	(18.12)	&	(65.20)	&	(1.66)	&	(19.01)	&	(12.46)	&	(43.81)	\\ 
	&	\multirow{2}{*}{MAE}	&	426.09	&	435.36	&	516.39	&	616.85	&	2053.43	&	532.12	&	625.68	&	1533.16	&	2053.28	&	\underline{\textbf{406.78}}	&	420.84	&	2041.85	\\ 
	&		&	(378.93)	&	(503.19)	&	(670.88)	&	(704.31)	&	(1600.66)	&	(671.79)	&	(609.94)	&	(950.68)	&	(1600.74)	&	(660.52)	&	(650.85)	&	(1657.19)	\\ 
	&	\multirow{2}{*}{MASE}	&	3.27	&	3.47	&	3.65	&	48.22	&	16.00	&	4.70	&	6.09	&	20.32	&	15.99	&	3.19	&	\underline{\textbf{2.93}}	&	16.33	\\ 
	&		&	(1.67)	&	(1.72)	&	(2.55)	&	(36.11)	&	(5.02)	&	(3.19)	&	(3.43)	&	(25.75)	&	(5.02)	&	(1.86)	&	(1.62)	&	(5.86)	\\ 
	&	\multirow{2}{*}{RMSE}	&	472.26	&	479.85	&	560.48	&	689.25	&	2062.90	&	630.89	&	765.93	&	1619.94	&	2062.75	&	454.04	&	\underline{\textbf{460.31}}	&	2120.00	\\ 
	&		&	(434.66)	&	(532.14)	&	(694.36)	&	(735.59)	&	(1610.42)	&	(743.31)	&	(630.43)	&	(950.09)	&	(1610.50)	&	(675.71)	&	(670.99)	&	(1642.06)	\\  \hline

    \multirow{8}{*}{3-month}	&	\multirow{2}{*}{SMAPE}	&	\underline{\textbf{15.49}}	&	19.65	&	25.22	&	29.89	&	198.50	&	30.81	&	37.75	&	139.28	&	198.39	&	23.21	&	28.45	&	118.23	\\ 
	&		&	(7.18)	&	(8.11)	&	(15.32)	&	(18.83)	&	(1.69)	&	(22.89)	&	(18.82)	&	(61.65)	&	(1.92)	&	(14.74)	&	(16.44)	&	(69.47)	\\ 
	&	\multirow{2}{*}{MAE}	&	335.69	&	\underline{\textbf{337.49}}	&	604.65	&	691.20	&	1913.85	&	500.00	&	772.95	&	1717.95	&	1913.69	&	484.61	&	532.22	&	1542.90	\\ 
	&		&	(312.41)	&	(204.39)	&	(632.28)	&	(656.96)	&	(1476.26)	&	(709.74)	&	(630.51)	&	(1616.61)	&	(1476.41)	&	(624.47)	&	(627.35)	&	(1141.89)	\\ 
	&	\multirow{2}{*}{MASE}	&	\underline{\textbf{3.52}}	&	4.55	&	5.20	&	141.43	&	20.01	&	5.39	&	8.03	&	20.80	&	20.01	&	4.76	&	6.02	&	22.67	\\ 
	&		&	(3.70)	&	(4.91)	&	(4.55)	&	(135.62)	&	(19.62)	&	(5.57)	&	(6.19)	&	(29.77)	&	(19.61)	&	(4.32)	&	(5.18)	&	(29.07)	\\ 
	&	\multirow{2}{*}{RMSE}	&	\underline{\textbf{358.44}}	&	364.29	&	623.90	&	754.38	&	1919.10	&	557.65	&	796.80	&	1731.41	&	1918.94	&	513.75	&	558.82	&	1553.01	\\ 
	&		&	(330.07)	&	(226.56)	&	(643.05)	&	(683.94)	&	(1482.70)	&	(709.34)	&	(640.83)	&	(1620.48)	&	(1482.85)	&	(631.53)	&	(631.70)	&	(1143.73)	\\ 

   \hline
\end{tabular}}
\end{table}

\begin{table}[]
\centering
\caption{\color{black}Mean forecast performance along with standard deviations (SDs) of spatiotemporal models, temporal deep learners, and Epidemic-Guided Deep Learning (EGDL)-Series approach for different forecast horizons. The best results are \underline{\textbf{highlighted}}.}
\label{Table_error_remodel_China}
\setlength\tabcolsep{0.5pt}
\tiny{
% \resizebox{0.99\textwidth}{!}{
\begin{tabular}{|c|c|cccc|cccc|cccc|}
   \hline \multirow{3}{*}{Horizon}&  \multirow{3}{*}{Metric}     & \multicolumn{4}{c|}{Spatiotemporal Models}& \multicolumn{4}{c|}{Temporal Models} & \multicolumn{4}{c|}{EGDL-Series Models}\\
   &       & \multirow{2}{*}{GSTAR} & \multirow{2}{*}{GpGp}   & \multirow{2}{*}{STGCN}  & \multirow{2}{*}{MN-SIR}      & Trans- & \multirow{2}{*}{NBeats} & \multirow{2}{*}{NHits}  & \multirow{2}{*}{TCN}    & EGS-Tran- & EGS- & EGS-  & EGS-\\
   &  & & & & & formers & & & & sformers & NBeats & NHits & TCN \\ \hline
    \multirow{8}{*}{12-month}	&	\multirow{2}{*}{SMAPE}	&	21.84	&	16.84	&	18.94	&	22.17	&	198.83	&	19.55	&	26.47	&	143.17	&	21.86	&	\underline{\textbf{16.75}}	&	24.39	&	31.19	\\ 
	&		&	(12.11)	&	(8.61)	&	(10.14)	&	(13.60)	&	(1.29)	&	(11.54)	&	(16.49)	&	(48.75)	&	(16.26)	&	(11.84)	&	(24.26)	&	(22.67)	\\ 
	&	\multirow{2}{*}{MAE}	&	453.05	&	353.22	&	442.08	&	492.10	&	2212.54	&	370.00	&	501.15	&	2010.94	&	444.42	&	\underline{\textbf{342.96}}	&	381.08	&	514.64	\\ 
	&		&	(587.06)	&	(442.97)	&	(502.33)	&	(518.47)	&	(1583.32)	&	(398.13)	&	(391.44)	&	(1657.91)	&	(621.26)	&	(533.44)	&	(483.18)	&	(595.12)	\\ 
	&	\multirow{2}{*}{MASE}	&	2.08	&	1.78	&	2.04	&	12.06	&	10.91	&	1.90	&	3.21	&	10.35	&	2.05	&	\underline{\textbf{1.74}}	&	2.23	&	2.95	\\ 
	&		&	(0.78)	&	(1.13)	&	(0.99)	&	(6.94)	&	(3.19)	&	(0.79)	&	(2.32)	&	(4.52)	&	(1.03)	&	(1.42)	&	(3.75)	&	(2.24)	\\ 
	&	\multirow{2}{*}{RMSE}	&	518.01	&	434.92	&	517.70	&	583.72	&	2237.18	&	456.66	&	579.61	&	2059.18	&	508.44	&	\underline{\textbf{423.05}}	&	571.85	&	615.30	\\ 
	&		&	(643.64)	&	(528.77)	&	(562.32)	&	(594.01)	&	(1604.79)	&	(479.04)	&	(473.67)	&	(1669.06)	&	(673.68)	&	(956.60)	&	(829.41)	&	(632.87)	\\  \hline

    \multirow{8}{*}{9-month}	&	\multirow{2}{*}{SMAPE}	&	19.05	&	19.92	&	18.93	&	23.72	&	198.72	&	22.75	&	27.40	&	103.18	&	20.47	&	\underline{\textbf{18.71}}	&	20.76	&	33.70	\\ 
	&		&	(7.48)	&	(9.77)	&	(12.85)	&	(15.80)	&	(1.43)	&	(19.62)	&	(18.22)	&	(66.02)	&	(18.00)	&	(16.51)	&	(27.33)	&	(30.66)	\\ 
	&	\multirow{2}{*}{MAE}	&	452.79	&	376.51	&	465.38	&	542.65	&	2148.93	&	400.40	&	535.43	&	1589.59	&	410.85	&	\underline{\textbf{362.38}}	&	402.76	&	495.99	\\ 
	&		&	(492.58)	&	(461.71)	&	(603.32)	&	(611.25)	&	(1597.28)	&	(501.76)	&	(538.17)	&	(951.47)	&	(701.54)	&	(658.49)	&	(767.64)	&	(580.14)	\\ 
	&	\multirow{2}{*}{MASE}	&	3.08	&	2.95	&	3.08	&	26.39	&	15.86	&	3.13	&	5.03	&	20.31	&	2.76	&	\underline{\textbf{2.65}}	&	3.25	&	4.56	\\ 
	&		&	(1.24)	&	(1.21)	&	(2.00)	&	(17.88)	&	(4.78)	&	(1.73)	&	(3.69)	&	(24.54)	&	(1.81)	&	(2.22)	&	(6.12)	&	(3.32)	\\ 
	&	\multirow{2}{*}{RMSE}	&	511.39	&	597.94	&	520.56	&	627.84	&	2165.74	&	471.71	&	590.48	&	1668.56	&	\underline{\textbf{459.24}}	&	507.32	&	491.50	&	631.52	\\ 
	&		&	(540.53)	&	(507.18)	&	(643.19)	&	(663.77)	&	(1612.70)	&	(548.37)	&	(575.84)	&	(950.27)	&	(731.33)	&	(1032.54)	&	(1388.78)	&	(620.99)	\\  \hline

    \multirow{8}{*}{6-month}	&	\multirow{2}{*}{SMAPE}	&	18.48	&	20.20	&	21.05	&	26.44	&	198.61	&	29.77	&	32.77	&	103.50	&	\underline{\textbf{17.89}}	&	19.19	&	31.58	&	34.60	\\ 
	&		&	(7.69)	&	(9.64)	&	(14.42)	&	(18.31)	&	(1.56)	&	(23.99)	&	(18.12)	&	(65.20)	&	(19.58)	&	(23.42)	&	(23.65)	&	(29.07)	\\ 
	&	\multirow{2}{*}{MAE}	&	426.09	&	435.36	&	516.39	&	616.85	&	2053.43	&	532.12	&	625.68	&	1533.16	&	\underline{\textbf{398.22}}	&	422.84	&	627.86	&	534.49	\\ 
	&		&	(378.93)	&	(503.19)	&	(670.88)	&	(704.31)	&	(1600.66)	&	(671.79)	&	(609.94)	&	(950.68)	&	(775.51)	&	(842.27)	&	(782.80)	&	(634.69)	\\ 
	&	\multirow{2}{*}{MASE}	&	3.27	&	3.47	&	3.65	&	48.22	&	16.00	&	4.70	&	6.09	&	20.32	&	\underline{\textbf{2.94}}	&	3.31	&	5.80	&	5.27	\\ 
	&		&	(1.67)	&	(1.72)	&	(2.55)	&	(36.11)	&	(5.02)	&	(3.19)	&	(3.43)	&	(25.75)	&	(2.54)	&	(3.25)	&	(5.98)	&	(3.95)	\\ 
	&	\multirow{2}{*}{RMSE}	&	472.26	&	479.85	&	560.48	&	689.25	&	2062.90	&	630.89	&	765.93	&	1619.94	&	\underline{\textbf{431.79}}	&	573.27	&	674.68	&	687.47	\\ 
	&		&	(434.66)	&	(532.14)	&	(694.36)	&	(735.59)	&	(1610.42)	&	(743.31)	&	(630.43)	&	(950.09)	&	(786.36)	&	(1277.02)	&	(1088.75)	&	(677.27)	\\  \hline

    \multirow{8}{*}{3-month}	&	\multirow{2}{*}{SMAPE}	&	\underline{\textbf{15.49}}	&	19.65	&	25.22	&	29.89	&	198.50	&	30.81	&	37.75	&	139.28	&	20.71	&	20.62	&	36.71	&	33.83	\\ 
	&		&	(7.18)	&	(8.11)	&	(15.32)	&	(18.83)	&	(1.69)	&	(22.89)	&	(18.82)	&	(61.65)	&	(18.61)	&	(27.09)	&	(31.11)	&	(35.55)	\\ 
	&	\multirow{2}{*}{MAE}	&	\underline{\textbf{335.69}}	&	337.49	&	604.65	&	691.20	&	1913.85	&	500.00	&	772.95	&	1717.95	&	402.83	&	560.50	&	769.31	&	474.60	\\ 
	&		&	(312.41)	&	(204.38)	&	(632.28)	&	(656.96)	&	(1476.26)	&	(709.74)	&	(630.51)	&	(1616.61)	&	(657.41)	&	(787.13)	&	(1003.12)	&	(631.28)	\\ 
	&	\multirow{2}{*}{MASE}	&	\underline{\textbf{3.52}}	&	4.55	&	5.20	&	141.43	&	20.01	&	5.39	&	8.03	&	20.80	&	4.27	&	5.82	&	7.85	&	6.04	\\ 
	&		&	(3.70)	&	(4.91)	&	(4.55)	&	(135.62)	&	(19.62)	&	(5.57)	&	(6.19)	&	(29.77)	&	(5.73)	&	(6.11)	&	(8.48)	&	(7.97)	\\ 
	&	\multirow{2}{*}{RMSE}	&	\underline{\textbf{358.44}}	&	364.29	&	623.90	&	754.38	&	1919.10	&	557.65	&	796.80	&	1731.41	&	423.93	&	612.03	&	858.49	&	507.97	\\ 
	&		&	(330.07)	&	(226.56)	&	(643.05)	&	(683.94)	&	(1482.70)	&	(709.34)	&	(640.83)	&	(1620.48)	&	(662.82)	&	(855.83)	&	(1104.88)	&	(637.82)	\\  \hline

   \end{tabular}
}
   \end{table}

\begin{figure}
    \centering
    \includegraphics[width=0.80\linewidth]{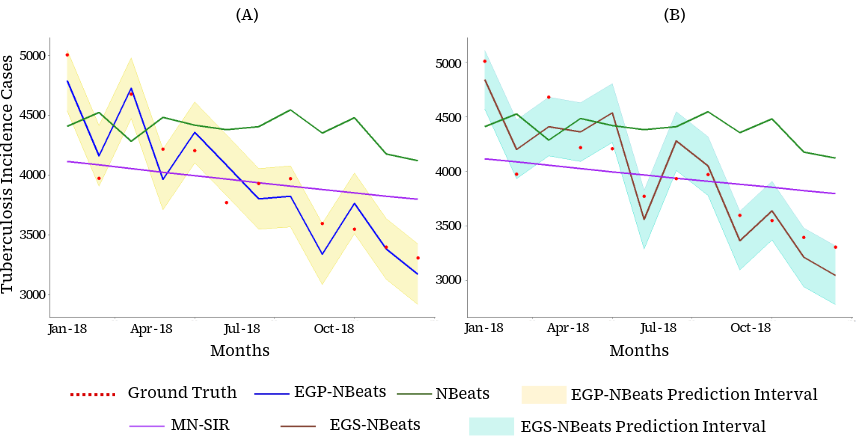}
    \caption{\color{black}{Visualization of the observed TB incidence cases (red dots) for province 23 (Sichuan) from January to December 2018, with the corresponding point forecasts from (A) EGP-NBeats (blue line), (B) EGS-NBeats (brown line), NBeats (green line), and the infection curve generated by the MN-SIR model (violet line). The shaded regions represent conformal prediction intervals: yellow for EGP-NBeats and light blue for EGS-NBeats.}}
    \label{Fig_Conformal_prediction_China}
\end{figure}

\subsection{Explainability and Model Transparency}
Despite the strong performance of deep learning models, their lack of transparency remains a significant challenge in public health applications \cite{rajkomar2019machine, tonekaboni2019clinicians}. Explainability is essential not only for scientific understanding but also for enabling trust, accountability, and policy relevance in health systems. End-users, including public health officials and epidemiologists, must be able to understand which factors drive predictions, particularly in the context of resource allocation, outbreak response, and intervention planning. Since EGDL frameworks rely on deep learning methods, they often lack transparency in how outputs are generated. Given the superior performance of the EGP-NHits architecture for forecasting TB incidence in both Japan and China, as identified by the MCB test, we focus on its explainability by leveraging the inherent hierarchical decomposition of the N-Hits model. Additionally, we apply a post hoc explainability technique, temporal gradient-based class activation maps (temporal Grad-CAM), to gain insights into model behavior and feature relevance \cite{selvaraju2017grad, das2022gradient}.

Fig. \ref{fig_blockwise_egpnhits} illustrates the inherent explainability analysis of the EGP-NHits framework, showing how individual forecast blocks contribute to the final prediction. Panel (A) corresponds to TB incidence data for prefecture 1 (Hokkaido) of Japan, while panel (B) illustrates results for province 1 (Beijing) of China. In each plot, the dashed black line represents historical TB data and the solid red line denotes the ground truth for the 12-month ahead forecast. The orange, green, and cyan lines correspond to outputs of individual blocks of the EGP-NHits architecture, each capturing different temporal resolutions and allowing us to analyze the contribution of diverse temporal scales to the overall forecast. The final forecast, depicted in blue, is computed by aggregating the individual block outputs. This block-wise decomposition enhances model output explainability by revealing how the EGP-NHits framework iteratively refines its predictions, with each block modeling distinct temporal resolution or scale. For instance, block 1 captures the short-term behavior and periodicity, block 2 models the low-frequency features, and block 3 is used for residual connections. The model finally generates the predictions by summing up all the block forecasts since some blocks capture the trend, whereas some may capture the noise terms. Notably, the combined forecast consistently aligns more closely with the ground truth than any individual block, highlighting the model's ability to integrate diverse temporal representations. Furthermore, to assess feature relevance and temporal importance across different spatial locations, we visualize the temporal Grad-CAM outputs for the EGP-NHits model in Fig. \ref{fig_EGP_NHITS_GRAD_CAM}, with the upper and lower panels demonstrating results for Japan and China, respectively. Within each panel, (A) highlights the importance of historical TB incidences across spatial units, while (B) captures the relevance of the infection curve estimated from the MN-SIR model. The x-axis represents time lags (in months) and the y-axis corresponds to prefectures (for Japan) or provinces (for China). The color intensity encodes the importance score derived from temporal Grad-CAM, with brighter regions indicating higher feature relevance. % The results reveal that certain temporal lags, particularly around 1 to 5 months and 15 to 18 months, consistently exhibit greater importance, suggesting that the EGP-NHits model leverages medium and long-term temporal dependencies to generate predictions. 
The temporal Grad-CAM results highlight that the EGP-NHits model focuses on a sparse set of specific time lags, rather than uniformly across the entire historical window. For Japan, distinct lags in both TB incidence cases and MN-SIR infection curves consistently receive higher relevance across multiple prefectures, indicating that the model identifies key temporal features that are particularly informative for forecasting. A similar pattern is observed for China, where certain discrete lags obtain higher importance scores across provinces, reflecting the model’s ability to capture region-specific temporal dependencies. Moreover, the variation in importance scores across spatial locations suggests that the model adapts to localized transmission patterns, potentially influenced by regional interventions. Notably, lagged observations of the MN-SIR model's infected curve (panel B in Fig. \ref{fig_EGP_NHITS_GRAD_CAM}) demonstrate distributed yet meaningful importance across both countries, emphasizing that incorporating mechanistic epidemiological signals improves model performance. Overall, these feature relevance patterns across time and space suggest that the EGP-NHits model effectively captures complex epidemiological dynamics while providing transparent and explainable insights into its forecasting process.

\begin{figure}
    \centering
    \includegraphics[width=1.0\linewidth]{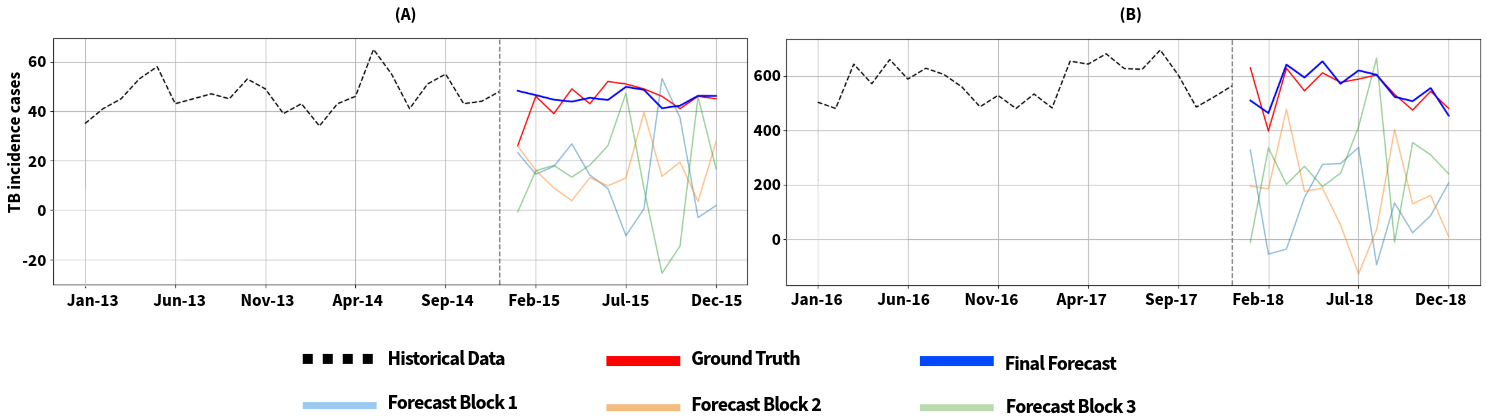}
    \caption{{\color{black}Block-wise forecast decomposition generated by the EGP-NHits architecture for TB incidence cases in (A) prefecture 1 (Hokkaido) of Japan and (B) province 1 (Beijing) of China, over a 12-month forecasting horizon.}}
    \label{fig_blockwise_egpnhits}
\end{figure}

\begin{figure}
    \centering
    \includegraphics[width=1.0\linewidth]{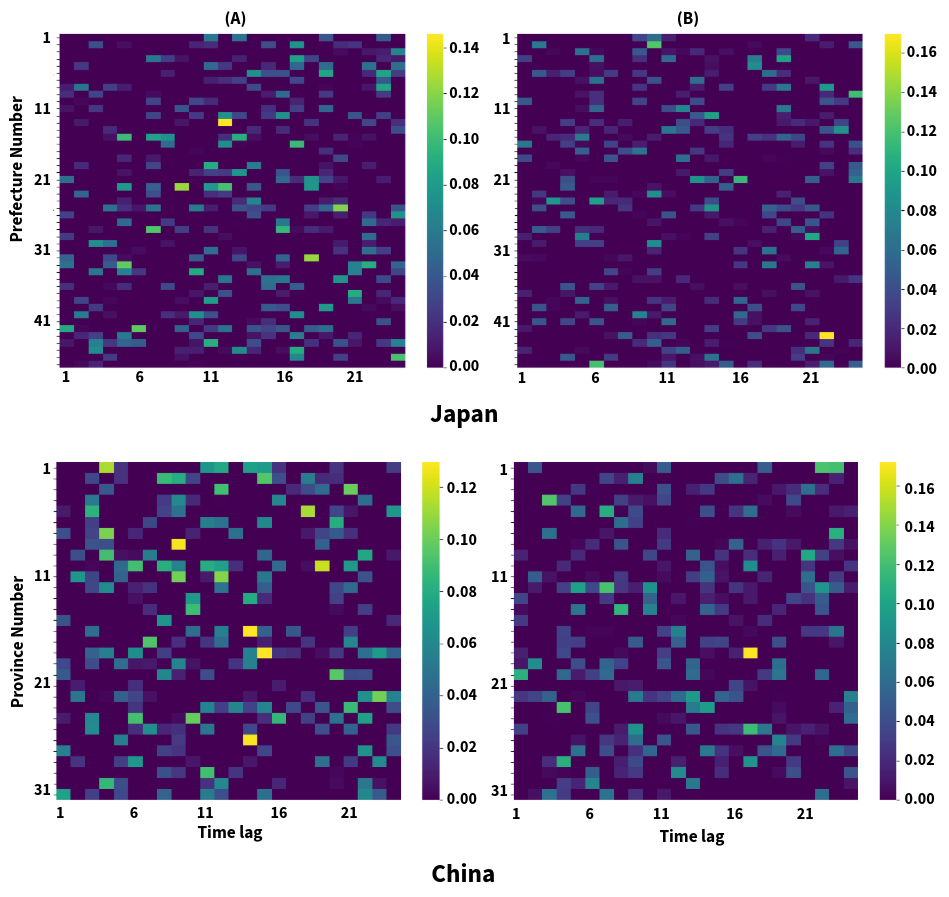}
    \caption{{\color{black}Feature attribution heatmaps for the 12-month ahead forecast of TB incidence cases in Japan (top) and China (bottom), computed using the temporal Grad-CAM approach within the EGP-NHits framework. Panel (A) shows feature attributions based on observed historical TB incidence data, while panel (B) presents attributions derived from the estimated MN-SIR model.}}
    \label{fig_EGP_NHITS_GRAD_CAM}
\end{figure}
}

{\color{black}
\section{Limitations and Future Scope of this study}
This study has several limitations. First, to define spatial connections between locations in the MN-SIR model, we adopted a simplified adjacency-based approach that relies on shared geographical borders. While this method offers interpretability and computational feasibility, it may fail to capture long-range mobility patterns influenced by socio-economic or infrastructural factors. Consequently, the spatial graph used in our framework might overlook transmission patterns between geographically distant but highly connected regions. Although more advanced spatial interaction approaches, such as Gravity \cite{anderson2011gravity} and Radiation \cite{simini2012universal} models, could address this limitation by estimating inter-regional mobility flows, their implementation typically requires high-resolution mobility data for calibration, which is often not publicly available across all locations in Japan and China. As a partial improvement, future work could consider a Gravity model using Haversine distances, computed from geographical coordinates, to approximate spatial interactions more effectively. However, mobility in both Japan and China is shaped not only by distance and population size, but also by complex socio-economic drivers such as railway infrastructure, economic zones, urban-rural linkages, and seasonal travel trends, that are not adequately captured by standard Gravity or Radiation models. Therefore, we consider our current spatial representation as a foundational step, which can be refined in future research as more detailed mobility datasets or well-validated parameterizations of the Gravity or Radiation model become available. Additionally, although the EGDL framework models spatiotemporal transmission dynamics through a combination of epidemic models and deep learning, it does not explicitly integrate socio-economic and environmental determinants that influence TB epidemiology. Variables such as income inequality, healthcare access, housing conditions, and environmental influences like air pollution and urbanization were not included due to the unavailability of consistent, high-resolution data. The exclusion of these factors may limit the model's ability to fully capture the complex socio-structural heterogeneities that shape real-world TB outbreaks. As an extension of our current study, we plan to include structured auxiliary covariates from demographic and environmental datasets into the EGDL framework as a future research direction. These variables can be integrated as additional input channels into the deep learning models, allowing the network to learn how these contextual features modulate TB incidence. Additionally, we will explore alternative architectures, such as attention mechanisms, causal diagnostics, and sparse feature attribution methods, to quantify the relative impact of these factors on transmission dynamics. These approaches can improve model explainability by identifying which input features or spatiotemporal patterns most influence the predictions, thereby making the model’s decision-making process more transparent. Such extensions will not only improve forecast accuracy but also enhance the policy relevance and epidemiological insight offered by the model. Finally, our current epidemic model does not explicitly incorporate an asymptomatic class, which is important for diseases like TB that may involve latent or subclinical carriers. Extending the mechanistic component to include an asymptomatic compartment \cite{juher2009analysis} would allow for more accurate modeling of hidden transmission dynamics.
}

\section{Conclusion and Discussion}\label{Sec_Conclusion}
% Tuberculosis (TB), a highly contagious airborne disease, was declared a global public health emergency by the WHO in 1993. Since then, numerous global initiatives have been undertaken to combat and prevent TB. The WHO's End TB Strategy aims to reduce the burden of the TB epidemic in alignment with the United Nations' Sustainable Development Goal 3: Good health and well-being. Despite these efforts, the 2024 Global TB Report highlighted a worrying increase in new TB cases, rising globally from 7.5 million in 2022 to 8.2 million in 2023. While low and middle-income countries remain high-burden regions, the impacts of rapid globalization have contributed to an uptick in TB cases in high-income countries. Japan, with the world's highest aging population and growing migration from TB-endemic regions, has shifted to a medium-burden country with rising TB incidence, presenting a significant public health and economic burden. Early detection and forecasting systems are vital for effective TB control. 

This study introduces a technological solution combining the MN-SIR model with advanced deep learning-based forecasting architectures to predict TB incidence cases. Our method bridges the gap between compartmental models, which leverage disease dynamics based on core epidemiological drivers, and data-driven forecasting techniques, which predict future disease trajectories. Integrating the MN-SIR model with deep learning frameworks allows for precise TB incidence forecasts while capturing the interactions of key disease drivers, spatial dynamics, and historical patterns. This ensures that our models provide accurate and realistic forecasts for designing effective public health interventions and making real-time policy adjustments during an outbreak.

The MN-SIR model, incorporating a saturated incidence rate and graph Laplacian diffusion, offers a robust estimation of TB transmission across Japan's 47 prefectures. The model's positivity and boundedness properties ensure realistic epidemic projections, while the global stability analysis reinforces its applicability to various populations and epidemic scenarios. Experimental results demonstrate that the EGDL-based models consistently outperform traditional data-driven models, yielding accurate forecasts across different time horizons (for short to medium-term forecasts). {\color{black}Moreover, the probabilistic outputs of the EGDL models, along with forecast explainability achieved through temporal Grad-CAM, provide valuable and transparent insights that help public health officials to track disease trends and tailor intervention strategies.} Our approach provides a comprehensive understanding of future TB dynamics, excelling in short to medium-term forecasting, while the MN-SIR model offers reliable long-term trajectory estimations. By forecasting TB incidence across Japan's 47 prefectures, this model identifies potential hotspots, guiding the design of region-specific interventions such as vaccination campaigns, resource allocation, and public health policy adjustments. {\color{black}To support the generalizability of our framework, we conducted additional experiments on TB incidence data from mainland China. The EGDL models maintained superior forecasting performance, demonstrating robustness to different population dynamics and surveillance contexts. These cross-country results confirm the broader applicability of our framework and validate the title’s claim of epidemic-guided forecasting for TB outbreaks in diverse settings.}
%While this study focuses on TB in Japan, the proposed frameworks can be adapted for other countries with varying epidemiological profiles and population dynamics. 
Additionally, the models can be extended to address the dynamics of vector-borne or waterborne diseases. Future research could also explore the application of these frameworks to multi-disease contexts, such as TB-HIV and TB-diabetes co-infections, from both global and regional perspectives. Further investigations may also examine how factors like socio-economic status, healthcare access, and environmental variables influence epidemic forecasts, improving the precision of intervention strategies.

\section*{Data and Code Availability Statement}
Data and codes are available in our github repository: \url{https://github.com/mad-stat/EGDL}.

\section*{Acknowledgments}
We are grateful to the editor, guest editors, and the three anonymous reviewers for their thoughtful comments and constructive feedback, which greatly improved the quality of this work. T.C. would also like to thank his student, Ms. Donia Besher, for her valuable assistance in collecting the data from China.

\newpage

\begin{appendices}
\setcounter{figure}{0} % Restart figure numbering
\renewcommand{\thefigure}{A\arabic{figure}}% Figure counter representation
\renewcommand{\theHfigure}{A\arabic{figure}}

\renewcommand{\thetable}{A\arabic{table}}

\section*{Appendix}\label{app}
\section{Mathematical Proofs}

\subsection{Proof of Lemma \ref{lim_glo}}\label{proof_lim_glo}
\noindent\textbf{Proof.} We consider the following two cases:
\begin{itemize}
    \item \textbf{Case-1} When $a < c$ :
    using the positivity condition of $u$, from Eq. \eqref{x_eqn} we have
    \begin{align*}
        \dot u \leq {a u} - c u = -(c-a)u, \quad \mbox{ for } t \in [0, \infty),\, c-a>0
    \end{align*}
    By standard Comparison principle and non-negativity of $u$, 
      $  \lim\limits_{t \to \infty} u = 0, \text{ provided } a< c.$
    \item \textbf{Case-2} When $a > c$ : define a Lyapunov functional 
        $W = u - u^* - u^* \ln \left(  \dfrac{u}{u^*}\right).$\\[1ex]
    Then, clearly, $W \geq 0$, for all $t \geq 0$, and $W = 0$ if and only if $u = u^*$. Differentiating $W$ with respect to $t$, it follows
    \begin{align*}
        W' &= \left( 1 - \dfrac{u^*}{u} \right) \dot u = \left( 1 - \dfrac{u^*}{u} \right) \left(  \dfrac{a u}{1+\alpha u} - c u \right)
        % &= \left( {u} - {u^*} \right) \left(  \dfrac{a }{1+\alpha u} - c  \right) = \left( {u} - {u^*} \right) \left(  \dfrac{a  - c - c\alpha u}{1+\alpha u}  \right)\\
        = -c \alpha \dfrac{\left( {u} - {u^*} \right)^2}{1+\alpha u} \leq 0,
    \end{align*}
 for $ u \geq 0,\,a > c$. {\color{black} We construct a suitable Lyapunov function \( W(u(t)) \) that is continuously differentiable and positive definite with respect to the equilibrium \( u^* \). We then show that its time derivative along the trajectories of the system satisfies \( {W}'(u(t)) \leq 0 \), indicating that \( W(u(t)) \) is non-increasing over time.} Hence,  by applying the Lyapunov-LaSalle invariance principle   \cite{martcheva2015introduction}, we obtain $\displaystyle{ \lim_{t \to \infty}u(t) = u^* = \dfrac{a-c}{\alpha c}}$.
\end{itemize}
%%%%%%%%%%%%%%%%%%%%%%%%%%%%%%%%%%%%%%%%%%%%%%%%%%%%%%%%%%
\subsection{Proof of Lemma \ref{frac_supinf}}\label{proof_frac_supinf}
\noindent\textbf{Proof.} Let us consider the following auxiliary problem
\begin{equation}
\begin{aligned}\label{limsup_abs}
       &\dfrac{\partial v}{\partial t} - \sigma \Delta v = \dfrac{\kappa v}{1+\alpha v}   - \nu v,\quad v(x, 0) = v_0(x) \geq 0\, (\not \equiv 0),\, x \in V.  
\end{aligned}
\end{equation}
By using strong maximum principle, Lemma 2.4 of \cite{tian2023asymptotic}, $v(x, t) > 0$ for all $(x, t) \in V \times(0, \infty)$, since $v_0(x) > 0$. For any sufficiently small $t_s > 0$, we define $\delta = \min\limits_{x \in V} v(x, t_s)$, which implies $\delta > 0$. Now, we consider $\hat v$ satisfies the following: 
\begin{equation}
\begin{aligned}
       &\dfrac{\partial \hat v}{\partial t} = \dfrac{\kappa \hat v}{1+\alpha \hat v}   - \nu\hat v,\quad  \hat v(x, t_s) = \delta> 0, \,  x \in V,\, t\in (t_s, \infty).  
\end{aligned}
\end{equation}
Since $V$ is finite, applying Lemma \ref{lim_glo}, we obtain
\begin{equation}\label{hatv}
  \lim_{t \to \infty} \hat v(x,t) =
  \begin{cases}
     \dfrac{\kappa - \nu}{\alpha \nu}, &\mbox{ if } \kappa > \nu,\\[0.75ex]
     0, &\mbox{ if } \kappa < \nu,
  \end{cases} \mbox{ uniformly in } x\in V.
\end{equation}
Furthermore, the Eq. \eqref{hfun} gives $\Delta \hat v(x, t) = \sum\limits_{y \in V, y\sim x} \left(\hat v(y, t) - \hat v(x, t)\right) \equiv 0$. For $t \in [t_s, \infty)$, $\hat v$ will be a lower solution of the system Eq. \eqref{limsup_abs}. By maximum principle yields $v(x, t) \geq \hat v(x, t)$ for all $(x, t) \in  V \times [t_s, \infty)$. From the relation Eq. \eqref{hatv}, it follows:
\begin{equation}\label{inf_les}
  \liminf_{t \to \infty} v \geq
  \begin{cases}
     \dfrac{\kappa - \nu}{\alpha \nu}, &\mbox{ if } \kappa > \nu,\\[0.75ex]
     0, &\mbox{ if } \kappa < \nu,
  \end{cases} \mbox{ uniformly in } x\in V.
\end{equation}
Similarly, one can get
\begin{equation}\label{sup_gret}
  \limsup_{t \to \infty} v \leq
  \begin{cases}
     \dfrac{\kappa - \nu}{\alpha \nu}, &\mbox{ if } \kappa > \nu,\\[0.75ex]
     0, &\mbox{ if } \kappa < \nu,
  \end{cases} \mbox{ uniformly in } x\in V.
\end{equation}
The above two results establish the following
\begin{equation}
  \lim_{t \to \infty} v = l = 
  \begin{cases}
     \dfrac{\kappa - \nu}{\alpha \nu}, &\mbox{ if } \kappa > \nu,\\[0.75ex]
     0, &\mbox{ if } \kappa < \nu,
  \end{cases} \mbox{ uniformly in } x\in V.
\end{equation}
The comparison principle deduces $ \liminf\limits_{t \to \infty} u(x, t) \geq l\, \Big( \limsup\limits_{t \to \infty}u(x, t) \leq  l \Big)$
uniformly in $x\in V$. Finally, the rest of the results are followed directly.
%%%%%%%%%%%%%%%%%%%%%%%%%%%%%%%%%%%%%%%%%%%%%%%%%%%
\subsection{Proof of the Lemma \ref{positive1}}\label{proof_positive1}
\noindent \textbf{Proof.}   On the hyperplane $\mathbb{D}_+$, we deduce for each $x \in V$
\begin{equation}\label{hyp}
\left\lbrace
\begin{aligned}
    \dfrac{\partial }{\partial t} S(x, t) \Big|_{S = 0} &= \Lambda + \sigma\sum_{ \substack {y \in {V}\\ y \sim x}} S > 0,\\
     \dfrac{\partial }{\partial t} I(x, t) \Big|_{I = 0} &=  \sigma\sum_{ \substack {y \in \textbf{V}\\ y \sim x}}  I \geq 0.
\end{aligned}
\right.
\end{equation}
{\color{black} The term \( \sum_{\substack{y \in V \\ y \sim x}} (S(y, t) - S(x, t)) \geq 0 \), using Eq.~\eqref{hfun}; here the negative part becomes zero as \( S(x, t) = 0 \). A similar argument can be made for \( I \).} Since $S(x,0)$ and $I(x, 0)$  in $\mathbb{D}_+$, Eq. \eqref{hyp} yields that there are no solutions $\lbrace (S, I) : x \in {V} \rbrace$ that leave the hyperplanes from $S = 0, I = 0$. The vector field is either pointing to the interior of $\mathbb{D}_+$ or tangent to the hyperplane.  Therefore, all the solutions starting in $\mathbb{D}_+$ stay in $\mathbb{D}_+$ for all time $t > 0$, i.e., the region $\mathbb{D}_+$ is positively invariant.
%%%%%%%%%%%%%%%%%%%%%%%%%%%%%%%%%%%%%%%
\subsection{Proof of the Lemma \ref{bdd}}\label{proof_bdd}
\noindent \textbf{Proof.} This can be proved easily by adding the equations of the system Eq. \eqref{submodel} and letting $P(x,t) = S(x, t) + I(x, t)$:  
\begin{align*}
     \dfrac{\partial P}{\partial t} - \sigma\Delta P &= \Lambda  - \mu P - \gamma I, \quad\,     P(x ,0) := P_0(x) = S_0(x) + I_0(x) \geq 0 \mbox{ for } x \in V\\
     &\leq  \Lambda  - \mu P, \qquad \quad\,\,     P(x ,0)  := P_0(x) = S_0(x) + I_0(x) \geq 0 \mbox{ for } x \in V
\end{align*}
Applying Lemma \ref{supinf}, we get 
\begin{align}\label{psup}
    \limsup\limits_{t\to\infty} P(x ,t)\leq \dfrac{\Lambda}{\mu}\, \mbox{ i.e.,}\, \limsup\limits_{t\to\infty} \left(S(x, t) + I(x, t) \right)\leq \dfrac{\Lambda}{\mu} \quad \mbox{ uniformly in } x \in V.
\end{align}
By maximum principle \cite{protter2012maximum, tian2020global}, we derive $P(x,t) \geq 0$. The positivity of the solutions gives  $S(x,t) \geq 0$ and $I(x,t) \geq 0$. Combining the above results, $0 \leq S(x, t), I(x, t) \leq K$  for all $(x, t) \in V \times (0, \infty)$, where the constant $K$ is given by the following expression
\begin{align*}
K = \max\limits_{x \in V} \Big\lbrace \dfrac{\Lambda}{\mu},\, \max\limits_{x \in V} P_0(x)  \Big\rbrace.
\end{align*}
%%%%%%%%%%%%%%%%%%%%%%%%%%%%%%%%%%%%%%%%%%%%%%%%
\subsection{Proof of the Theorem \ref{dfe_glo}}\label{proof_dfe_glo}
\noindent\textbf{Proof.} According to Lemma \ref{positive1}, $(S, I), \mbox{ for }(x, t) \in V\times [0, \infty)$, represents the non-negative solution of system Eq. \eqref{submodel}. We choose $\epsilon$ be sufficiently small positive real number such that $\mathcal{R}_0 < 1$, then one can write
\begin{align}\label{R0_inq}
   \beta\left(\dfrac{\Lambda}{\mu}+\epsilon\right) < (\gamma + \mu). 
\end{align}
From Eq. \eqref{submodel_a}, we obtain
\begin{align*}
                \dfrac{\partial S}{\partial t} - \sigma\Delta S \leq  \Lambda  - \mu S, \qquad\,     S(x ,0) = S_0(x) \mbox{ for } x \in V.
\end{align*}
By Lemma \ref{supinf}, we have 
\begin{align}\label{ssup}
    \limsup\limits_{t\to\infty} S(x ,t)\leq \dfrac{\Lambda}{\mu}, \quad \mbox{ uniformly in } x \in V,
\end{align}
and so, for any $\epsilon > 0$, there is a $T_1$ such that if $t > T_1,$  
\begin{align}\label{s1_less}
    S(x ,t) < \dfrac{\Lambda}{\mu} + \epsilon, \quad \mbox{ uniformly in } x \in V.
\end{align}
For any $\epsilon > 0$ satisfying Eq. \eqref{R0_inq}, substituting Eq. \eqref{s1_less} into  Eq. \eqref{submodel_b} and for $t > T_1$, we deduce
\begin{align}
    \dfrac{\partial I}{\partial t} - \sigma \Delta I &\leq \dfrac{\beta \left({\Lambda}/{\mu}+\epsilon\right) I}{1 + \alpha I} - (\gamma + \mu)I,\qquad I(x ,T_1) = I_1(x) \mbox{ for } x \in V.
\end{align}
Using Lemma \ref{frac_supinf} with the condition Eq. \eqref{R0_inq}, it follows
\begin{align}\label{isup}
    \limsup\limits_{t\to\infty} I(x, t)\leq 0, \quad \mbox{ uniformly in } x \in V,
\end{align}
hence, for any $\epsilon > 0$, there is a $T_2>T_1$ such that if $t>T_2$,
\begin{align}\label{i_les}
    I(x, t) < \epsilon, \quad \mbox{ uniformly in } x \in V.
\end{align}
The non-negativity fact of $I$ and the relation Eq. \eqref{isup} results $\lim\limits_{t\to\infty} I(x, t)= 0$  uniformly in $x \in V$.
For $t > T_2$, putting Eq. \eqref{i_les} into Eq. \eqref{submodel_a}, we find
\begin{align}
    \dfrac{\partial S}{\partial t} - \sigma\Delta S &\geq \Lambda - \left(\dfrac{\beta  \epsilon}{1 + \alpha \epsilon} +\mu \right)S,\qquad S(x ,T_2) = S_2(x) \mbox{ for } x \in V.
\end{align}
By the Lemma \ref{supinf}, 
\begin{align}\label{s1_gret}
    \liminf_{t\to \infty}S(x ,t) \geq \dfrac{\Lambda(1 + \alpha \epsilon)}{{\mu +(\beta  +\alpha\mu)\epsilon} }, \quad \mbox{ uniformly in } x \in V.
\end{align}
Since $\epsilon$ is an arbitrary real number, we let $\epsilon \to 0$, which gives
\begin{align}\label{sinf}
    \liminf\limits_{t\to\infty} S(x, t)\geq \dfrac{\Lambda}{\mu}, \quad \mbox{ uniformly in } x \in V.
\end{align}
Combining the relations Eq. \eqref{ssup} and Eq. \eqref{sinf}, we have $\lim\limits_{t\to\infty} S(x, t) = \dfrac{\Lambda}{\mu}$  uniformly in $x \in V$. This completes the proof.
%%%%%%%%%%%%%%%%%%%%%%%
\subsection{Proof of the Theorem \ref{ee_glo}}\label{proof_ee_glo}
\noindent\textbf{Proof.} Denote
\begin{align}
\begin{cases}
       \,\overline S(x) = \limsup\limits_{t \to \infty} S(x, t), \quad  \underline S(x) = \liminf\limits_{t \to \infty} S(x, t),\\[1ex]
        \,\overline I(x) = \limsup\limits_{t \to \infty} I(x, t), \quad\,\,\,  \underline I(x) = \liminf\limits_{t \to \infty} I(x, t),
\end{cases}
\quad \mbox{for all } x \in V.
\end{align}
Our claim will be $\overline S(x) = \underline S(x)=S^*,\, \overline I(x) = \underline I(x)=I^*$ uniformly in $x \in V$, which is established in the following:\\[1ex]
\noindent we choose $\epsilon$ be sufficiently small positive real number such that if $\mathcal{R}_0 > 1$,
\begin{align}\label{R0_grt}
   \beta\left(\dfrac{\Lambda}{\mu}+\epsilon\right) > (\gamma + \mu). 
\end{align}
From Eq. \eqref{submodel_a}, we obtain
\begin{align*}
                \dfrac{\partial S}{\partial t} - \sigma\Delta S \leq  \Lambda  - \mu S, \qquad\,     S(x ,0) = S_0(x) \mbox{ for } x \in V.
\end{align*}
By Lemma \ref{supinf}, we have 
\begin{align}\label{A1S}
    \limsup\limits_{t\to\infty} S(x ,t)\leq \dfrac{\Lambda}{\mu} :=A_1^S, \quad \mbox{ uniformly in } x \in V,
\end{align}
and so, for any $\epsilon > 0$, there is a $T_1$ such that if $t > T_1,$  
\begin{align}\label{e_s1_less}
    S(x ,t) < A_1^S + \epsilon, \quad \mbox{ uniformly in } x \in V.
\end{align}
For any $\epsilon > 0$ satisfying Eq. \eqref{R0_grt}, substituting Eq. \eqref{e_s1_less} into Eq. \eqref{submodel_b} and for $t > T_1$, we deduce
\begin{align}
    \dfrac{\partial I}{\partial t} - \sigma \Delta I &\leq \dfrac{\beta \left(A_1^S+\epsilon\right) I}{1 + \alpha I} - (\gamma + \mu)I,\qquad I(x ,T_1) = I_1(x) \mbox{ for } x \in V.
\end{align}
Using Lemma \ref{frac_supinf} with the condition Eq. \eqref{R0_grt}, it follows
\begin{align*}
    \limsup_{t\to\infty} I(x, t) \leq \dfrac{\beta \left(A_1^S+\epsilon\right) - (\gamma + \mu)}{\alpha(\gamma + \mu)}, \quad \mbox{ uniformly in } x \in V.
\end{align*}
As the above inequality holds for any arbitrary $\epsilon > 0$, we derive $\limsup\limits_{t\to\infty} I(x, t) \leq A_1^I,$ where
\begin{align}\label{A1I}
    A_1^I = \dfrac{\beta A_1^S - (\gamma + \mu)}{\alpha(\gamma + \mu)}.
\end{align}
We acquire $A_1^I > 0$, since $\mathcal{R}_0 > 1$. So, for any $\epsilon > 0$ there is a $T_2 > T_1$ such that if $t > T_2$, $I(x, t) < A_1^I + \epsilon$ uniformly in $x\in V$. Substituting the last result into Eq. \eqref{submodel_a}, when $t>T_2$, we obtain
\begin{align*}
    \dfrac{\partial S}{\partial t} - \sigma\Delta S &\geq \Lambda - \dfrac{\beta(A_1^S+\epsilon)(A_1^I + \epsilon)}{1 + \alpha (A_1^I + \epsilon)} - \mu S, \qquad\,     S(x ,T_2) = S_2(x) \mbox{ for } x \in V.
\end{align*}
By  Lemma \ref{supinf}, we derive
\begin{align*}
    \liminf_{n \to \infty} S(x, t) \geq  \dfrac{1}{\mu}\left[ \Lambda - \dfrac{\beta(A_1^S+\epsilon)(A_1^I + \epsilon)}{1 + \alpha (A_1^I + \epsilon)} \right], \quad \mbox{ uniformly in } x \in V.
\end{align*}
Due to arbitrariness of $\epsilon > 0$, we deduce $\liminf\limits_{t \to\infty} S(x, t) \geq B_1^S,\mbox{ uniformly in } x \in V$, where
\begin{align}\label{B1S}
    B_1^S = \dfrac{1}{\mu}\left[ \Lambda - \dfrac{\beta A_1^S A_1^I }{1 + \alpha A_1^I } \right].
\end{align}
Some simple calculations and the condition $\mathcal{R}_0 > 1$ yield $B_1^S > 0$. Hence, for any $\epsilon > 0$ there is $T_3 > T_2$ such that if $t > T_3$, $ S(x, t) > B_1^S - \epsilon$, uniformly in $x\in V$. Using the latest inequality in Eq. \eqref{submodel_b}, when $t>T_3,$  we derive
\begin{align*}
    \dfrac{\partial I}{\partial t} - \sigma \Delta I &\geq  \dfrac{\beta (B_1^S - \epsilon)I}{1 + \alpha I} - (\gamma + \mu)I,\qquad I(x ,T_3) = I_3(x) \mbox{ for } x \in V.
\end{align*}
With a few derivation and $\mathcal{R}_0>1$,  $\beta (B_1^S - \epsilon) > (\gamma + \mu)$. By Lemma \ref{frac_supinf}, we have
\begin{align*}
    \liminf_{t \to \infty} I(x, t)  \geq \dfrac{ \beta (B_1^S - \epsilon)-(\gamma + \mu)}{\alpha(\gamma + \mu)}, \quad\mbox{ uniformly in } x \in V.
\end{align*}
This inequality satisfies for every $\epsilon>0$, we conclude $\liminf\limits_{t \to \infty} I(x, t) \geq B_1^I$, where
\begin{align}\label{B1I}
    B_1^I = \dfrac{ \beta B_1^S-(\gamma + \mu)}{\alpha(\gamma + \mu)}.
\end{align}
Again one can easily show that $B_1^I > 0$, when $\mathcal{R}_0 > 1.$ Thus, there is $T_4>T_3$ such that if $t>T_4$, $ I(x, t)  \geq B_1^I - \epsilon$, uniformly in $x\in V$. In a similar manner, we derive from  Eq. \eqref{submodel_a},  for $t>T_4$
\begin{align*}
     \dfrac{\partial S}{\partial t} - \sigma\Delta S &\leq  \Lambda - \dfrac{\beta (B_1^S-\epsilon) (B_1^I-\epsilon)}{1 + \alpha  (B_1^I-\epsilon)} - \mu S, \qquad\,     S(x ,T_4) = S_4(x) \mbox{ for } x \in V.
\end{align*}
Applying Lemma \ref{supinf}, we acquire
\begin{align*}
    \limsup_{t \to\infty} S(x, t) \leq \dfrac{1}{\mu} \left[\Lambda - \dfrac{\beta (B_1^S-\epsilon) (B_1^I-\epsilon)}{1 + \alpha  (B_1^I-\epsilon)} \right], \quad\mbox{ uniformly in } x \in V
\end{align*}
which is true for any  $\epsilon > 0$. We deduce $\limsup\limits_{t \to\infty} S(x, t) \leq A_2^S$, where
\begin{align}\label{A2S}
    A_2^S = \dfrac{1}{\mu} \left[\Lambda - \dfrac{\beta B_1^S B_1^I}{1 + \alpha  B_1^I} \right].
\end{align}
Again it can be shown directly that $A_2^S>0$, if $\mathcal{R}_0 >1$. Therefore, for any $\epsilon>0$, there is $T_5>T_4$ such that if $t>T_5$, $S(x, t) < A_2^S +\epsilon$, uniformly in $x \in V$. From Eq. \eqref{submodel_b},  we derive  
\begin{align}
    \dfrac{\partial I}{\partial t} - \sigma \Delta I &\leq \dfrac{\beta \left(A_2^S+\epsilon\right) I}{1 + \alpha I} - (\gamma + \mu)I,\qquad I(x ,T_5) = I_5(x) \mbox{ for } x \in V.
\end{align}
By Lemma \ref{frac_supinf} with the condition Eq. \eqref{R0_grt}, it gives
\begin{align*}
    \limsup_{t\to\infty} I(x, t) \leq \dfrac{\beta \left(A_2^S+\epsilon\right) - (\gamma + \mu)}{\alpha(\gamma + \mu)}, \quad \mbox{ uniformly in } x \in V.
\end{align*}
Since the above inequality satisfies for any $\epsilon > 0$, we get  $\limsup\limits_{t\to\infty} I(x, t) \leq A_2^I,$ where
\begin{align}\label{A2I}
    A_2^I = \dfrac{\beta A_2^S - (\gamma + \mu)}{\alpha(\gamma + \mu)}.
\end{align}
By some direct calculations, we obtain $A_1^I > 0$, as $\mathcal{R}_0 > 1$. So, for any $\epsilon > 0$ there is a $T_6 > T_5$ such that if $t > T_6$, $I(x, t) < A_2^I + \epsilon$ uniformly in $x\in V$. From Eq. \eqref{submodel_a}, when $t>T_2$, we find
\begin{align*}
    \dfrac{\partial S}{\partial t} - \sigma\Delta S &\geq \Lambda - \dfrac{\beta(A_2^S+\epsilon)(A_2^I + \epsilon)}{1 + \alpha (A_2^I + \epsilon)} - \mu S, \qquad\,     S(x ,T_6) = S_6(x) \mbox{ for } x \in V,
\end{align*}
Using Lemma \ref{supinf}, we derive
\begin{align*}
    \liminf_{n \to \infty} S(x, t) \geq  \dfrac{1}{\mu}\left[ \Lambda - \dfrac{\beta(A_2^S+\epsilon)(A_2^I + \epsilon)}{1 + \alpha (A_2^I + \epsilon)} \right], \quad \mbox{ uniformly in } x \in V.
\end{align*}
Since above inequality is true for any $\epsilon > 0$, we conclude $\liminf\limits_{t \to\infty} S(x, t) \geq B_2^S,\mbox{ uniformly in } x \in V$, where
\begin{align}\label{B2S}
    B_2^S = \dfrac{1}{\mu}\left[ \Lambda - \dfrac{\beta A_2^S A_2^I }{1 + \alpha A_2^I } \right].
\end{align}
A few simple computations and the condition $\mathcal{R}_0 > 1$ lead $B_1^S > 0$. Thus, for any $\epsilon > 0$ there is $T_7 > T_6$ such that if $t > T_7$, $ S(x, t) > B_2^S - \epsilon$, uniformly in $x\in V$. Again Eq. \eqref{submodel_b}, when $t>T_7,$  it follows
\begin{align*}
    \dfrac{\partial I}{\partial t} - \sigma \Delta I &\geq  \dfrac{\beta (B_2^S - \epsilon)I}{1 + \alpha I} - (\gamma + \mu)I,\qquad I(x ,T_7) = I_7(x) \mbox{ for } x \in V
\end{align*}
With a few easy calculations and $\mathcal{R}_0>1$, by Lemma \ref{frac_supinf}, we have
\begin{align*}
    \liminf_{t \to \infty} I(x, t)  \geq \dfrac{ \beta (B_2^S - \epsilon)-(\gamma + \mu)}{\alpha(\gamma + \mu)}, \quad\mbox{ uniformly in } x \in V.
\end{align*}
This inequality holds for any $\epsilon>0$, we obtain $\liminf\limits_{t \to \infty} I(x, t) \geq B_2^I$, where
\begin{align}\label{B2I}
    B_2^I = \dfrac{ \beta B_2^S-(\gamma + \mu)}{\alpha(\gamma + \mu)}.
\end{align}
Since $\mathcal{R}_0 > 1$, we obtain $B_2^I > 0$. Hence, for any $\epsilon > 0$ there is a $T_8 > T_7$ such that if $t>T_8$, $ I(x, t) > B_2^I - \epsilon$ uniformly in $x \in V$.\\[1ex]
\noindent If we carry on similar arguments, we acquire four sequences $\lbrace A_n^S\rbrace_{n=1}^\infty, \lbrace A_n^I\rbrace_{n=1}^\infty, \lbrace B_n^S\rbrace_{n=1}^\infty$ and $\lbrace B_n^I\rbrace_{n=1}^\infty$, where 
\begin{equation}
    \begin{aligned}\label{ABs}
            A_n^S &= \dfrac{1}{\mu} \left[\Lambda - \dfrac{\beta B_{n-1}^S B_{n-1}^I}{1 + \alpha  B_{n-1}^I} \right],\quad  A_n^I = \dfrac{\beta A_n^S - (\gamma + \mu)}{\alpha(\gamma + \mu)},\\
            B_n^S &= \dfrac{1}{\mu}\left[ \Lambda - \dfrac{\beta A_n^S A_n^I }{1 + \alpha A_n^I } \right],\quad\quad B_n^I = \dfrac{ \beta B_n^S-(\gamma + \mu)}{\alpha(\gamma + \mu)},\, n\geq 2. 
    \end{aligned}
\end{equation}
The term $A_2^S$ can be expressed as $A_2^S = A_1^S - \dfrac{\beta B_1^S B_1^I}{\mu(1+\alpha B_1^i)}$. Since $\dfrac{\beta B_1^S B_1^I}{\mu(1+\alpha B_1^i)} = \dfrac{\beta A_1^S - (\gamma + \mu)}{\alpha \mu} > 0, $ when $\mathcal{R}_0>1$, we have $A_2^S < A_1^S$. Hence $A_2^I < A_1^I$. We calculate from Eq. \eqref{B1S} and Eq. \eqref{B2S},
\begin{align*}
    B_2^S = \dfrac{1}{\mu}\left[\Lambda - (\gamma+\mu)A_2^I  \right], \quad B_1^S = \dfrac{1}{\mu}\left[\Lambda - (\gamma+\mu)A_1^I  \right].
\end{align*}
Using the fact $A_2^I < A_1^I$, it is clear that $B_2^S > B_1^S$. Applying induction, one can prove easily that $A_{n+1}^S \leq A_{n}^S$  and $B_{n+1}^S \geq B_{n}^S$  for all $n \geq 2$. Therefore, $\lbrace A_n^S\rbrace_{n=1}^\infty$ and $\lbrace B_n^S\rbrace_{n=1}^\infty$ are non-increasing sequence and non-decreasing sequence, respectively. Continuing similar arguments for the sequences $\lbrace A_n^I\rbrace_{n=1}^\infty$ and $\lbrace B_n^I\rbrace_{n=1}^\infty$, we obtain,  $\lbrace A_n^I\rbrace_{n=1}^\infty$ and $\lbrace B_n^I\rbrace_{n=1}^\infty$ are non-increasing sequence and non-decreasing sequence, respectively. Combining $\liminf \leq \limsup$ and the above results, we have
\begin{align}
    B_n^S \leq \underline S(x) \leq \overline S(x) \leq A_n^S,\, \mbox{ and } B_n^I \leq \underline I(x) \leq \overline I(x) \leq A_n^I, \mbox{ for all } x \in V. 
    \end{align}
Thus, the limits of all sequences exist such that
\begin{align}\label{limsup-liminf}
    \underline S(x) = \lim_{ n \to \infty} B_n^S, \quad  \overline S(x) = \lim_{ n \to \infty} A_n^S,\quad \underline I(x) = \lim_{ n \to \infty} B_n^I,\mbox{ and }  \overline I(x) = \lim_{ n \to \infty} A_n^I, 
    \end{align}
for all  $x \in V$. From Eq. \eqref{ABs}, we obtain
\begin{equation}\label{compI}
\begin{aligned}
     A_{n+1}^S &= \dfrac{1}{\mu} \left[\Lambda - \dfrac{\beta B_{n}^S B_{n}^I}{1 + \alpha  B_{n}^I} \right] = \dfrac{1}{\mu} \left[\Lambda - \dfrac{\beta B_n^S-(\gamma+\mu)}{\alpha} \right],\\
     \quad B_n^S &= \dfrac{1}{\mu}\left[ \Lambda - \dfrac{\beta A_n^S A_n^I }{1 + \alpha A_n^I } \right] = \dfrac{1}{\mu} \left[\Lambda - \dfrac{\beta A_n^S-(\gamma+\mu)}{\alpha} \right].
\end{aligned}
\end{equation}
Taking limit as $t \to \infty$ and using equation (\ref{limsup-liminf}), we acquire
\begin{align}\label{barS}
     \overline S(x) & = \dfrac{1}{\mu} \left[\Lambda - \dfrac{\beta \underline S(x)-(\gamma+\mu)}{\alpha} \right],
     \quad \underline S(x) =  \dfrac{1}{\mu} \left[\Lambda - \dfrac{\beta \overline S(x)-(\gamma+\mu)}{\alpha} \right],
\end{align}
for each $ x \in V$. With some direct calculations after subtracting the above two equations, it gives  
\begin{align}
    (\alpha \mu - \beta) \left( \overline S(x) - \underline S(x) \right) = 0, \quad \mbox{for } x \in V.
\end{align}
If $\alpha \mu \neq \beta$, we get $\overline S(x) = \underline S(x), \quad \mbox{for } x \in V.$ Hence, we have from Eq. \eqref{barS}, $ \overline S(x) = \underline S(x) = \dfrac{\alpha \Lambda + (\gamma + \mu)}{\beta + \alpha \mu} = S^*$, for all $x\in V$. It implies $\lim\limits_{t \to \infty} S(x, t) = S^*$, uniformly in $x\in V$.  Using this and Eq. \eqref{compI}, one may derive that   $\lim\limits_{t \to \infty} I(x, t) = I^*$, uniformly in $x\in V$. Hence, the endemic equilibrium $(S^*, I^*)$ is globally asymptotically stable if $\mathcal{R}_0 > 1$.

\section{Workflow of EGDL Frameworks}\label{Appendix_EGDL_Algo}
The workflow of the EGDL-Parallel and EGDL-Series frameworks is presented in Algorithm \ref{algo_EGDL_Parallel}.

\begin{algorithm}
%\renewcommand{\thealgorithm}{}
%\begin{algorithmic}
    %\SetKwFunction{isOddNumber}{isOddNumber}
    \SetKwInOut{KwIn}{Input}
    \SetKwInOut{KwOut}{Output}
    \KwIn{Incidence cases recorded at $n$ locations in $T$ consecutive timestamps as $\left\{Y(x, t)\right\}_{x \in V, \; t \in \mathcal{T}}; \; V = \{1, 2, \ldots, n\}; \; \mathcal{T} = \{1, 2, \ldots, T\}$.}
    \KwOut{$q$-step ahead forecast of the disease incidence cases for the $n$ locations as $\left\{\widehat{Y}(x, T+v)\right\}_{x \in V};  \; v = 1, 2, \ldots, q$ ($\geq 1$).}
    \vspace{0.15cm}
\textbf{Model Epidemic Information:}\\
\nl  Build a MN-SIR epidemic model with saturated incidence rate and graph Laplacian diffusion by considering disease drivers across $n$ locations.\\  %such as transmission, recovery, saturation, death rates, and population mobility 
\nl Calculate the posterior distribution of the MN-SIR model parameters using the MCMC approach and select the best-fitted parameters to quantify the spatial dynamics of disease transmission and the effect of key epidemic drivers. \\  %input data $\left\{Y(1, t), Y(2, t), \ldots, Y(n, t)\right\};  \; t = 1, 2, \ldots, T$ by employing
\nl Solve the MN-SIR model to estimate the infected curve across $n$ locations for $\tilde{\tau}$ timestamps as $\left\{I(x, \tilde{\tau})\right\}_{x \in V};  \; \tilde{\tau} = 1, 2, \ldots, T+q$. \\

\setcounter{AlgoLine}{0}
\noindent \textbf{Forecasting Procedure for EGDL-Parallel Framework:}\\
\nl Initialize a data-driven forecasting model $f_{EGDLP} \in \mathcal{F}$, where $\mathcal{F}$ represents set of global deep learning-based forecasting tools. \\
\nl Provide a hybrid input to $f_{EGDLP}$ framework combining the historical incidence cases $Y(x,t)$ and the estimated infected curve of MN-SIR $I(x,t)$, formulated as $\left\{\left\{Y\left(x, \tilde{t} - t_0 \right)\right\}_{t_0 = 0}^{t_w - 1}, \; \left\{I\left(x, \tilde{t} - t_0 \right)\right\}_{t_0 = 0}^{t_w - 1}\right\}_{\tilde{t} = t_w} ^ T,$ with $t_w$ lagged time steps, where $Y(x,t)$ serves as endogenous variable and $I(x,t)$ acts as the exogenous variable.\\
\nl Train the $f_{EGDLP}$ model to encode the hidden temporal patterns from the input data and epidemic knowledge from the infected curve to forecast $q$-step ahead future trajectory of the endogenous variables as 
$$ \widehat{Y}(x, \tilde{t} + v) = f_{EGDLP} \left(\left\{Y\left(x, \tilde{t} - t_0 \right)\right\}_{t_0 = 0}^{t_w - 1}, \; \left\{I\left(x, \tilde{t} - t_0 \right)\right\}_{t_0 = 0}^{t_w - 1}\right), \; v = 1, 2, \ldots, q.$$ 
% \vspace{0.2cm}
\setcounter{AlgoLine}{0}
\textbf{Forecasting Procedure for EGDL-Series Framework:}\\
\nl Obtain the residuals of the MN-SIR model by subtracting the estimated infected curve from the ground truth observations as  ${e(x, t) = Y(x, t) - I(x, t)};  \; x \in V, \; t \in \mathcal{T}$. \\
% \nl Initialize a data-driven forecasting model $f_{EGDLS} \in \mathcal{F}$ and provide the $t_w$ lagged values of the residual series as input to generate the q-step ahead forecasts as:
% $$
%     \hat{e}(x, \tilde{t}+v) = f_{EGDLS}\left(\left\{e\left(x, \tilde{t} - t_0 \right)\right\}_{t_0 = 0}^{t_w - 1}\right), \; \tilde{t} = t_w, \ldots, T; \; v = 1, 2, \ldots, q.
% $$ 
% % \nl Obtain the q-step ahead predictions from the trained $f_{EGDLS}$ model as $\left\{\hat{e}(1, T+v), \hat{e}(2, T+v), \ldots, \hat{e}(n, T+v)\right\};  \; v = 1, 2, \ldots, q$ \\
% \nl Combine the deterministic predictions of the MN-SIR model $I(x, \tilde{t}+v)$, with the forecasts of stochastic components from $f_{EGDLS}$ framework as 
% $$
% \widehat{Y}(x, \tilde{t}+v) = I(x, \tilde{t}+v) + \hat{e}(x, \tilde{t}+v),
% $$
% to obtain the $q$-step ahead forecast of the disease incidence cases.
\nl Initialize a data-driven forecasting model $f_{M} \in \mathcal{F}$ and provide the $t_w$ lagged values of the residual series as input to generate the q-step ahead forecasts as:
$$
    \hat{e}(x, \tilde{t}+v) = f_{M}\left(\left\{e\left(x, \tilde{t} - t_0 \right)\right\}_{t_0 = 0}^{t_w - 1}\right), \; \tilde{t} = t_w, \ldots, T; \; v = 1, 2, \ldots, q.
$$ 
% \nl Obtain the q-step ahead predictions from the trained $f_{EGDLS}$ model as $\left\{\hat{e}(1, T+v), \hat{e}(2, T+v), \ldots, \hat{e}(n, T+v)\right\};  \; v = 1, 2, \ldots, q$ \\
\nl Combine the deterministic components of the MN-SIR model $I(x, \tilde{t}+v)$ with the forecasts of stochastic components to generate the predictions of the EGDL-Series methods as:  
$$
    \widehat{Y}(x, \tilde{t}+v) = \hat{f}_{EGDLS}(x, \tilde{t}+v) = I(x, \tilde{t}+v) + \hat{e}(x, \tilde{t}+v).
$$
% \vspace{0.1cm}
%\vspace{0.2cm}
\caption{{\bf Proposed EGDL Framework}
\label{algo_EGDL_Parallel}
}
\end{algorithm}

\section{Empirical Results}\label{Appendix_Global_Char}
{\color{black}Tables \ref{Table_Global_Features} and \ref{Table_Features_CHina} summarize the descriptive statistics, including measures of central tendency (Mean), standard deviation (Sd), coefficient of variation (CV), skewness, and kurtosis for TB datasets of Japan and China, respectively}. Additionally, to investigate key statistical characteristics of these datasets, we study their global patterns, as outlined below: \textit{Long-term dependency} reflects the self-similarity of the time series with its lagged observations. This feature is crucial for probabilistic time series modeling approaches. In this study, the Hurst exponent is computed to detect whether TB incidence trends exhibit long-range dependence over time. \textit{Stationarity} ensures the mean and variance of a time series remain constant over time, implying no systematic changes in trend or seasonal patterns. This property is essential for many classical forecasting models. The Kwiatkowski–Phillips–Schmidt–Shin (KPSS) test is employed to assess the level and trend stationarity of the TB incidence data. \textit{Linearity} of a time series determines whether its future values can be expressed as a linear combination of past observations. This feature influences model selection based on their ability to capture linear or non-linear dynamics. The Teräsvirta neural network test is used in this study to detect the presence of non-linear patterns in TB incidence cases. \textit{Seasonality} represents recurring patterns within a defined time interval. We apply the Ollech and Webel combined seasonality test to identify seasonal behavior in TB incidence. All the statistical analyses were performed using R statistical software, following \cite{panja2023epicasting}.

\begin{table}[!ht]
\caption{Global features of TB incidence cases at 47 prefectures of Japan. In the table, (A) and (Q) indicate annual and quarterly seasonality.}
\setlength\tabcolsep{2pt}
\tiny{
% \resizebox{0.99\textwidth}{!}{
\begin{tabular}{|c|cccccc|cc|}
\hline
\multirow{2}{*}{Prefecture No.}& \multicolumn{6}{c|}{Descriptive Statistics} & \multicolumn{2}{c|}{Statistical Properties}\\
 & Mean   & Range       & Sd     & CV    & Skewness & Kurtosis & Hurst Exponent & Characteristics                       \\ \hline
1             & 72.56  & (33 , 163)  & 25.59  & 35.26 & 0.76     & -0.03    & 0.82            & Non-stationary, Nonlinear, Seasonality (A) \\
2             & 24.57  & (8 , 61)    & 9.99   & 40.66 & 0.83     & 0.53     & 0.80            & Non-stationary, Linear                     \\
3             & 17.86  & (5 , 48)    & 7.91   & 44.32 & 1.06     & 1.46     & 0.79            & Non-stationary, Linear                     \\
4             & 28.24  & (6 , 71)    & 11.52  & 40.78 & 0.90     & 0.58     & 0.81            & Non-stationary, Nonlinear                  \\
5             & 14.95  & (2 , 38)    & 6.84   & 45.79 & 0.81     & 0.49     & 0.78            & Non-stationary, Linear                     \\
6             & 13.81  & (4 , 36)    & 5.92   & 42.84 & 1.37     & 2.20     & 0.75            & Non-stationary, Linear                     \\
7             & 26.63  & (7 , 75)    & 11.27  & 42.33 & 1.12     & 1.68     & 0.80            & Non-stationary, Nonlinear                  \\
8             & 44.24  & (22 , 84)   & 12.97  & 29.33 & 0.68     & -0.18    & 0.81            & Non-stationary, Linear, Seasonality (A)    \\
9             & 28.00  & (9 , 59)    & 9.62   & 34.33 & 0.71     & 0.17     & 0.80            & Non-stationary, Linear                     \\
10            & 25.77  & (10 , 70)   & 10.06  & 39.04 & 1.17     & 2.05     & 0.79            & Non-stationary, Linear                     \\
11            & 113.38 & (59 , 232)  & 26.85  & 23.68 & 0.98     & 1.50     & 0.81            & Non-stationary, Linear, Seasonality (A)    \\
12            & 108.09 & (53 , 190)  & 29.65  & 27.43 & 0.61     & -0.14    & 0.81            & Non-stationary, Linear, Seasonality (A)    \\
13            & 294.84 & (165 , 544) & 60.51  & 20.52 & 0.62     & 0.79     & 0.81            & Non-stationary, Linear, Seasonality (A, Q) \\
14            & 152.68 & (91 , 306)  & 37.16  & 24.34 & 0.94     & 1.22     & 0.81            & Non-stationary, Linear, Seasonality (A)    \\
15            & 32.01  & (11 , 65)   & 11.92  & 37.23 & 0.61     & -0.19    & 0.81            & Non-stationary, Linear, Seasonality (A)    \\
16            & 17.85  & (5 , 48)    & 7.54   & 42.25 & 0.85     & 0.77     & 0.79            & Non-stationary, Nonlinear                  \\
17            & 18.46  & (5 , 47)    & 7.34   & 39.76 & 1.03     & 1.32     & 0.77            & Non-stationary, Linear                     \\
18            & 12.15  & (2 , 29)    & 4.67   & 38.46 & 0.69     & 0.78     & 0.75            & Non-stationary, Linear                     \\
19            & 10.13  & (2 , 28)    & 4.37   & 43.14 & 0.74     & 0.89     & 0.75            & Non-stationary, Linear                     \\
20            & 20.68  & (7 , 42)    & 6.27   & 30.32 & 0.70     & 1.15     & 0.75            & Non-stationary, Linear                     \\
21            & 43.77  & (18 , 107)  & 14.32  & 32.72 & 1.07     & 1.81     & 0.80            & Non-stationary, Linear                     \\
22            & 62.58  & (30 , 130)  & 17.26  & 27.58 & 0.46     & 0.16     & 0.81            & Non-stationary, Linear, Seasonality (A)    \\
23            & 154.22 & (83 , 280)  & 36.57  & 23.72 & 0.68     & 0.34     & 0.79            & Non-stationary, Nonlinear, Seasonality (A) \\
24            & 30.89  & (11 , 68)   & 10.82  & 35.03 & 0.69     & 0.21     & 0.79            & Non-stationary, Linear, Seasonality (A)    \\
25            & 21.20  & (6 , 49)    & 7.39   & 34.88 & 0.78     & 0.94     & 0.77            & Non-stationary, Nonlinear, Seasonality (A) \\
26            & 61.24  & (25 , 178)  & 28.35  & 46.29 & 2.02     & 4.69     & 0.78            & Non-stationary, Linear, Seasonality (A)    \\
27            & 286.82 & (119 , 623) & 113.25 & 39.48 & 0.89     & -0.19    & 0.83            & Non-stationary, Nonlinear, Seasonality (A) \\
28            & 129.98 & (57 , 320)  & 47.93  & 36.88 & 1.07     & 0.59     & 0.83            & Non-stationary, Nonlinear, Seasonality (A) \\
29            & 28.42  & (12 , 58)   & 9.66   & 33.99 & 0.66     & -0.16    & 0.80            & Non-stationary, Linear, Seasonality (A)    \\
30            & 23.28  & (7 , 57)    & 9.29   & 39.89 & 1.07     & 0.96     & 0.79            & Non-stationary, Nonlinear                  \\
31            & 9.49   & (1 , 27)    & 4.39   & 46.31 & 0.90     & 1.33     & 0.76            & Non-stationary, Linear                     \\
32            & 11.85  & (3 , 27)    & 4.24   & 35.77 & 0.59     & 0.44     & 0.73            & Non-stationary, Linear                     \\
33            & 31.19  & (11 , 63)   & 10.72  & 34.37 & 0.76     & 0.36     & 0.79            & Non-stationary, Linear, Seasonality (A)    \\
34            & 45.82  & (19 , 91)   & 14.17  & 30.92 & 0.79     & 0.04     & 0.79            & Non-stationary, Linear                     \\
35            & 27.77  & (9 , 71)    & 12.04  & 43.34 & 1.05     & 0.82     & 0.80            & Non-stationary, Linear                     \\
36            & 17.42  & (6 , 42)    & 7.34   & 42.15 & 1.11     & 1.11     & 0.79            & Non-stationary, Nonlinear, Seasonality (A) \\
37            & 19.75  & (4 , 52)    & 7.87   & 39.84 & 0.80     & 0.93     & 0.79            & Non-stationary, Nonlinear, Seasonality (A) \\
38            & 23.68  & (8 , 49)    & 8.75   & 36.95 & 0.82     & 0.45     & 0.78            & Non-stationary, Linear, Seasonality (A)    \\
39            & 15.19  & (1 , 40)    & 6.83   & 44.95 & 0.79     & 0.31     & 0.80            & Non-stationary, Linear, Seasonality (A)    \\
40            & 99.09  & (38 , 201)  & 30.50  & 30.78 & 0.79     & 0.25     & 0.83            & Non-stationary, Linear, Seasonality (A)    \\
41            & 15.28  & (4 , 33)    & 5.21   & 34.11 & 0.41     & 0.26     & 0.74            & Non-stationary, Linear, Seasonality (A)    \\
42            & 32.10  & (12 , 63)   & 9.71   & 30.25 & 0.68     & 0.08     & 0.78            & Non-stationary, Linear, Seasonality (A)    \\
43            & 31.11  & (14 , 60)   & 8.42   & 27.07 & 0.64     & 0.29     & 0.76            & Non-stationary, Nonlinear, Seasonality (A) \\
44            & 24.86  & (9 , 69)    & 9.71   & 39.07 & 1.47     & 3.04     & 0.78            & Non-stationary, Linear                     \\
45            & 19.43  & (7 , 50)    & 8.27   & 42.57 & 1.02     & 0.81     & 0.80            & Non-stationary, Nonlinear, Seasonality (A) \\
46            & 34.78  & (10 , 65)   & 11.59  & 33.32 & 0.55     & -0.26    & 0.80            & Non-stationary, Linear, Seasonality (A)    \\
47            & 24.90  & (9 , 45)    & 6.75   & 27.11 & 0.35     & -0.24    & 0.75            & Non-stationary, Linear    \\ \hline                
\end{tabular}}
\label{Table_Global_Features}
\end{table}

{\color{black}
\begin{table}[!ht]
\caption{\color{black}Global features of TB incidence cases at 31 prefectures of China. In the table, (A) indicate annual seasonality.}
\setlength\tabcolsep{2pt}
\tiny{
% \resizebox{0.99\textwidth}{!}{
\begin{tabular}{|c|cccccc|cc|}
\hline
\multirow{2}{*}{Province No.}& \multicolumn{6}{c|}{Descriptive Statistics} & \multicolumn{2}{c|}{Statistical Properties}\\
 & Mean   & Range       & Sd     & CV    & Skewness & Kurtosis & Hurst Exponent & Characteristics                       \\ \hline
1	&	583.23	& (436, 706) &	69.13	&	0.12	&	-0.18	&	-0.96	&	0.60	&	Stationary, Nonlinear, Seasonality (A) 	\\
2	&	263.13	& (171, 344) &	38.28	&	0.15	&	-0.57	&	0.13	&	0.55	&	Stationary, Linear, Seasonality (A) 	\\
3	&	2798.79	& (2155, 3369) &	291.60	&	0.10	&	-0.16	&	-0.52	&	0.67	&	Stationary, Linear, Seasonality (A) 	\\
4	&	1252.46	& (857, 1625) &	183.02	&	0.15	&	-0.03	&	-0.44	&	0.68	&	Non-stationary, Linear	\\
5	&	1015.60	& (733, 1311) &	148.08	&	0.15	&	0.23	&	-0.58	&	0.54	&	Stationary, Linear, Seasonality (A) 	\\
6	&	1975.31	& (1357, 2584) &	263.55	&	0.13	&	0.23	&	-0.20	&	0.58	&	Stationary, Linear, Seasonality (A) 	\\
7	&	1243.17	& (737, 1732) &	234.19	&	0.19	&	0.18	&	-0.48	&	0.66	&	Non-stationary, Linear, Seasonality (A) 	\\
8	&	2619.27	& (1873, 3598) &	437.06	&	0.17	&	0.17	&	-0.67	&	0.69	&	Non-stationary, Linear, Seasonality (A) 	\\
9	&	552.88	& (435, 657) &	65.29	&	0.12	&	-0.24	&	-1.03	&	0.50	&	Stationary, Linear, Seasonality (A) 	\\
10	&	2503.90	& (1718, 3193) &	333.38	&	0.13	&	-0.12	&	-0.38	&	0.72	&	Non-stationary, Linear, Seasonality (A) 	\\
11	&	2318.58	& (1745, 2825) &	302.19	&	0.13	&	-0.23	&	-1.11	&	0.62	&	Stationary, Linear, Seasonality (A) 	\\
12	&	2951.65	& (2314, 3721) &	340.14	&	0.12	&	0.18	&	-0.52	&	0.62	&	Non-stationary, Linear, Seasonality (A) 	\\
13	&	1413.56	& (1046, 1741) &	178.00	&	0.13	&	-0.23	&	-0.74	&	0.61	&	Non-stationary, Linear, Seasonality (A) 	\\
14	&	3194.77	& (1976, 24882) &	3218.01	&	1.01	&	6.79	&	46.67	&	0.51	&	Stationary, Linear, Seasonality (A) 	\\
15	&	2649.19	& (1882, 4059) &	443.85	&	0.17	&	0.88	&	1.38	&	0.62	&	Non-stationary, Linear, Seasonality (A) 	\\
16	&	4877.46	& (3688, 5769) &	470.66	&	0.10	&	-0.10	&	-0.04	&	0.68	&	Non-stationary, Linear, Seasonality (A) 	\\
17	&	3656.77	& (2695, 4550) &	460.29	&	0.13	&	-0.23	&	-0.50	&	0.63	&	Non-stationary, Linear, Seasonality (A) 	\\
18	&	4508.90	& (3443, 5801) &	587.06	&	0.13	&	0.24	&	-0.73	&	0.65	&	Non-stationary, Linear, Seasonality (A) 	\\
19	&	6687.92	& (4987, 7976) &	729.69	&	0.11	&	-0.38	&	-0.76	&	0.57	&	Stationary, Linear, Seasonality (A) 	\\
20	&	3691.96	& (2726, 5086) &	554.50	&	0.15	&	0.30	&	-0.38	&	0.64	&	Non-stationary, Linear, Seasonality (A) 	\\
21	&	689.88	& (485, 857) &	87.91	&	0.13	&	-0.27	&	-0.42	&	0.68	&	Non-stationary, Linear, Seasonality (A) 	\\
22	&	1892.35	& (1328, 2592) &	289.06	&	0.15	&	0.09	&	-0.23	&	0.58	&	Stationary, Linear, Seasonality (A) 	\\
23	&	4523.88	& (3586, 6044) &	535.51	&	0.12	&	0.47	&	0.32	&	0.55	&	Non-stationary, Linear, Seasonality (A) 	\\
24	&	3784.88	& (2774, 5251) &	546.27	&	0.14	&	0.33	&	0.47	&	0.53	&	Stationary, Linear, Seasonality (A) 	\\
25	&	2201.13	& (1616, 2997) &	298.40	&	0.14	&	0.34	&	0.71	&	0.54	&	Stationary, Linear, Seasonality (A) 	\\
26	&	399.88	& (176, 553) &	85.51	&	0.21	&	-0.47	&	0.18	&	0.59	&	Stationary, Linear, Seasonality (A) 	\\
27	&	1809.21	& (1462, 2261) &	202.04	&	0.11	&	0.30	&	-0.95	&	0.58	&	Stationary, Linear, Seasonality (A) 	\\
28	&	1265.38	& (921, 2039) &	238.73	&	0.19	&	0.88	&	0.94	&	0.57	&	Stationary, Linear, Seasonality (A) 	\\
29	&	595.15	& (375, 831) &	106.01	&	0.18	&	0.14	&	-0.64	&	0.66	&	Non-stationary, Linear, Seasonality (A) 	\\
30	&	243.67	& (146, 394) &	48.72	&	0.20	&	0.92	&	1.47	&	0.66	&	Non-stationary, Linear, Seasonality (A) 	\\
31	&	3638.77	& (2212, 5142) &	689.76	&	0.19	&	0.19	&	-0.73	&	0.58	&	Stationary, Linear, Seasonality (A) 	\\ \hline
\end{tabular}}
\label{Table_Features_CHina}
\end{table}
}

\subsection{Parameter Calibration of the Mechanistic Model}\label{App_Param_Calli}
\begin{figure}
\centering
% \resizebox{0.75\textwidth}{!}{
\begin{tabular}[!ht]{cc}
(a) Time evolution for $S$    & (b)  Time evolution for $I$ \\
  \includegraphics[width=50mm, height=40mm]{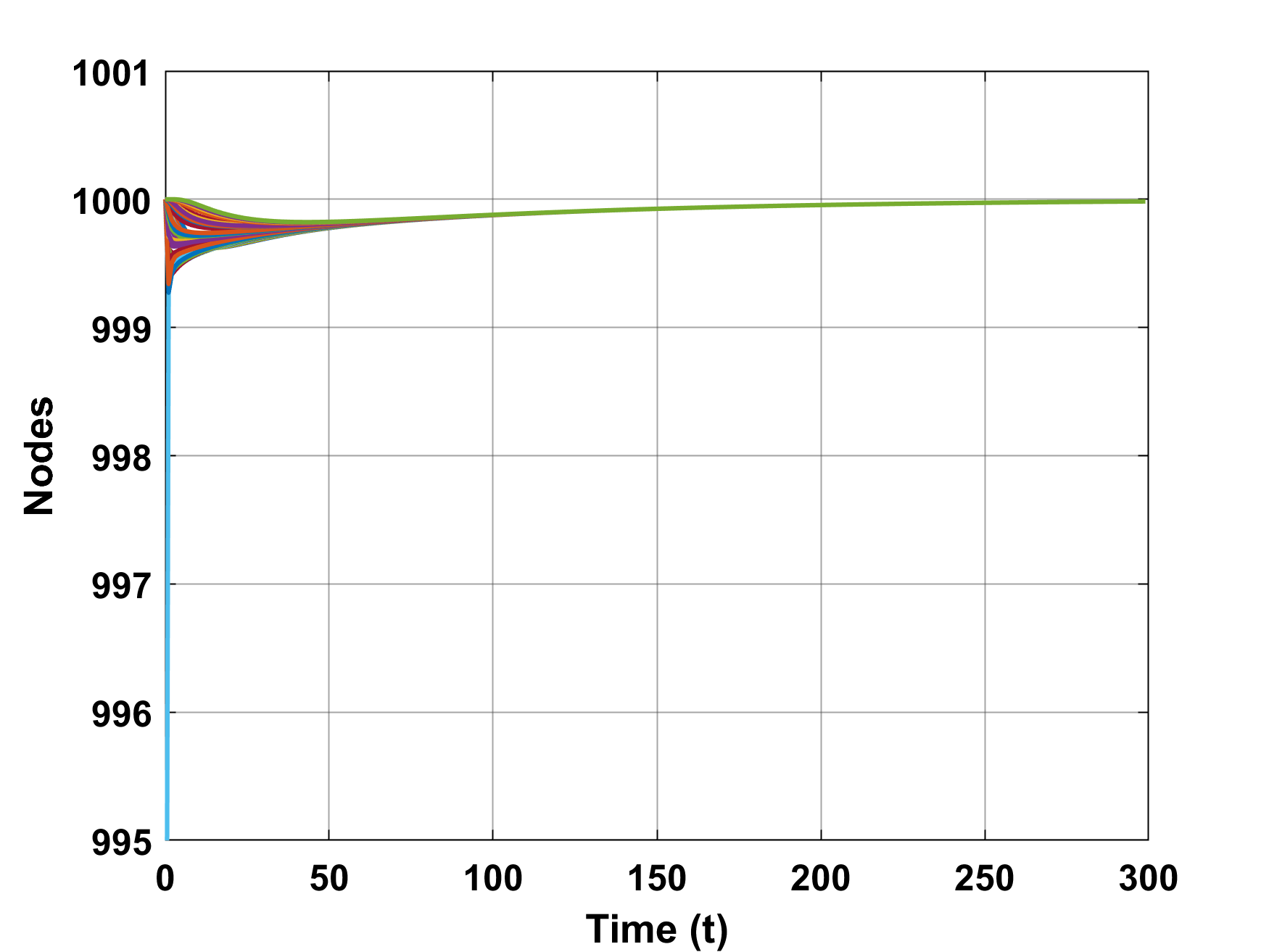} &   \includegraphics[width=50mm, height=40mm]{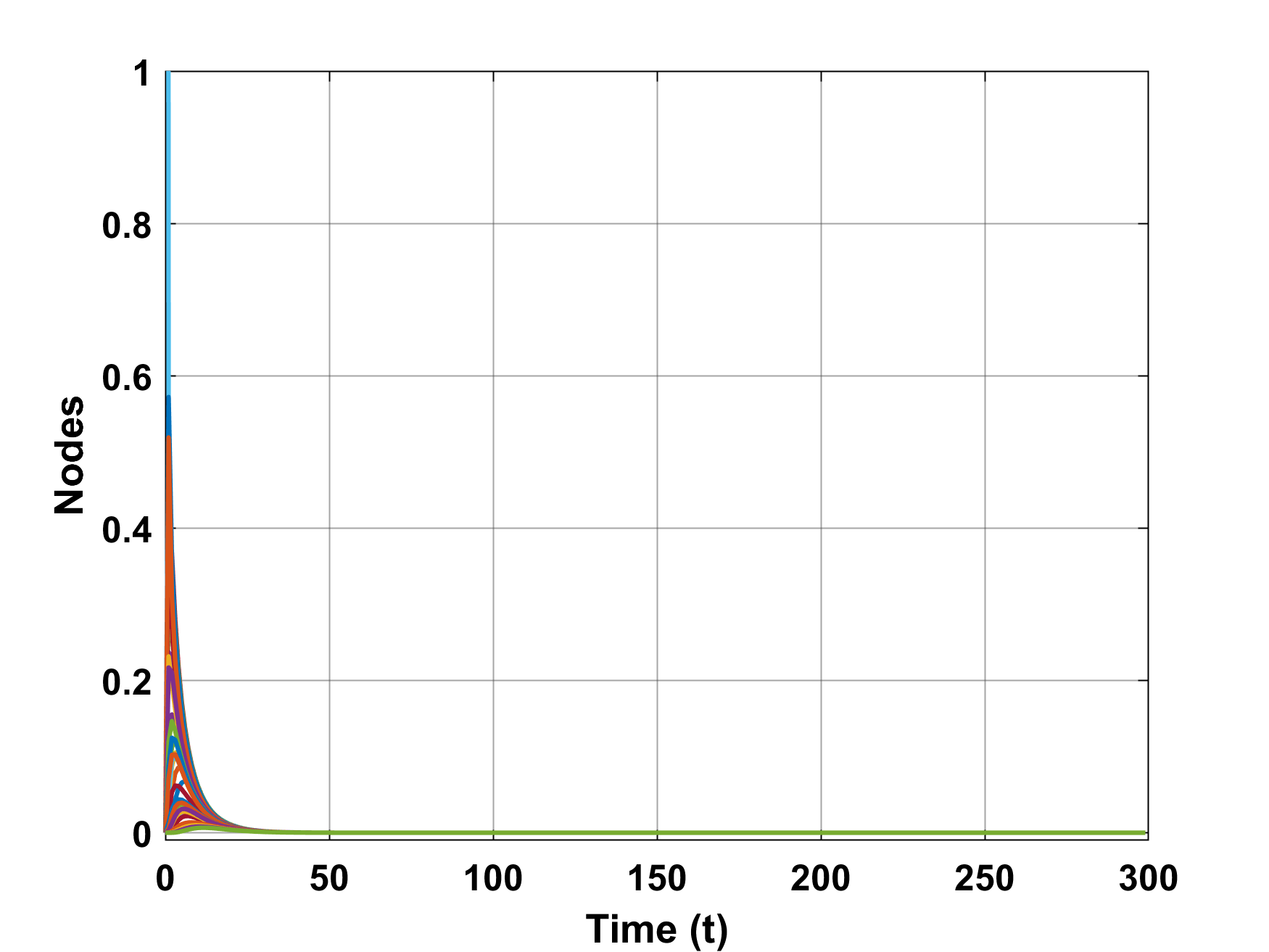} \\
(c) Time evolution for $S$    & (d) Time evolution for $I$ \\
\includegraphics[width=50mm, height=40mm]{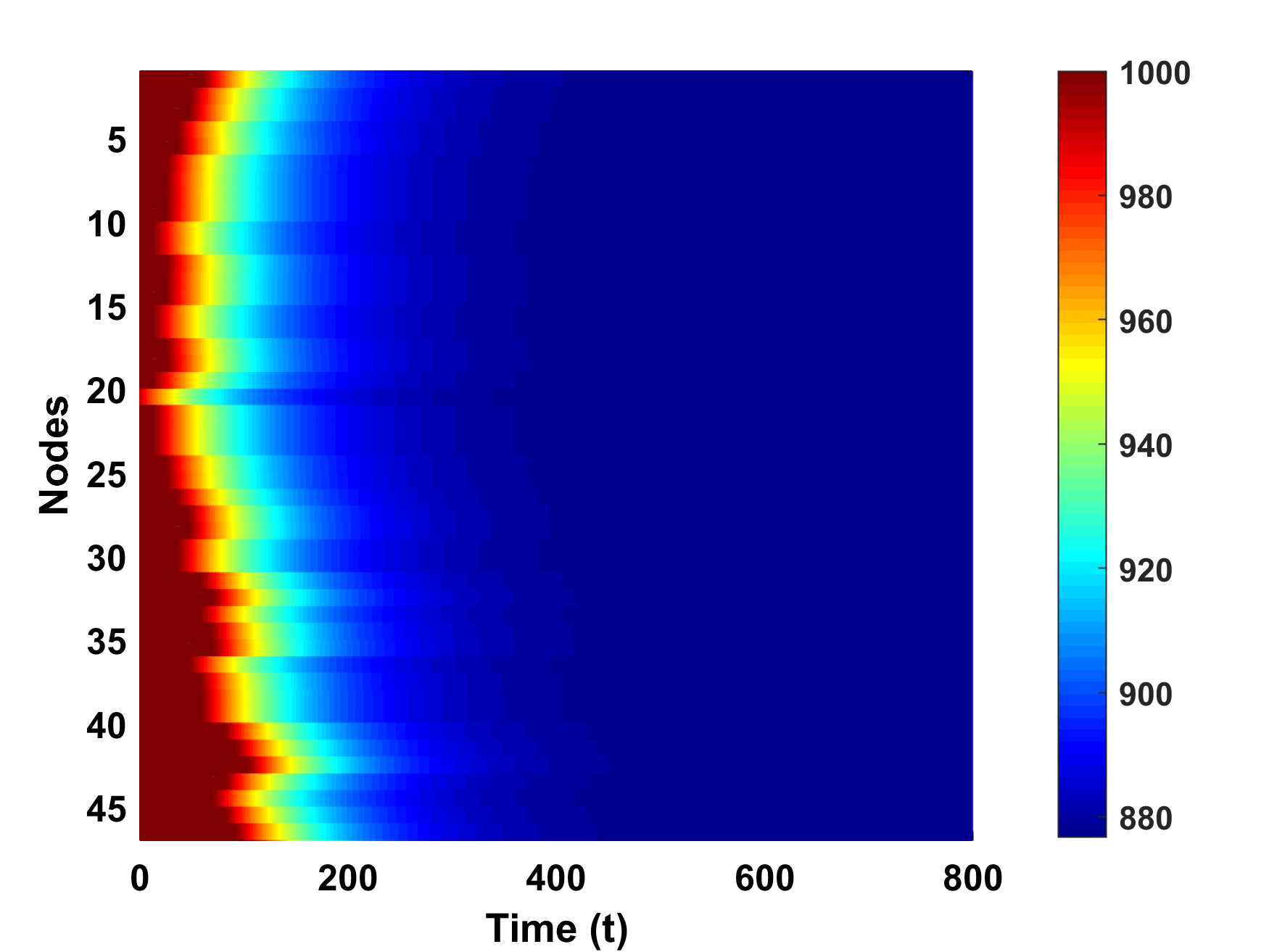} &   \includegraphics[width=50mm, height=40mm]{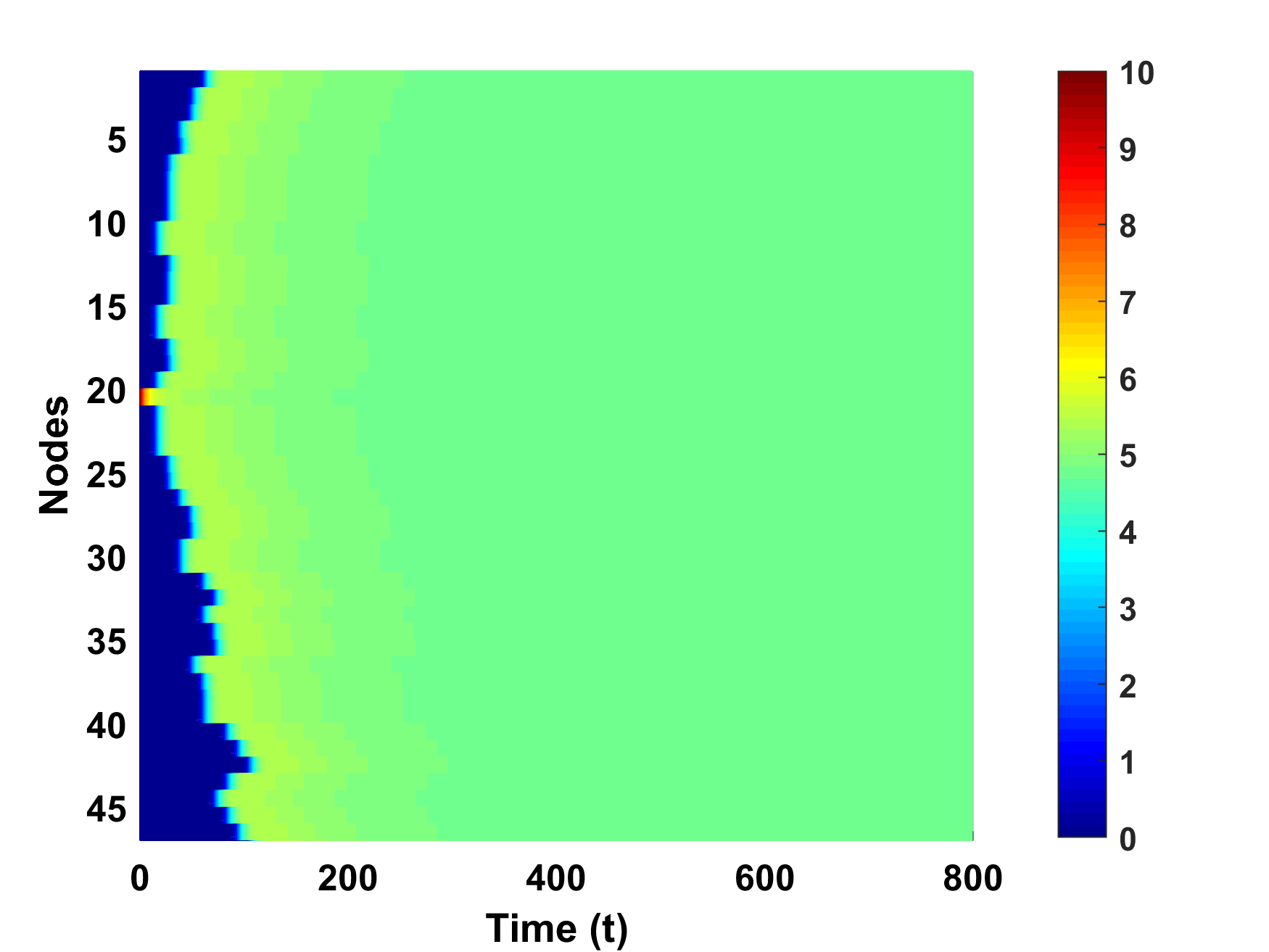}   
\end{tabular}
%85mm,height=65mm
\caption{(a) - (b) illustrates the stability plot of the disease-free equilibrium for $\mathcal{R}_0 = 0.38$ ($< 1$). (c) - (d) illustrates the stability plot of the endemic equilibrium for $\mathcal{R}_0 = 3.8$ ($> 1$). Note that along the y-axis, individuals are represented by node numbers in sequential order rather than preserving the graph structure.}
\label{fig:stab}
\end{figure}

\begin{figure}[!ht]
    \centering
    % \resizebox{0.75\textwidth}{!}{
    \begin{tabular}{cc}
        (a) Prefecture 20 & (b) Prefecture 10 \\
           \includegraphics[scale=0.3]{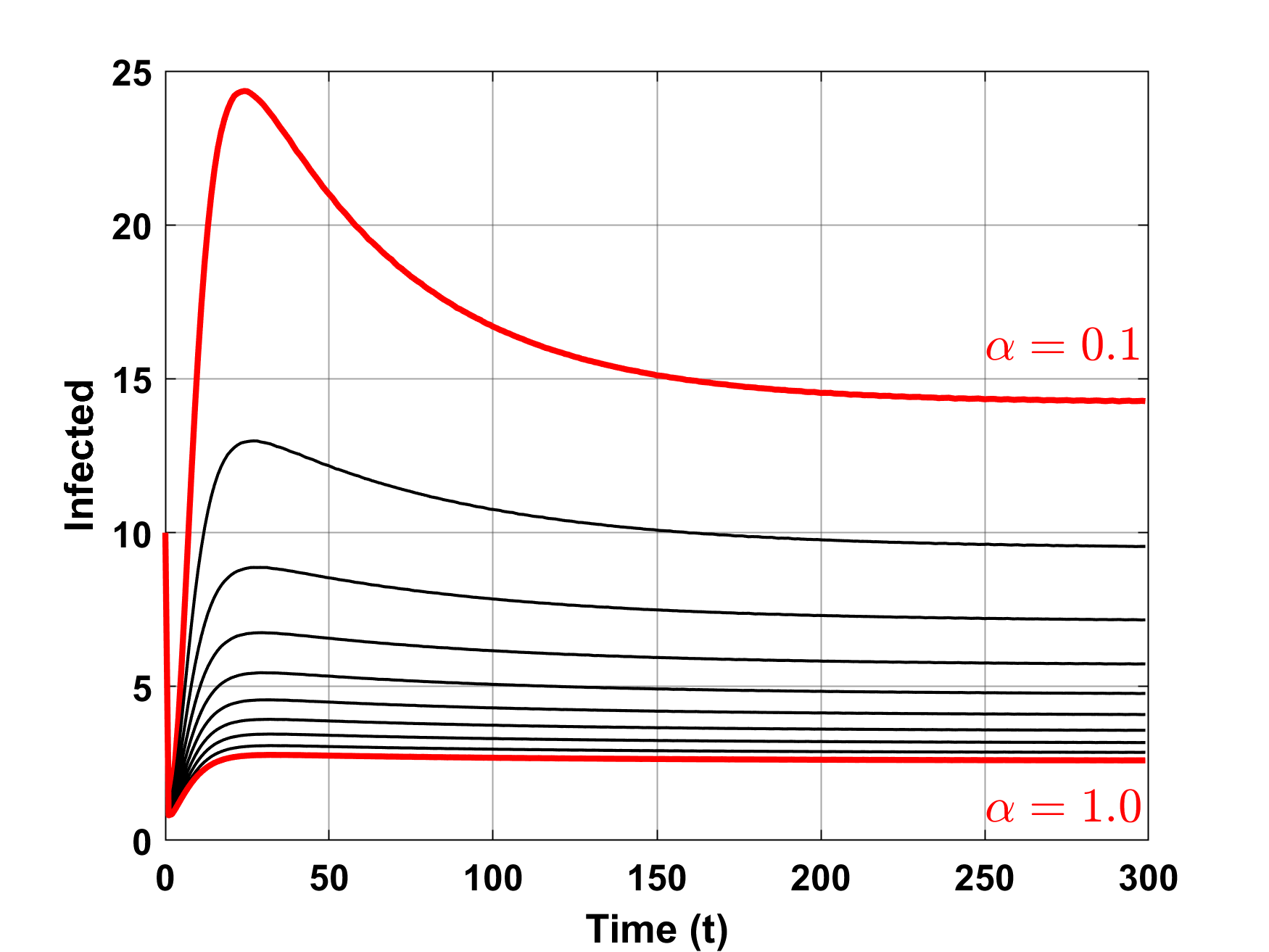}  &     \includegraphics[scale=0.3]{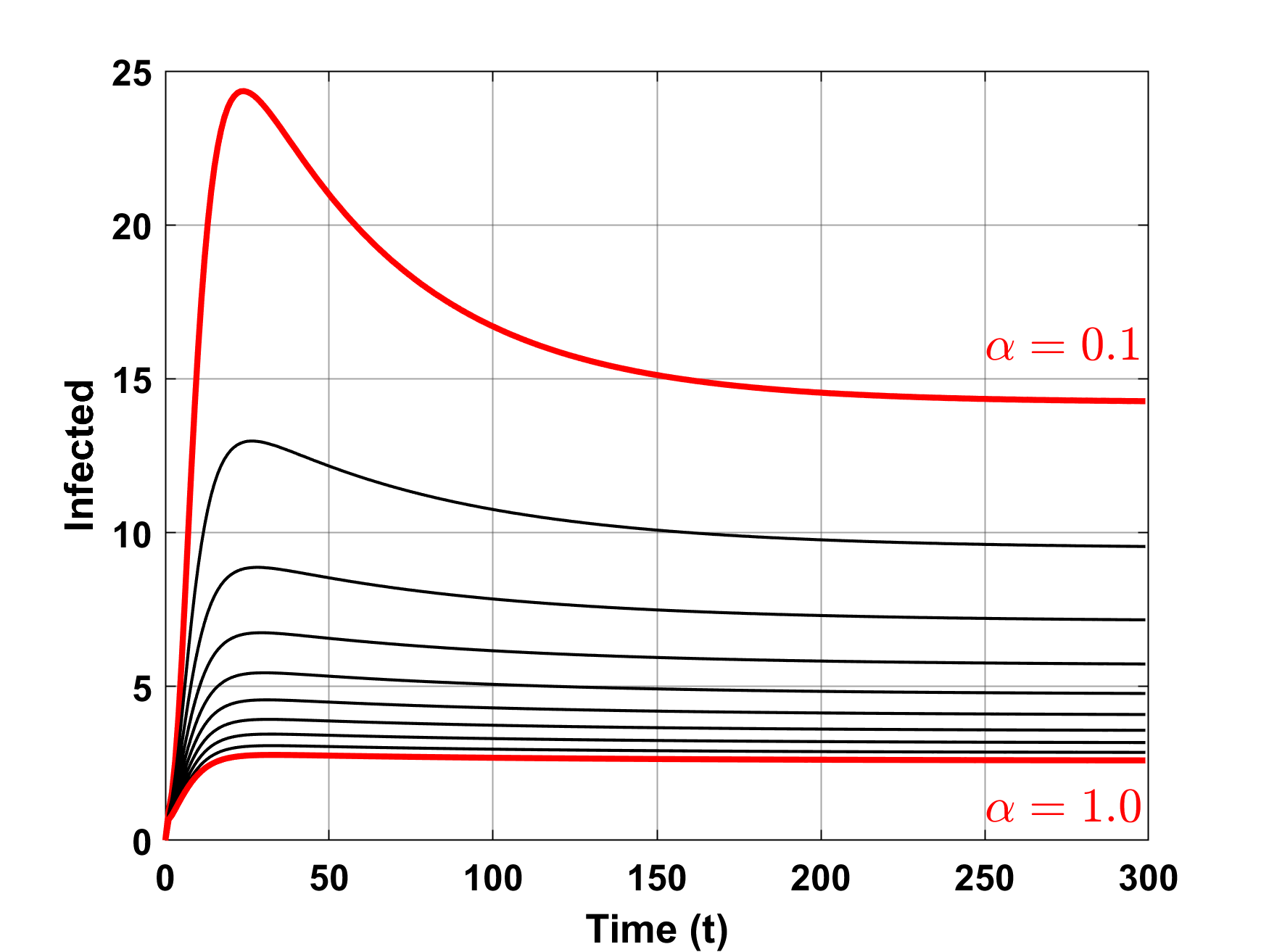}
    \end{tabular}
    \caption{Time evolution of infected individuals at (a) prefecture 20 and (b) prefecture 10 when the saturation factor  ($\alpha$) varies from $0.1$ to $1.0$.}
    \label{fig:alpha_var}
\end{figure}

\begin{figure}[!ht] 
\centering
% \resizebox{0.75\textwidth}{!}{
\begin{tabular}[!ht]{cc}
(a) Time evolution for $S$    & (b)  Time evolution for $I$ \\
  \includegraphics[width=50mm, height=40mm]{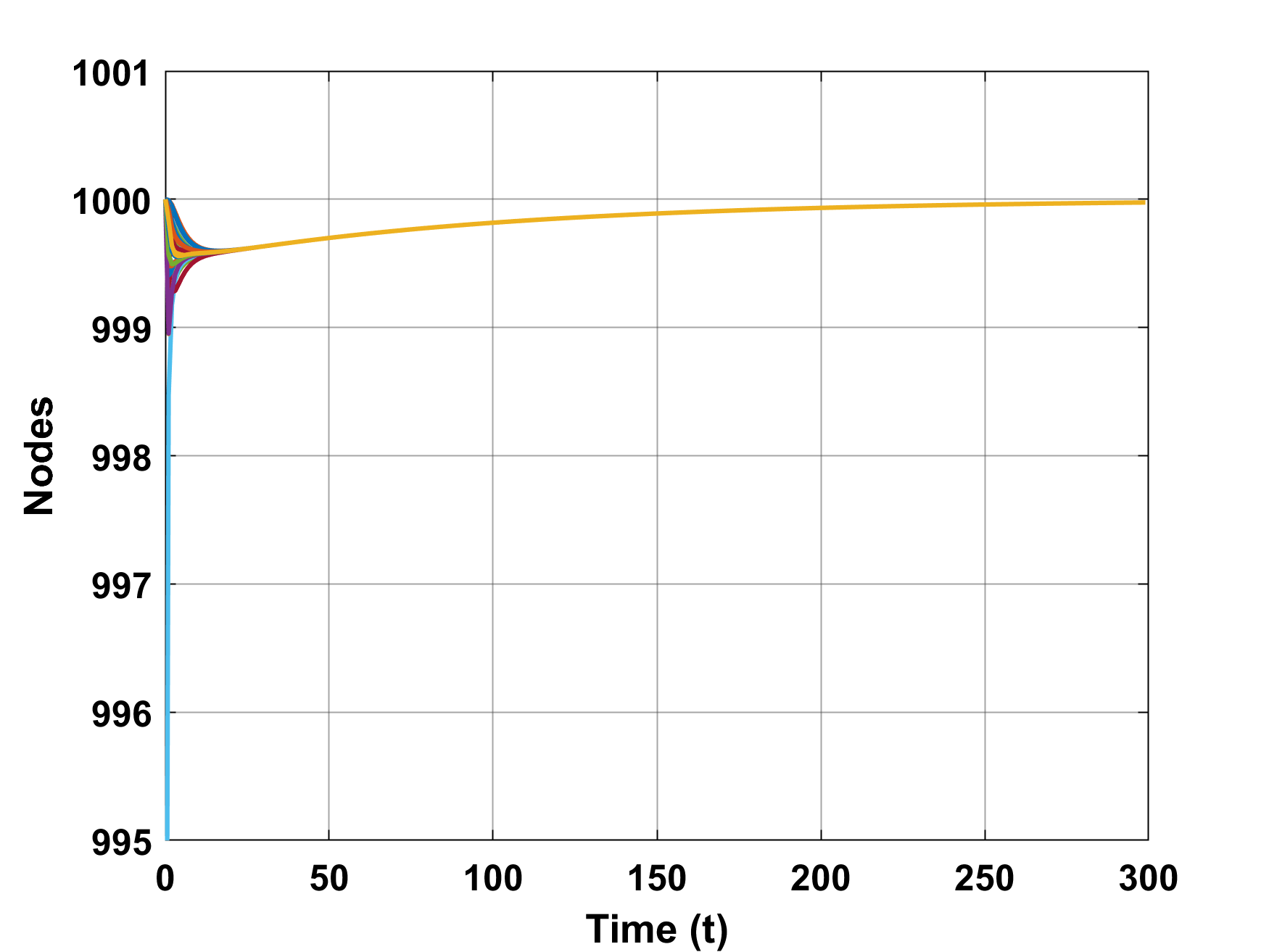} &   \includegraphics[width=50mm, height=40mm]{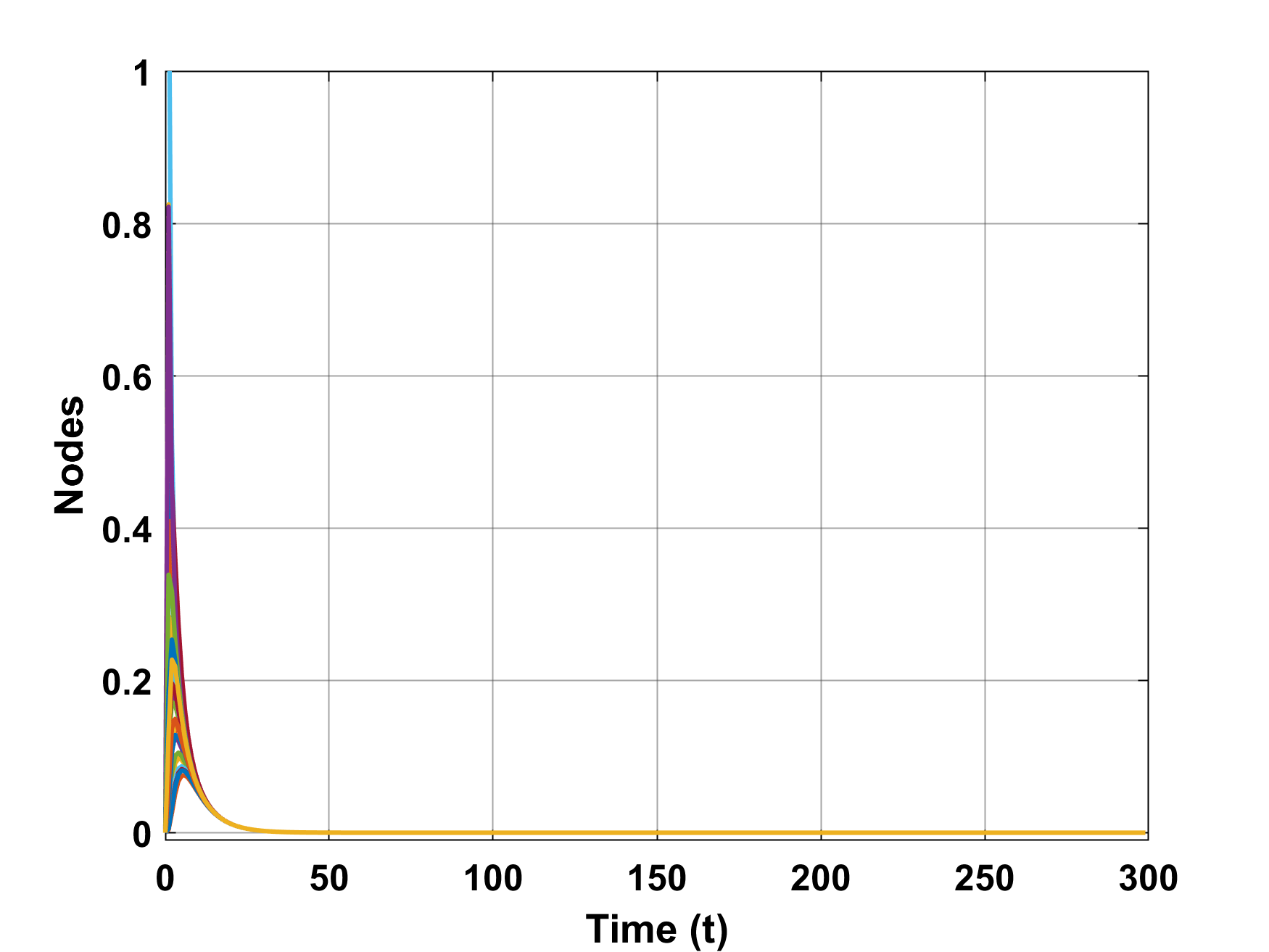} \\
(c) Time evolution for $S$    & (d) Time evolution for $I$ \\
\includegraphics[width=50mm, height=40mm]{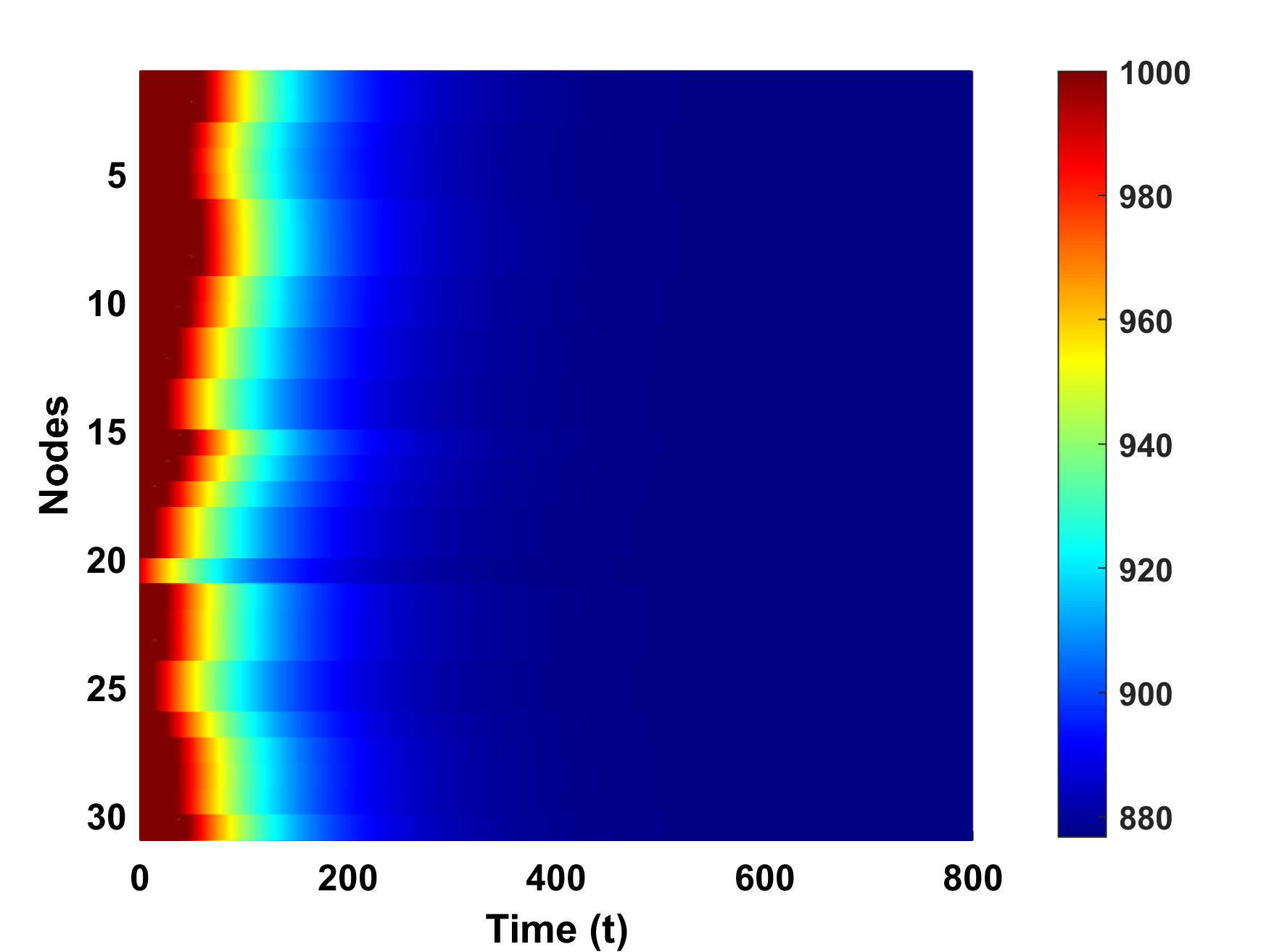} &   \includegraphics[width=50mm, height=40mm]{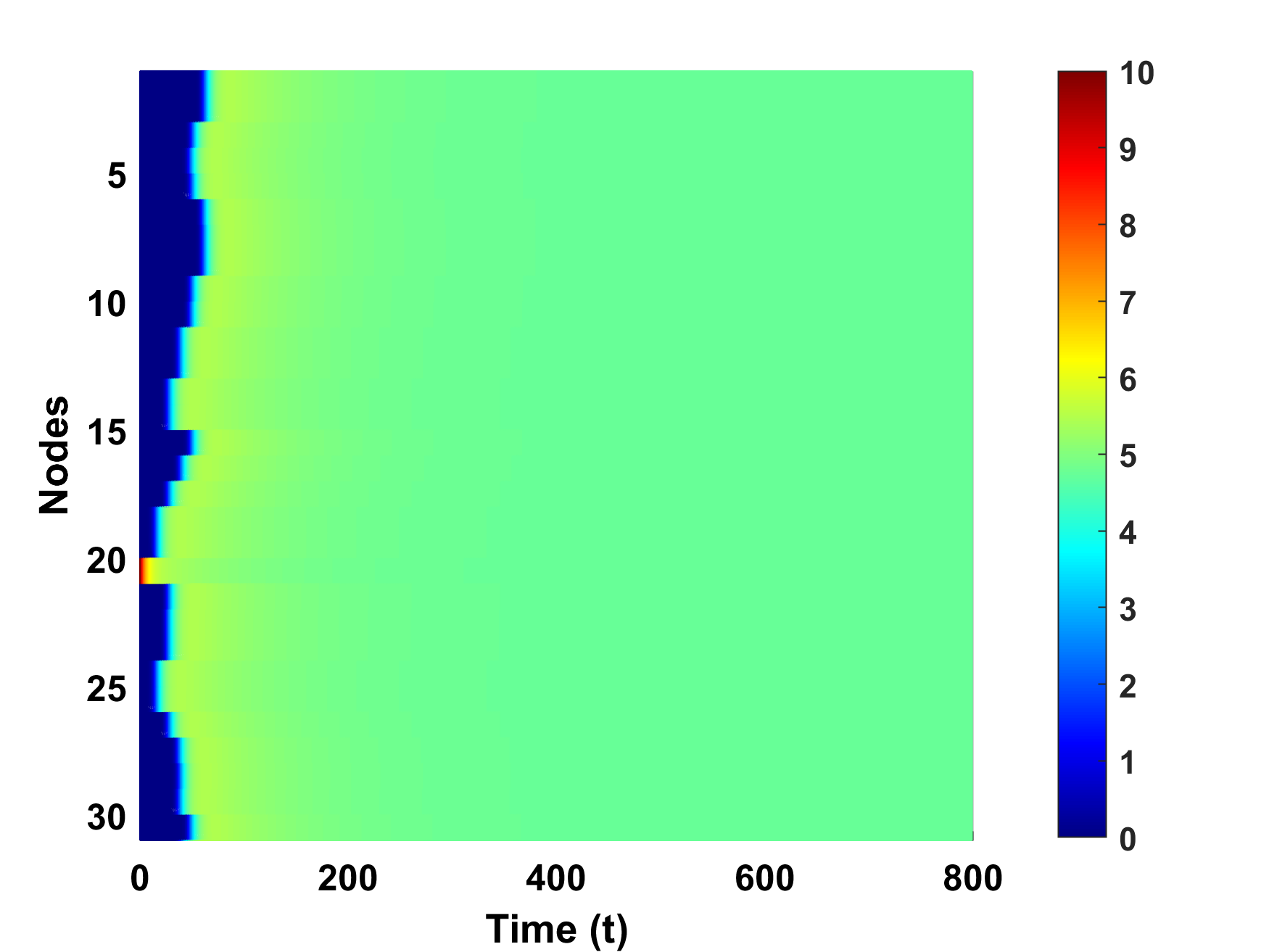}   
\end{tabular}
%85mm,height=65mm
\caption{\color{black} (a) - (b) illustrates the stability plot of the disease-free equilibrium for $\mathcal{R}_0 = 0.38$ ($< 1$) with China network. (c) - (d) illustrates the stability plot of the endemic equilibrium for $\mathcal{R}_0 = 3.8$ ($> 1$) with China network.}
\label{fig:stab_china}
\end{figure}

\begin{figure}[!ht]
    \centering
    % \resizebox{0.75\textwidth}{!}{
    \begin{tabular}{cc}
        (a) Province 20 & (b) Province 10 \\
           \includegraphics[scale=0.3]{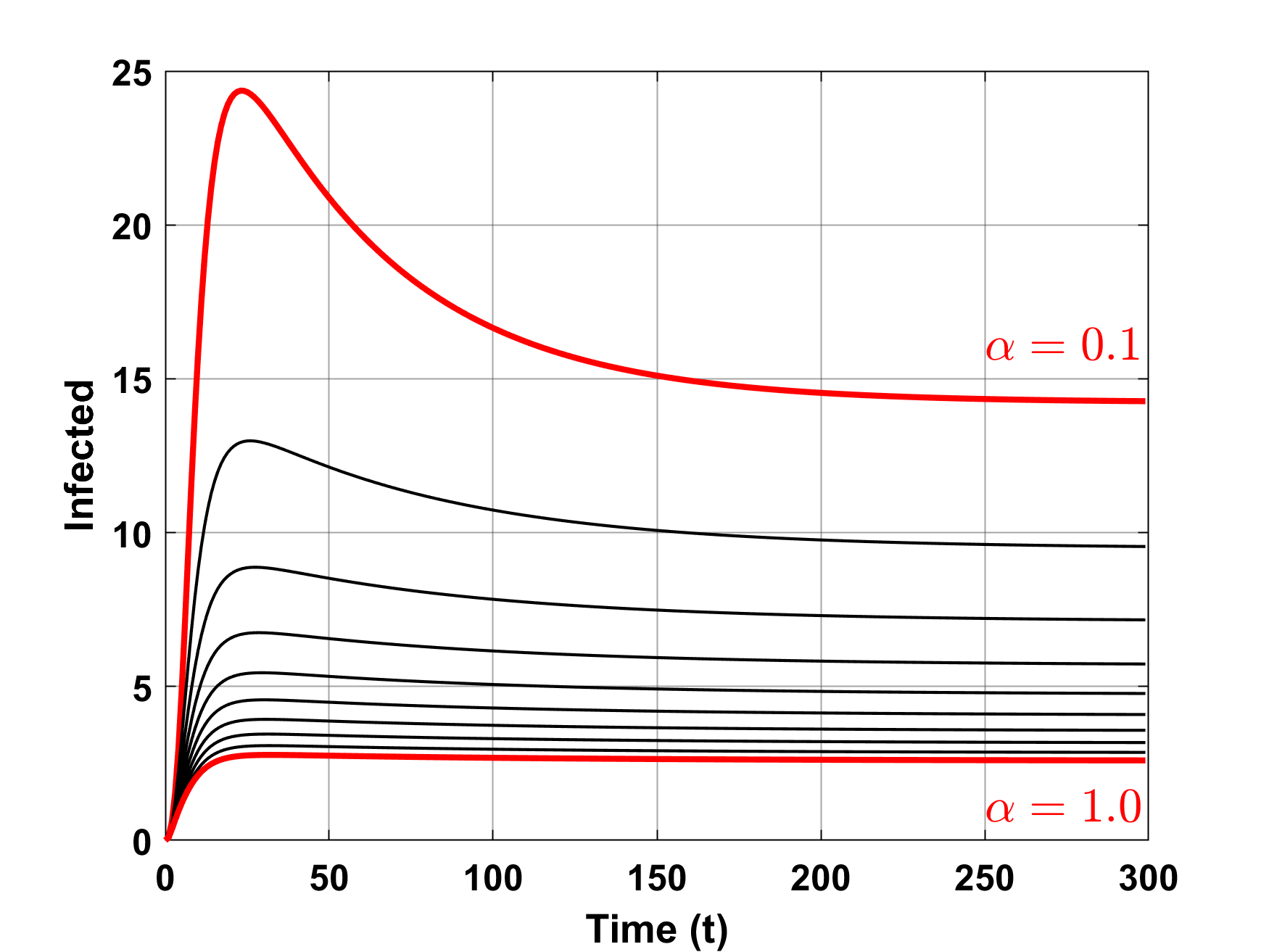}  &     \includegraphics[scale=0.3]{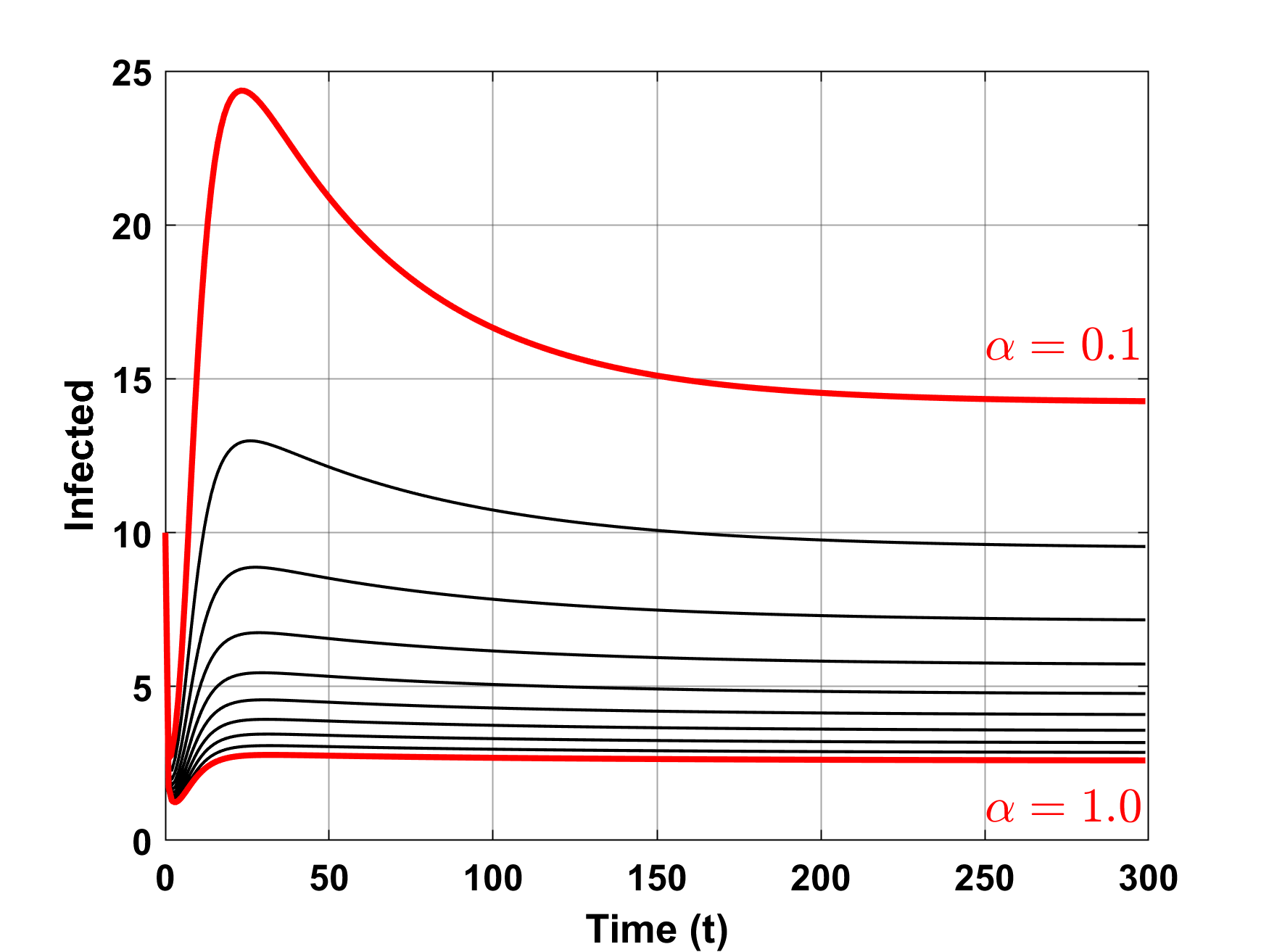}
    \end{tabular}
    \caption{\color{black} Time evolution of infected individuals with China network at (a) province 20 and (b) province 10 when the saturation factor ($\alpha$) varies from $0.1$ to $1.0$.}
    \label{fig:alpha_var_china}
\end{figure}

{\color{black}Fig. \ref{fig:stab} and \ref{fig:stab_china} depicts the stability plot of the disease-free equilibrium ($\mathcal{R}_0 = 0.38$) and endemic equilibrium ($\mathcal{R}_0 = 3.8$) for Japan and China, respectively. The time evolution of the infected individuals for selected locations of Japan and China, estimated using the MN-SIR model is presented in Fig. \ref{fig:alpha_var} and \ref{fig:alpha_var_china}, respectively.}

\subsection{Performance Measures}\label{Sec_Performance_measures}
The mathematical expression for calculating the error metrics corresponding to location $x$ is defined as follows:
{\footnotesize
$$
\begin{aligned}
    \text{ SMAPE} &= \frac{1}{q} \sum_{v=1}^q \frac{2\left|\widehat{Y}(x,{T+v}) - Y(x,{T+v})\right|}{\left|\widehat{Y}(x,{T+v})\right|+ \left|Y(x,{T+v})\right|} \times 100 \%;
    \text{ MAE} = \frac{1}{q}\sum_{v =1}^{q} \left|\widehat{Y}(x,{T+v}) - Y(x,{T+v})\right|;\\
    \text{ MASE} &= \frac{\sum_{v = 1}^{q} \left|\widehat{Y}(x,{T+v}) - Y(x,{T+v})\right|}{\frac{q}{T-1} \sum_{v = 2}^T \left|Y(x,{v}) - Y(x,{v-1})\right|}; 
    \text{ RMSE} = \sqrt{\frac{1}{q}\sum_{v =1}^{q} \left(\widehat{Y}(x,{T+v}) - Y(x,{T+v})\right)^2};
    \end{aligned}
$$}
where $q$ denotes the forecast horizon, $\widehat{Y}(x,{T+v})$ is the $v^{th}$-step ahead forecast based on $T$ historical data for location $x$, and $Y(x,{T+v})$ represents the corresponding ground truth.

\subsection{Statistical Significance}\label{Sec_Stat_Signif}
\begin{figure}
    \centering
    \includegraphics[width=0.9\linewidth]{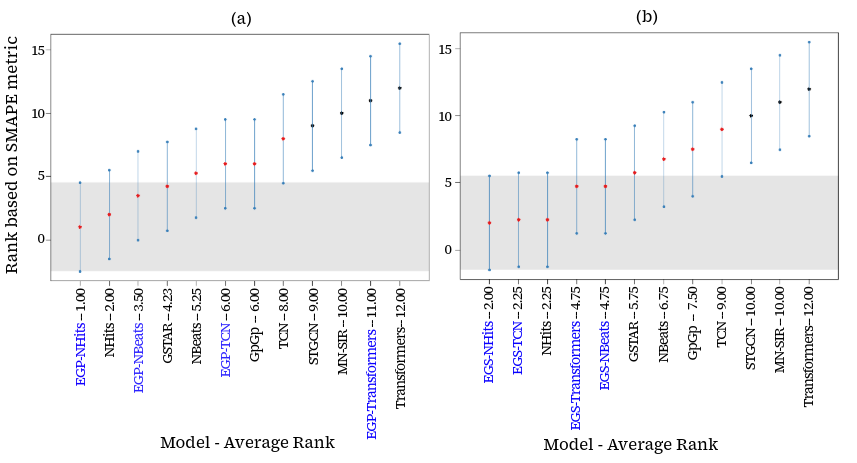}
    \caption{The MCB test results, comparing (a) EGDL-Parallel and (b) EGDL-Series architectures with baseline models using the SMAPE metric. In the plots, the Y-axis represents the average ranks of the models, while the X-axis labels each model along with its corresponding average rank. For instance, the label `EGP-NHits - 1.00' indicates that the EGP-NHits model achieved an average rank of 1.00, similar to other approaches. A lower rank indicates a better-performing model based on the evaluation metric used.}
    \label{fig:MCB_Test}
\end{figure}

\begin{figure}
    \centering
    \includegraphics[width=0.85\linewidth]{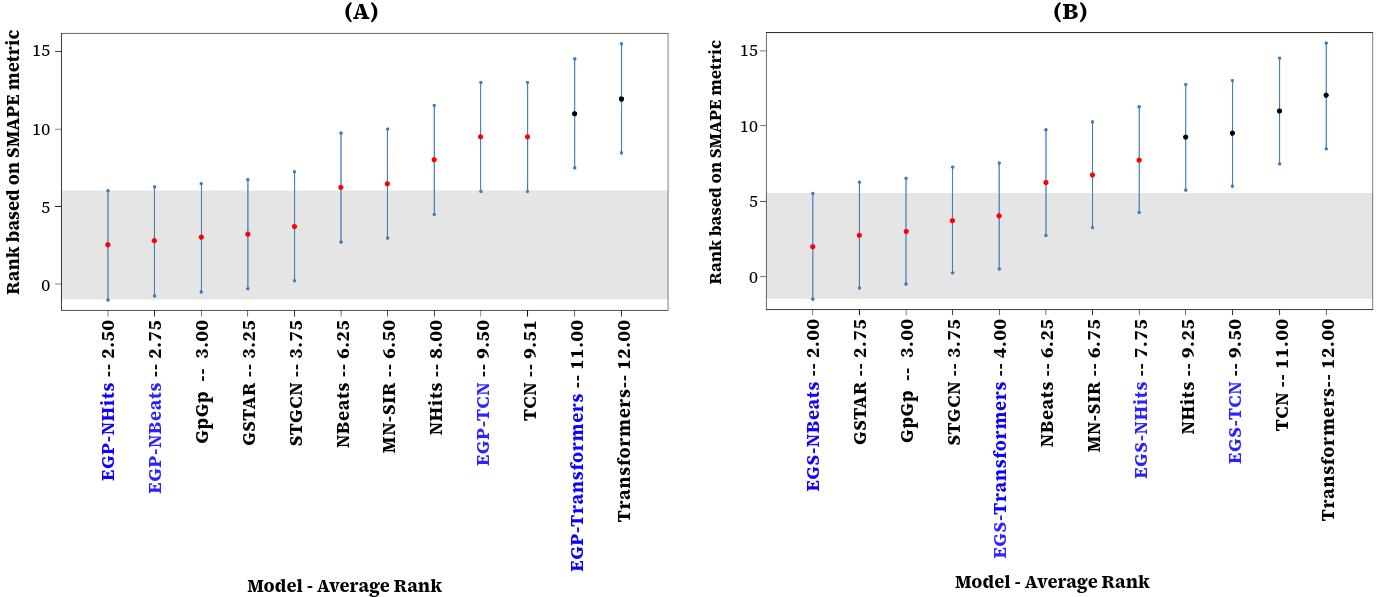}
    \caption{{\color{black}The MCB test results, comparing (A) EGDL-Parallel and (B) EGDL-Series architectures with baseline models using the SMAPE metric for forecasting TB incidence cases of China. In the plots, the Y-axis represents the average ranks of the models, while the X-axis labels each model along with its corresponding average rank. For instance, the label `EGP-NHits - 2.50' indicates that the EGP-NHits model achieved an average rank of 2.50, similar to other approaches. A lower rank indicates a better-performing model based on the evaluation metric used.}}
    \label{MCB_Plot_China}
\end{figure}

To validate the robustness of the proposed EGDL-Parallel and EGDL-Series frameworks, we employ the multiple comparisons with the best (MCB) test \cite{edwards1983multiple}. This model-agnostic procedure identifies the framework with the lowest average rank as the `best' performing model. To compare the relative performance of the architectures, the MCB test calculates a critical distance (CD) based on the Tukey distribution and utilizes the CD of the `best' model as the reference threshold. {\color{black}The MCB test results, based on the SMAPE metric, comparing EGDL-Parallel and EGDL-Series forecasters with baseline models, are visualized in Fig. \ref{fig:MCB_Test} and \ref{MCB_Plot_China} for Japan and China, respectively.} For Japan's TB dataset, among the EGDL-Parallel architectures, the EGP-NHits framework achieves the lowest rank of 1.00, followed by the NHits and EGP-NBeats frameworks. Similarly, for the EGDL-Series architectures, the EGS-NHits model has the lowest rank of 2.00, followed by EGS-TCN and NHits. {\color{black}In case of TB incidence cases of China, the EGP-NHits and EGS-NBeats frameworks achieve the minimum ranks of 2.50 and 2.00, respectively.} These results indicate that the epidemic-guided NHits and NBeats models consistently deliver the `best' performance among the evaluated frameworks. The CD for the `best' models, highlighted by the shaded regions, serves as the reference value in the MCB test. The competing architectures with CDs exceeding this reference value are deemed to perform significantly worse than the `best' performing frameworks. %Additionally, the performance of the STGCN model is notably inferior to most of the competing architectures, primarily due to the small sample size of the dataset. 
Notably, all epidemic-guided forecasting approaches exhibit lower average ranks compared to their respective baseline architectures, underscoring the effectiveness of incorporating epidemic principles into data-driven forecasting models.

\end{appendices}

\newpage
\bibliography{bibliography}% common bib file

\end{document}